\documentclass[11pt]{article}
\usepackage{smile}
\usepackage{enumerate,mathrsfs}
\usepackage[OT1]{fontenc}
\usepackage[colorlinks,linkcolor=red,anchorcolor=blue,citecolor=blue]{hyperref}
\usepackage{hhline}
\usepackage{fullpage}

\usepackage[protrusion=true, expansion=true, tracking=true, spacing=true, final, babel]{microtype}
\def\##1\#{\begin{align}#1\end{align}}
\def\$#1\${\begin{align*}#1\end{align*}}
\usepackage{setspace,chngcntr}
\usepackage{makeidx}

\newcommand{\smallvertiii}[1]{{\vert\kern-0.25ex\vert\kern-0.25ex\vert #1  \vert\kern-0.25ex\vert\kern-0.25ex\vert}}    
\newcommand{\vertiii}[1]{{\big\vert\kern-0.25ex\big\vert\kern-0.25ex\big\vert #1  \big\vert\kern-0.25ex\big\vert\kern-0.25ex\big\vert}}

\allowdisplaybreaks 

\begin{document}

\title{\huge Tensor Graphical Model: Non-convex Optimization and Statistical Inference}

\author
{
Xiang Lyu\thanks{PhD Student, Division of Biostatistics, University of California, Berkeley, Berkeley, CA 94720,  xianglyu@berkeley.edu.}
\quad Will Wei Sun\thanks{Assistant Professor, Department of Management Science, University of Miami, FL 33146, wsun@bus.miami.edu}
\quad Zhaoran Wang\thanks{Assistant Professor, Department of Industrial Engineering and Management Sciences, Northwestern University, Evanston, IL 60208, zhaoranwang@gmail.com.}
\quad Han Liu\thanks{Associate Professor, Department of Electrical Engineering and Computer Science and Department of Statistics, Northwestern University, Evanston, IL 60208, hanliu@northwestern.edu.}
\quad Jian Yang \thanks{Senior Director, Yahoo Research, Sunnyvale, CA 94089, jianyang@oath.com.}
\quad Guang Cheng\thanks{Professor, Department of Statistics, Purdue University, West Lafayette, IN 47906,  chengg@purdue.edu.}
}
 
\date{\vspace{-0.9em}}
\maketitle

\begin{abstract}
We consider the estimation and inference of graphical models that characterize the dependency structure of high-dimensional tensor-valued data. To facilitate the estimation of the precision matrix corresponding to each way of the tensor, we assume the data follow a tensor normal distribution whose covariance has a Kronecker product structure. A critical challenge in the estimation and inference of this model is the fact that its penalized maximum likelihood estimation involves minimizing a non-convex objective function. To address it, this paper makes two contributions: (i) In spite of the non-convexity of this estimation problem, we prove that an alternating minimization algorithm, which iteratively estimates each sparse precision matrix while fixing the others, attains an estimator with an optimal statistical rate of convergence. (ii) We propose a de-biased statistical inference procedure for testing hypotheses on the true support of the sparse precision matrices, and employ it for testing a growing number of hypothesis with false discovery rate (FDR) control. The asymptotic normality of our test statistic and the consistency of FDR control procedure are established. Our theoretical results are backed up by thorough numerical studies and our real applications on neuroimaging studies of Autism spectrum disorder and users' advertising click analysis bring new scientific findings and business insights. The proposed methods are encoded into a publicly available R package {\bf Tlasso}.
\end{abstract}


\section{Introduction}
\label{sec:introduction}

High-dimensional tensor-valued data are observed in many fields such as personalized recommendation systems and imaging research \cite{JT05, zheng2010, rendle2010, karatzoglou2010, allen2012, liu2013, WYSGYTRT13, LMWY13, Chu2016}. Traditional recommendation systems are mainly based on the user-item matrix, whose entry denotes each user's preference for a particular item. To incorporate additional information into the analysis, such as the temporal behavior of users, we need to consider tensor data, e.g., user-item-time tensor. For another example, functional magnetic resonance imaging (fMRI) data can be viewed as a three-way tensor since it contains brain measurements taken on different locations over time under various experimental conditions. Also, in the example of microarray study for aging \cite{zahn2007}, thousands of gene expression measurements are recorded on $16$ tissue types on $40$ mice with varying ages, which forms a four-way gene-tissue-mouse-age tensor.

In this paper, we study the estimation and inference of conditional independence structure within tensor data. For example, in the microarray study for aging we are interested in the dependency structure across different genes, tissues, ages and even mice. Assuming data are drawn from a tensor normal distribution, a straightforward way to estimate this structure is to vectorize the tensor and estimate the underlying Gaussian graphical model associated with the vector. Such an approach ignores the tensor structure and requires estimating a rather high dimensional precision matrix with an insufficient sample size. For instance, in the aforementioned fMRI application the sample size is one if we aim to estimate the dependency structure across different locations, time and experimental conditions. To address such a problem, a popular approach is to assume the covariance matrix of the tensor normal distribution is separable in the sense that it is the Kronecker product of small covariance matrices, each of which corresponds to one way of the tensor. Under this assumption, our goal is to estimate the precision matrix corresponding to each way of the tensor and recover its support. See \S\ref{sec:rw} for a detailed survey of previous work. 

The separable normal assumption imposes non-convexity on the penalized negative log-likelihood function. However, most existing literatures do not fix this gap between computational and statistical theory. As we will show in \S\ref{sec:rw}, previous work mainly focus on establishing the existence of a local optimum, rather than offering efficient algorithmic procedures that provably achieve the desired local optima. In contrast, we analyze an alternating minimization algorithm, named as Tlasso, that attains a consistent estimator after only one iteration. This algorithm iteratively minimizes the non-convex objective function with respect to each individual precision matrix while fixing the others. 

The established theoretical guarantees of the Tlasso algorithm are as follows. Suppose that we have $n$ observations from a $K$ order tensor normal distribution. We denote by $m_k$, $s_k$, $d_k$ ($k=1,\ldots, K$) the dimension, sparsity, and max number of non-zero entries in each row of the $k$-th way precision matrix. Besides, we define $m=\prod_{k=1}^K m_k$. The $k$-th precision matrix estimator from the Tlasso algorithm achieves a $\sqrt{ m_k(m_k+s_k)\log m_k/(nm)}$ convergence rate in Frobenius norm, which is minimax-optimal in the sense it is the optimal rate one can obtain even when the rest $K-1$ true precision matrices are known \cite{cai2015}. Moreover, under an extra irrepresentability condition, we establish a $\sqrt{ {m_k\log m_k }/{(nm)} }$ convergence rate in max norm, which is also optimal, and a $d_k\sqrt{ {m_k\log m_k }/{(nm)} }$ convergence rate in spectral norm. These estimation consistency results, together with a sufficiently large signal strength condition, further imply the model selection consistency of edge recovery. Notably, these results demonstrate that, when $K \geq 3$, the Tlasso algorithm achieves above estimation consistency even if we only have access to one tensor sample, which is often the case in practice. This phenomenon was never observed in the previous work.

The dependency structure in tensor makes support recovery very challenging. To the best of our knowledge, no previous work has been established on tensor precision matrix inference. In contrast, we propose a multiple testing method. This method tests all the off-diagonal entries of precision matrix, built upon the estimator from the Tlasso algorithm. To further balance the performance of multiple tests, we develop a false discovery rate (FDR) control procedure. This procedure selects a sufficiently small critical value across all tests. In theory, the test statistic is shown to be asymptotic normal after standardization, and hence provides a valid way to construct confidence interval for the entries of interest. Meanwhile, FDR asymptotically converges to a pre-specific level. An interesting theoretical finding is that our testing method and FDR control are still valid even for any fixed sample size as long as dimensionality diverges. This phenomenon is mainly due to the utilization of tensor structure information corresponding to each mode.

In the end, we conduct extensive experiments to evaluate the numerical performance of the proposed estimation and testing procedures. 
Under the guidance of theory, we also propose a way to significantly accelerate the alternating minimization algorithm without sacrificing estimation accuracy. 
In the multiple testing method, we empirically justify the proposed FDR control procedure by comparing the results with the oracle inference results which assume the true precision matrices are known. Additionally, analyses of two real data, i.e., the Autism spectrum disorder neuroimaging data and advertisement click data from a major Internet company, are conducted,  in which several interesting findings are revealed. For example, differential brain functional connectivities appear on postcentral gyrus, thalamus, and temporal lobe between autism patients and normal controls. Also, sports news and weather news are strongly dependent only on PC, while magazines are significantly interchained only on mobile.  


\subsection{Related Work and Contribution}\label{sec:rw}

A special case of our sparse tensor graphical model (when $K=2$) is the sparse matrix graphical model, which is studied by \cite{leng2012, yin2012,tsiligkaridis2013,zhou2014}. In particular, \cite{leng2012} and \cite{yin2012} only establish the existence of a local optima with desired statistical guarantees. Meanwhile, \cite{tsiligkaridis2013} considers an algorithm that is similar to ours. However, the statistical rates of convergence obtained by \cite{yin2012, tsiligkaridis2013} are much slower than ours when $K=2$. See Remark \ref{remark:matrix} for a detailed comparison. For $K=2$, our statistical rate of convergence in Frobenius norm recovers the result of \cite{leng2012}. In other words, our theory confirms that the desired local optimum studied by \cite{leng2012} not only exists, but is also attainable by an efficient algorithm. In addition, for matrix graphical models, \cite{zhou2014} establishes the statistical rates of convergence in spectral and Frobenius norms for the estimator attained by a similar algorithm. Their results achieve estimation consistency in spectral norm with only one matrix observation. However, their rate is slower than ours with $K=2$. See Remark \ref{remark:spectral} for detailed discussions. Furthermore, we allow $K$ to increase and establish estimation consistency even in Frobenius norm for $n=1$. Most importantly, all these results focus on matrix graphical model and can not handle the aforementioned motivating applications such as the gene-tissue-mouse-age tensor dataset.

In the context of sparse tensor graphical model with a general $K$, \cite{he2014} show the existence of a local optimum with desired rates, but do not prove whether there exists an efficient algorithm that provably attains such a local optimum. In contrast, we prove that our alternating minimization algorithm achieves an estimator with desired statistical rates. To achieve it, we apply a novel theoretical framework to consider the population and sample optimizers separately, and then establish the one-step convergence for the population optimizer (Theorem \ref{thm:population_tensor}) and the optimal rate of convergence for the sample optimizer (Theorem \ref{thm:statistical_error}). A new concentration result (Lemma \ref{lemma:sample_cov}) is developed for this purpose, which is also of independent interest. Moreover, we establish additional theoretical guarantees including the optimal rate of convergence in max norm, the estimation consistency in spectral norm, and the graph recovery consistency of the proposed sparse precision matrix estimator.

In addition to the literature on graphical models, our work is also related to another line of work about nonconvex optimization problems. See, e.g., \cite{jain2013low, agarwal2013learning, netrapalli2013phase, yi2013alternating, arora2013new,  hardt2014fast, hardt2014understanding, hardt2014computational, arora2014more, sun2015complete, arora2015simple, HV15, STLZ17} among others. These existing results mostly focus on problems such as dictionary learning, phase retrieval and matrix decomposition. Hence, our statistical model and analysis are completely different from theirs.

Our work also connects with a recent line of work on Bayesian tensor factorization \cite{hoff2011,  CZ09, XCHSG10, XYQ12, PWGCDC14, hoff2016, ZZC15}. In particular, they model covariance structure along each mode of a single tensor as an intermediate step in their tensor factorization. These covariance structures are imposed on core tensor or factor matrices to serve as the priors. Our work is fundamentally different from these procedures as they focus on the accuracy of tensor factorization while we focus on the graphical model structure within tensor-variate data. In addition, their tensor factorization is applied on a single tensor while our procedure learns dependency structure of multiple high-dimensional tensor-valued data.




In the end, the tensor inference part of our work is related to the recent high dimensional inference work, \cite{zhang2014}, \cite{van2014} and \cite{JM15}. The other two related work are \cite{ning2016} and \cite{ZC16}. To consider the statistical inference in the vector-variate high-dimensional Gaussian graphical model, \cite{liu2013gaussian} proposes the multiple testing procedure with FDR control, \cite{jankova2014} extend the de-biased estimator to precision matrix estimation, and \cite{ren2015} consider a scaled-Lasso-based inference procedure. To extend the inference methods from the vector-variate Gaussian graphical model to the matrix-variate Gaussian graphical model, \cite{chen2015, xia2015} propose multiple testing methods with FDR control and establish their asymptotic properties. However, these existing inference work can not be directly applied to our tensor graphical model. 


\vskip4pt
\noindent{\bf Notation:}
In this paper, scalar, vector and matrix are denoted by lowercase letter, boldface lowercase letter and boldface capital letter, respectively. For a matrix $\Ab = (\Ab_{i,j})\in \RR^{d\times d}$, we denote $\|\Ab\|_{\infty}, \|\Ab\|_2, \|\Ab\|_F$ as its max, spectral, and Frobenius norm, respectively. We define $\|\Ab\|_{1,\textrm{off}} := \sum_{i\ne j} |\Ab_{i,j}|$ as its off-diagonal $\ell_1$ norm and $\smallvertiii{\Ab}_{\infty} := \max_{i} \sum_{j} |\Ab_{i,j}|$ as the maximum absolute row sum. We denote $\textrm{vec}(\Ab)$ as the vectorization of $\Ab$ which stacks the columns of the matrix $\Ab$. Let $\textrm{tr}(\Ab)$ be the trace of $\Ab$. For an index set ${\SSS} = \{(i,j), i,j \in \{1,\ldots, d\}\}$, we define $[\Ab]_{{\SSS}}$ as the matrix whose entry indexed by $(i,j)\in {\SSS}$ is equal to $\Ab_{i,j}$, and zero otherwise. For two matrices $\Ab_1 \in \RR^{m \times n}, \Ab_2 \in \RR^{p \times q}$, we denote $\Ab_1 \otimes \Ab_2  \in \RR^{mp \times nq}$ as the Kronecker product of $\Ab_1$ and $\Ab_2$. We denote $\ind_d$ as the  identity matrix with dimension $d\times d$. Throughout this paper, we use $C, C_1, C_2, \ldots$ to denote generic absolute constants, whose values may vary from line to line.

\vskip4pt
\noindent{\bf Organization:} \S\ref{sec:main} introduces the main result of sparse tensor graphical model and its efficient implementation, followed by the theoretical study of the proposed estimator in \S\ref{sec:theory_optimization}. \S\ref{sec:tensor_inference} contains all the statistical inference results including a novel test statistic for constructing confidence interval and a multiple testing procedure with FDR control. \S\ref{sec:simulation} demonstrates the superior performance of the proposed methods and performs extensive comparisons with existing methods in both parameter estimation and statistical inference. 
\S\ref{sec: real_data} illustrates analyses of two real data sets, i.e., the Autism spectrum disorder neuroimaging data and advertisement click data from a major Internet company, via the proposed testing method. 
\S\ref{sec:discussion} summarizes this article and points out a few interesting future work. Detailed technical proofs are available in supplementary material.

\section{Tensor Graphical Model}
\label{sec:main}

This section introduces our sparse tensor graphical model and an alternating minimization algorithm for solving the associated nonconvex optimization problem.

\subsection{Preliminary}
\label{sec:prelim}

We first introduce the preliminary background on tensors and adopt the notations used by \cite{kolda2009}. Throughout this paper, higher order tensors are denoted by boldface Euler script letters, e.g. ${\cal T}$. We consider a $K$-way tensor ${\cal T} \in \mathbb R^{m_1 \times m_2 \times \cdots \times m_K}$. When $K=1$ it reduces to a vector and when $K=2$ it reduces to a matrix. The $(i_1,\ldots,i_K)$-th element of the tensor ${\cal T}$ is denoted as ${\cal T}_{i_1,\ldots,i_K}$. We denote the vectorization of ${\cal T}$ as $\textrm{vec}({\cal T}) := ({\cal T}_{1,1,\ldots,1},\ldots, {\cal T}_{m_1,1,\ldots,1},\ldots,{\cal T}_{1,m_2,\ldots,m_K},{\cal T}_{m_1,m_2,\ldots,m_K} )^{\top} \in \mathbb R^{m}$ with $m = \prod_k m_k$. In addition, we define the Frobenius norm of a tensor ${\cal T}$ as 
$$
\|{\cal T}\|_F := \sqrt{ \sum_{i_1,\ldots,i_K} {\cal T}_{i_1,\ldots,i_K}^2}.
$$

In tensors, a fiber refers to a higher order analogue of matrix row and column. A fiber is obtained by fixing all but one of the indices of the tensor, e.g., for a tensor ${\cal T}$, the mode-$k$ fiber is given by ${\cal T}_{i_1,\ldots,,i_{k-1},:,i_{k+1},\ldots,i_K}$. Matricization, also known as unfolding, is a process to transform a tensor into a matrix. We denote ${\cal T}_{(k)}$ as the mode-$k$ matricization of a tensor ${\cal T}$. It arranges the mode-$k$ fibers to be the columns of the resulting matrix. Another useful operation in tensor is the $k$-mode product. The $k$-mode (matrix) product of a tensor ${\cal T} \in \mathbb R^{m_1 \times m_2 \times \cdots \times m_K}$ with a matrix $\Ab \in \mathbb R^{J \times m_k}$ is denoted as ${\cal T} \times_k \Ab$ and is of the size $m_1 \times \cdots \times m_{k-1} \times J \times m_{k+1} \times \cdots \times m_K$. Its entry is defined as 
$$({\cal T} \times_k \Ab)_{i_1, \ldots, i_{k-1}, j, i_{k+1}, \ldots, i_K} := \sum_{i_k =1}^{m_k} {\cal T}_{i_1,\ldots,i_K} {\Ab}_{j, i_k}.$$
Furthermore, for a list of matrices $\{\Ab_1,\ldots, \Ab_K\}$ with $\Ab_k\in \mathbb R^{m_k \times m_k}$, we define
$$
{\cal T} \times \{\Ab_1,\ldots, \Ab_K\} := {\cal T} \times_1 \Ab_1 \times_2 \cdots \times_K \Ab_K.
$$

\subsection{Statistical Model}

A tensor ${\cal T} \in \mathbb R^{m_1 \times m_2 \times \cdots \times m_K}$ follows the tensor normal distribution with zero mean and covariance matrices $\bSigma_1, \ldots, \bSigma_K$, denoted as ${\cal T} \sim \textrm{TN}({\bf0}; \bSigma_1, \ldots, \bSigma_K)$, if its probability density function is $p({\cal T}| \bSigma_1,\ldots,\bSigma_K) =$
\begin{equation}
 (2\pi)^{\frac{-m}{ 2}} \biggl\{ \prod_{k=1}^K |\bSigma_k|^{\frac{-m}{ 2m_k}}  \biggr\} \exp \big(- \|{\cal T} \times \bSigma^{\frac{-1}{2}}\|_F^2/2 \big),
\label{eqn:pdf}
\end{equation}
where $m = \prod_{k=1}^K m_k$ and $\bSigma^{-1/2} := \{\bSigma_1^{-1/2},\ldots,\bSigma_K^{-1/2}\}$. When $K=1$, this tensor normal distribution reduces to the vector normal distribution with zero mean and covariance $\bSigma_1$. According to \cite{kolda2009}, it can be shown that ${\cal T} \sim \textrm{TN}({\mathbf{0}}; \bSigma_1, \ldots, \bSigma_K)$ if and only if $\textrm{vec}({\cal T}) \sim \textrm{N}(\textrm{vec}({\bf0}); \bSigma_K \otimes \cdots \otimes \bSigma_1)$, where $\textrm{vec}({\bf 0}) \in \mathbb R^{m}$ and $\otimes$ is the matrix Kronecker product.

We consider the parameter estimation for the tensor normal model. Assume that we observe independently and identically distributed tensor samples ${\cal T}_1,\ldots,{\cal T}_n$ from $\textrm{TN}({\bf0}; \bSigma_1^*, \ldots, \bSigma_K^*)$. We aim to estimate the true covariance matrices $(\bSigma_1^*, \ldots, \bSigma_K^*)$ and their corresponding true precision matrices $(\bOmega_1^*, \ldots, \bOmega_K^*)$ where $\bOmega_k^* = \bSigma_k^{*-1}\ (k=1,\ldots,K)$. To address the identifiability issue in the parameterization of the tensor normal distribution, we assume that  $\|\bOmega_k^*\|_F= 1$ for $k=1,\ldots, K$. This renormalization does not change the graph structure of the original precision matrix.

A standard approach to estimate $\bOmega_k^*$, $k=1,\ldots,K$, is to use the maximum likelihood method via \eqref{eqn:pdf}. Up to a constant, the negative log-likelihood function of the tensor normal distribution is
$$
\ell(\bOmega_1, \ldots, \bOmega_K) := \frac{1}{2}\textrm{tr}[\Sbb (\bOmega_K \otimes \cdots \otimes \bOmega_1)] - \frac{1}{2}\sum_{k=1}^K \frac{m}{m_k} \log |\bOmega_k|,
$$
where $\Sbb := \frac{1}{n} \sum_{i=1}^n \textrm{vec}({\cal T}_i) \textrm{vec}({\cal T}_i)^{\top}$. To encourage the sparsity of each precision matrix in the high-dimensional scenario, we propose a penalized log-likelihood estimator which minimizes $q_n(\bOmega_1, \ldots, \bOmega_K) :=$
\begin{equation}
\frac{1}{m}\textrm{tr}[\Sbb (\bOmega_K \otimes \cdots \otimes \bOmega_1)] - \sum_{k=1}^K \frac{1}{m_k} \log |\bOmega_k| + \sum_{k=1}^K P_{\lambda_k}(\bOmega_k),
\label{eqn:sample_qn}
\end{equation}
where $P_{\lambda_k}(\cdot)$ is a penalty function indexed by the tuning parameter $\lambda_k$. In this paper, we focus on the lasso penalty \cite{tibshirani1996} $P_{\lambda_k}(\bOmega_k) = \lambda_k \sum_{i\ne j} |[\bOmega_{k}]_{i,j}| $. The estimation procedure applies similarly to a broad family of penalty functions, for example, the SCAD penalty \cite{fan2001}, the adaptive lasso penalty \cite{zou2006}, the MCP penalty \cite{zhang2010}, and the truncated $\ell_1$ penalty \cite{shen2012}. 

The penalized model from $(\ref{eqn:sample_qn})$ is called the sparse tensor graphical model. It reduces to the $m_1$-dimensional sparse gaussian graphical model \cite{yuan2007, banerjee2008, friedman2008} when $K=1$, and the sparse matrix graphical model \cite{leng2012, yin2012, tsiligkaridis2013,zhou2014} when $K=2$. Our framework generalizes them to fulfill the demand of capturing the graphical structure of the higher-order tensor-valued data.

\subsection{Estimation}

This section introduces the estimation procedure for the proposed sparse tensor graphical model. A computationally efficient algorithm is provided to alternatively estimate all precision matrices. 

Recall that in \eqref{eqn:sample_qn}, $q_n(\bOmega_1, \ldots, \bOmega_K)$ is jointly non-convex with respect to $\bOmega_1, \ldots, \bOmega_K$. Nevertheless, it turns out to be a bi-convex problem since $q_n(\bOmega_1, \ldots, \bOmega_K)$ is convex in $\bOmega_k$ when the rest $K-1$ precision matrices are fixed. The nice bi-convex property plays a critical role in our algorithm construction and its theoretical analysis in \S\ref{sec:theory_optimization}.

Based on the bi-convex property, we propose to solve this non-convex problem by alternatively updating one precision matrix with other matrices being fixed. Note that, for any $k = 1,\ldots, K$, minimizing \eqref{eqn:sample_qn} with respect to $\bOmega_k$ while fixing the rest $K-1$ precision matrices is equivalent to minimizing 
\begin{equation}
L(\bOmega_k) :=  \frac{1}{m_k}\textrm{tr}(\Sbb_k \bOmega_k) - \frac{1}{m_k} \log |\bOmega_k| + \lambda_k \|\bOmega_k\|_{1,\textrm{off}}. 
\label{eqn:one_qn}
\end{equation}
Here, $\Sbb_k := \frac{m_k}{n m}\sum_{i=1}^n \Vb_i^k \Vb_i^{k\top}$, where $\Vb_i^k := \big[ {\cal T}_i \times \big\{\bOmega_1^{1/2},\ldots,\bOmega_{k-1}^{1/2}, \ind_{m_k}, \bOmega_{k+1}^{1/2},\ldots,\bOmega_{K}^{1/2}    \big\} \big]_{(k)}$ with $\times$ the tensor product operation and $[\cdot]_{(k)}$ the mode-$k$ matricization operation defined in \S\ref{sec:prelim}. The result in \eqref{eqn:one_qn} can be shown by noting that $\Vb_i^k = [{\cal T}_i]_{(k)} \big( \bOmega_{K}^{1/2} \otimes \cdots \otimes \bOmega_{k+1}^{1/2} \otimes \bOmega_{k-1}^{1/2} \otimes \cdots \otimes \bOmega_1^{1/2}  \big)^{\top}$ according to the properties of mode-$k$ matricization shown by \cite{kolda2009}. Hereafter, we drop the superscript $k$ of $\Vb_i^k$ if there is no confusion. Note that minimizing \eqref{eqn:one_qn} corresponds to estimating vector-valued Gaussian graphical model and can be solved efficiently via the glasso algorithm \cite{friedman2008}.

\begin{algorithm}[h!]
\caption{Solve sparse tensor graphical model via Tensor lasso (Tlasso)}
\begin{algorithmic}[1]
\STATE \textbf{Input:} Tensor samples ${\cal T}_1\ldots, {\cal T}_n$, tuning parameters $\lambda_1, \ldots, \lambda_K$, max number of iterations $T$.
\STATE \textbf{Initialize} $\bOmega_1^{(0)},\ldots,\bOmega_K^{(0)}$ randomly as symmetric and positive definite matrices and set $t=0$.
\STATE \textbf{Repeat}:
\STATE $t = t + 1$.
\STATE \textbf{For} $k = 1,\ldots,K$:
\STATE \hspace{10pt} Given $\bOmega_1^{(t)}, \ldots, \bOmega_{k-1}^{(t)}, \bOmega_{k+1}^{(t-1)},\ldots, \bOmega_K^{(t-1)}$, solve \eqref{eqn:one_qn} for ${\bOmega}_k^{(t)}$ via glasso.
\STATE \hspace{10pt} Normalize ${\bOmega}_k^{(t)}$ such that $\|{\bOmega}_k^{(t)}\|_F = 1$.
\STATE \textbf{End For}
\STATE \textbf{Until $t = T$}.
\STATE \textbf{Output:} $\widehat{\bOmega}_k = {\bOmega}_k^{(T)} \ (k=1,\ldots,K)$.
\end{algorithmic}\label{alg:tlasso}
\end{algorithm}

The details of our Tensor lasso (Tlasso) algorithm are shown in Algorithm \ref{alg:tlasso}. It starts with a random initialization and then alternatively updates each precision matrix until it converges. In \S\ref{sec:theory_optimization}, we will illustrate that the statistical properties of the obtained estimator are insensitive to the choice of the initialization (see the discussion following Theorem \ref{thm:final_error}). In our numerical experiments, for each $k=1,\ldots,K$, we set the initialization of $k$-th precision matrix as $\ind_{m_k}$, which leads to superior numerical performance.

\section{Theory of Statistical Optimization}
\label{sec:theory_optimization}

We first prove the estimation errors in terms of Frobenius norm, max norm, and spectral norm, and then provide the model selection consistency of the estimator output from the Tlasso algorithm. For compactness, we defer the proofs of theorems to supplementary material.

\subsection{Estimation Error in Frobenius Norm}
\label{sec:error_F}

Based on the penalized log-likelihood in \eqref{eqn:sample_qn}, we define the population log-likelihood function as $q(\bOmega_1, \ldots, \bOmega_K) := $
\begin{equation}
 \frac{1}{m} \mathbb E \big \{   \textrm{tr} \big[ \textrm{vec}({\cal T}) \textrm{vec}({\cal T})^{\top} (\bOmega_K \otimes \cdots \otimes \bOmega_1) \big]   \big \} - \sum_{k=1}^K \frac{1}{m_k} \log |\bOmega_k|.
\label{eqn:population_q}
\end{equation}

By minimizing $q(\bOmega_1, \ldots, \bOmega_K)$ with respect to $\bOmega_k$, $k=1,\ldots, K$, we obtain the population minimization function with the parameter $\bOmega_{[K]-k} := \{\bOmega_1, \ldots, \bOmega_{k-1},\bOmega_{k+1},\ldots,\bOmega_K\}$, i.e.,
\begin{equation}
M_{k}(\bOmega_{[K]-k}) :=  \argmin_{\bOmega_k} q(\bOmega_1, \ldots, \bOmega_K).
\label{eqn:population_Mk}
\end{equation}

Our first theorem shows an interesting result that the above population minimization function recovers the true parameter in only one iteration.

\begin{theorem}
\label{thm:population_tensor}
For any $k=1,\ldots, K$, if $\bOmega_j$ $(j\ne k)$ satisfies $\textrm{tr}(\bSigma_j^* \bOmega_j) \ne 0$, then the population minimization function in \eqref{eqn:population_Mk} satisfies
$
M_{k}(\bOmega_{[K]-k}) = m \bigl[m_k \prod_{j\ne k} \textrm{tr}(\bSigma_j^* \bOmega_j)\bigr]^{-1} \bOmega^*_k.
$
\end{theorem}

Theorem \ref{thm:population_tensor} indicates that the population minimization function recovers the true precision matrix up to a constant in only one iteration. If $\bOmega_j = \bOmega_j^*, j\ne k$, then $M_{k}(\bOmega_{[K]-k}) = \bOmega^*_k$. Otherwise, after a normalization such that $\|M_{k}(\bOmega_{[K]-k})\|_F = 1$, the normalized population minimization function still fully recovers  $\bOmega^*_k$. This observation suggests that setting $T=1$ in Algorithm \ref{alg:tlasso} is sufficient. Such a theoretical suggestion will be further supported by our numeric results. 

In practice, when the population log-likelihood function \eqref{eqn:population_q} is unknown, we can approximate it by its sample version $q_n(\bOmega_1, \ldots, \bOmega_K)$ defined in \eqref{eqn:sample_qn}, which gives rise to the statistical estimation error. Similar as \eqref{eqn:population_Mk}, we define the sample-based minimization function with parameter $\bOmega_{[K]-k} = \{\bOmega_1, \ldots, \bOmega_{k-1},\bOmega_{k+1},\ldots,\bOmega_K\}$ as
\begin{eqnarray} 
\hat{M}_k(\bOmega_{[K]-k}) :=  \argmin_{\bOmega_k} q_n(\bOmega_1, \ldots, \bOmega_K).
\label{eqn:sample_Mk}
\end{eqnarray}

In order to derive the estimation error, it remains to quantify the statistical error induced from finite samples. The following two regularity conditions are assumed for this purpose.

\begin{condition}[({Bounded Eigenvalues})]
\label{con:eigenvalue_tensor} 
For any $k=1,\ldots,K$, there is a constant $C_1>0$ such that, 
\begin{eqnarray*}
0 < C_1 \le \lambda_{\min}(\bSigma^*_k) \le \lambda_{\max}(\bSigma^*_k) \le 1/C_1 < \infty,
\end{eqnarray*}
where $\lambda_{\min}(\bSigma^*_k)$ and $\lambda_{\max}(\bSigma^*_k)$ refer to the minimal and maximal eigenvalue of $\bSigma^*_k$, respectively.
\end{condition}
Condition \ref{con:eigenvalue_tensor} has been commonly assumed in the precision matrix estimation literature in order to facilitate the proof of estimation consistency \cite{bickel2008,rothman2008,lam2009}.

\begin{condition}[({Tuning})]
\label{con:tuning_tensor} 
For any $k=1,\ldots,K$ and some constant $C_2>0$, the tuning parameter $\lambda_{k}$ satisfies $1/C_2 \sqrt{\log m_k / (n m m_k)} \le \lambda_{k} \le C_2 \sqrt{\log m_k/( n m m_k)}$.
\end{condition}
Before characterizing the statistical error, we define a sparsity parameter for $\bOmega^*_k$, $k=1,\ldots,K$. Let $\SSS_k := \{(i,j): [\bOmega^*_k]_{i,j} \ne 0\}$. Denote the sparsity parameter $s_k := |\SSS_k| - m_k$, which is the number of nonzero entries in the off-diagonal component of $\bOmega^*_k$. For each $k=1,\ldots,K$, we define $\BB(\bOmega_k^*)$ as the set containing $\bOmega_k^*$ and its neighborhood for some sufficiently large radius $\alpha>0$, i.e., $\BB(\bOmega_k^*) :=$
\begin{equation}
 \{ \bOmega \in \mathbb R^{m_k \times m_k}: \bOmega = \bOmega^{\top}; \bOmega \succ 0; \|\bOmega - \bOmega_k^*\|_F \le \alpha \}.\label{eqn:Bset}
\end{equation}

\begin{theorem}
\label{thm:statistical_error}
Suppose that Conditions \ref{con:eigenvalue_tensor} and \ref{con:tuning_tensor} hold. For any $k=1,\ldots, K$, the statistical error of the sample-based minimization function defined in \eqref{eqn:sample_Mk} satisfies that, for any fixed $\bOmega_j \in \BB(\bOmega_j^*)\ (j\ne k)$, 
\begin{align}  
& \bigl\| \hat{M}_k(\bOmega_{[K]-k}) - M_{k}(\bOmega_{[K]-k}) \bigr\|_F  =   O_P\left(\sqrt{ \frac{m_k(m_k+s_k)\log m_k }{nm} } \right),   \label{eqn:error_tensor}
\end{align}
where $M_{k}(\bOmega_{[K]-k})$ and $\hat{M}_k(\bOmega_{[K]-k})$ are defined in \eqref{eqn:population_Mk} and \eqref{eqn:sample_Mk}, and $m = \prod_{k=1}^K  m_k$.
\end{theorem}

Theorem \ref{thm:statistical_error} establishes the estimation error associated with $\hat{M}_k(\bOmega_{[K]-k})$ for arbitrary $\bOmega_j \in \BB(\bOmega_j^*)$ with $j\ne k$. In comparison, previous work on the existence of a local solution with desired statistical property only establishes theorems similar to Theorem \ref{thm:statistical_error} for $\bOmega_j = \bOmega_j^*$ with $j\ne k$. The extension to an arbitrary $\bOmega_j \in \BB(\bOmega_j^*)$ involves non-trivial technical barriers. Specifically, we first establish the rate of convergence of the difference between a sample-based quadratic form and its expectation (Lemma \ref{lemma:sample_cov}) via Talagrand's concentration inequality \cite{ledoux2011}. This result is also of independent interest. We then carefully characterize the rate of convergence of $\Sbb_k$ defined in \eqref{eqn:one_qn} (Lemma \ref{lemma:sample_cov_tensor}). Finally, we develop \eqref{eqn:error_tensor} using the results for vector-valued graphical models developed by \cite{fan2009}.

According to Theorem \ref{thm:population_tensor} and Theorem \ref{thm:statistical_error}, we obtain the rate of convergence of the Tlasso estimator in terms of Frobenius norm, which is our main result. 

\begin{theorem}
\label{thm:final_error}
Assume that Conditions \ref{con:eigenvalue_tensor} and \ref{con:tuning_tensor} hold. For any $k=1,\ldots,K$, if the initialization satisfies $\bOmega_j^{(0)} \in \BB(\bOmega_j^*)$ for any $j\ne k$, then the estimator $\widehat{\bOmega}_k$ from Algorithm \ref{alg:tlasso} with $T=1$ satisfies, 
\begin{equation}
\bigl\| \widehat{\bOmega}_k - \bOmega^*_k \bigr\|_F = O_P\Biggl(\sqrt{ \frac{m_k(m_k+s_k)\log m_k }{nm} } \Biggl),   \label{eqn:final_error_tensor}
\end{equation}
where $m = \prod_{k=1}^K  m_k$ and $\BB(\bOmega_j^*)$ is defined in \eqref{eqn:Bset}.
\end{theorem}

Theorem \ref{thm:final_error} suggests that as long as the initialization is within a constant distance to the truth, the Tlasso algorithm attains a consistent estimator after only one iteration. This consistency is insensitive to the initialization since the constant $\alpha$ in \eqref{eqn:Bset} can be arbitrarily large. In literature, \cite{he2014} show that there exists a local minimizer of \eqref{eqn:sample_qn} whose convergence rate can achieve \eqref{eqn:final_error_tensor}. However, it is unknown if their algorithm can find such a minimizer since there could be many other local minimizers. 

A notable implication of Theorem \ref{thm:final_error} is that, when $K\ge3$, the estimator from the Tlasso algorithm can achieve estimation consistency even if we only have access to one observation, i.e., $n=1$, which is often the case in practice. To see it, suppose that $K=3$ and $n=1$. When the dimensions $m_1,m_2$, and $m_3$ are of the same order of magnitude and $s_k = O(m_k)$ for $k=1,2,3$, all the three error rates corresponding to $k=1,2,3$ in \eqref{eqn:final_error_tensor} converge to zero.

Theorem \ref{thm:final_error} implies that the estimation of the $k$-th precision matrix takes advantage of the information from the $j$-th way ($j\ne k$) of the tensor data. Consider a simple case that $K=2$ and one precision matrix $\bOmega^*_1 = \ind_{m_1}$ is known. In this scenario the rows of the matrix data are independent and hence the effective sample size for estimating $\bOmega^*_2$ is in fact $nm_1$. The optimality result for the vector-valued graphical model \cite{cai2015} implies that the optimal rate for estimating $\bOmega^*_2$ is $\sqrt{(m_2+s_2)\log m_2/(nm_1)}$, which is consistent with our result in \eqref{eqn:final_error_tensor}. Therefore, the rate in \eqref{eqn:final_error_tensor} obtained by the Tlasso estimator is minimax-optimal since it is the best rate one can obtain even when $\bOmega^*_j$ ($j\ne k$) were known. As far as we know, this phenomenon has not been discovered by any previous work in tensor graphical model.

\begin{remark}
\label{remark:matrix}
For $K=2$, our tensor graphical model reduces to matrix graphical model with Kronecker product covariance structure \cite{yin2012, leng2012, tsiligkaridis2013, zhou2014}. In this case, the rate of convergence of $\widehat{\bOmega}_1$ in \eqref{eqn:final_error_tensor} reduces to $\sqrt{ (m_1+s_1)\log m_1/(nm_2)}$, which is much faster than $\sqrt{m_2 (m_1+s_1)(\log m_1 + \log m_2)/n}$ established by \cite{yin2012} and $\sqrt{ (m_1+m_2) \log[ \max(m_1, m_2, n)]/(nm_2)}$ established by \cite{tsiligkaridis2013}. In literature, \cite{leng2012} shows that there exists a local minimizer of the objective function whose estimation errors match ours. However, it is unknown if their estimator can achieve such convergence rate. On the other hand, our theorem confirms that our algorithm is able to find such estimator with an optimal rate of convergence.
\end{remark}

\subsection{Estimation Error in Max Norm and Spectral Norm}
\label{sec:error_max}

We next derive the estimation error in max norm and spectral norm. Trivially, these estimation errors are bounded by that in Frobenius norm shown in Theorem \ref{thm:final_error}. To develop improved rates of convergence in max and spectral norms, we need to impose stronger conditions on true parameters. 

We first introduce some important notations. Denote $d_k$ as the maximum number of non-zeros in any row of the true precision matrices $\bOmega^*_k$, that is,
\begin{equation}
d_k := \max_{i\in \{1,\ldots,m_k\}} \big|\{j\in \{1,\ldots,m_k\}:  [\bOmega^*_k]_{i,j} \ne 0 \} \big|, \label{eqn:dk}
\end{equation}
with $|\cdot|$ the set cardinality. For each covariance matrix $\bSigma_k^*$, we define $\kappa_{\bSigma_k^*} := \smallvertiii{ \bSigma_k^* }_{\infty}$. Denote the Hessian matrix $\bGamma_k^* := \bOmega_k^{* -1} \otimes  \bOmega_k^{* -1} \in \mathbb R^{m_k^2 \times m_k^2}$, whose entry $[\bGamma_k^*]_{(i,j),(s,t)}$ corresponds to the second order partial derivative of the objective function with respect to $[\bOmega_k]_{i,j}$ and $[\bOmega_k]_{s,t}$. We define its sub-matrix indexed by $\SSS_k$ as $[\bGamma_k^*]_{\SSS_k, \SSS_k} = [\bOmega_k^{* -1} \otimes  \bOmega_k^{* -1}]_{\SSS_k, \SSS_k}$, which is the $|\SSS_k| \times |\SSS_k|$ matrix with rows and columns of $\bGamma_k^*$ indexed by $\SSS_k$ and $\SSS_k$, respectively. Moreover, we define $\kappa_{\bGamma_k^*} := \vertiii{ ( [\bGamma_k^*]_{\SSS_k, \SSS_k} )^{-1} }_{\infty}$. In order to establish the rate of convergence in max norm, we need to impose an irrepresentability condition on the Hessian matrix.

\begin{condition}[({Irrepresentability})]
\label{con:IR} 
For each $k=1,\ldots,K$, there exists some $\alpha_k \in (0,1]$ such that 
$$
\max_{e \in \SSS_k^{c}} \big\| [\bGamma_k^*]_{e, \SSS_k} \big([\bGamma_k^*]_{\SSS_k,\SSS_k} \big)^{-1}   \big\|_1 \le 1 - \alpha_k.
$$
\end{condition}
Condition \ref{con:IR} controls the influence of the non-connected terms in $\SSS_k^{c}$ on the connected edges in $\SSS_k$. This condition has been widely applied for developing the theoretical properties of lasso-type estimator \cite{zhao2006, ravikumar2011,jankova2014}.

\begin{condition}[({Bounded Complexity})]
\label{con:bound_kappa} 
For each $k=1,\ldots,K$, the parameters $\kappa_{\bSigma_k^*}$ and $\kappa_{\bGamma_k^*}$ are bounded and the parameter $d_k$ in \eqref{eqn:dk} satisfies $d_k = o\big({\sqrt{nm}}/{(m_k \log m_k)} \big)$. 
\end{condition}

\begin{theorem}
\label{thm:maxnorm}
Suppose Conditions \ref{con:eigenvalue_tensor}, \ref{con:tuning_tensor}, \ref{con:IR} and \ref{con:bound_kappa} hold. Assume $s_k = O(m_k)$ for $k=1,\ldots, K$ and assume $m_k's$ are in the same order, i.e., $m_1 \asymp m_2 \asymp \cdots \asymp m_K$. For each $k$, if the initialization satisfies $\bOmega_j^{(0)} \in \BB(\bOmega_j^*)$ for any $j\ne k$, then the estimator $\widehat{\bOmega}_k$ from Algorithm \ref{alg:tlasso} with $T=2$ satisfies, 
\begin{equation}
\big\| \widehat{\bOmega}_k -  \bOmega_k^* \big\|_{\infty} = O_P\left(\sqrt{\frac{m_k \log m_k}{n m }}\right). \label{eqn:maxnorm}
\end{equation}
In addition, the edge set of $\widehat{\bOmega}_k$ is a subset of the true edge set of $\bOmega_k^*$, that is, $\textrm{supp}(\widehat{\bOmega}_k) \subseteq \textrm{supp}(\bOmega_k^*)$.
\end{theorem}

Theorem \ref{thm:maxnorm} shows that the Tlasso estimator achieves the optimal rate of convergence in max norm \cite{cai2015}. Here we consider the estimator obtained after two iterations since we require a new concentration inequality (Lemma \ref{lemma:sample_cov_max}) for the sample covariance matrix, which is built upon the estimator in Theorem \ref{thm:final_error}. 

\begin{remark}
Theorem \ref{thm:maxnorm} ensures that the estimated precision matrix correctly excludes all non-informative edges and includes all the true edges $(i,j)$ with $| [\bOmega_k^*]_{i,j} | > C \sqrt{{m_k \log m_k}/{(n m) }} $ for some constant $C>0$. Therefore, in order to achieve the variable selection consistency $\textrm{sign}\bigl( \widehat{\bOmega}_k \bigr) = \textrm{sign}(\bOmega_k^*)$, a sufficient condition is to assume that the minimal signal $\min_{(i,j)\in \textrm{supp}(\bOmega^*_k)}| [\bOmega_k^*]_{i,j}| \ge C\sqrt{{m_k \log m_k}/({n m })}$ for each $k$. This confirms that the Tlasso estimator is able to correctly recover the graphical structure of each way of the high-dimensional tensor data. 
\end{remark}





A direct consequence from Theorem \ref{thm:maxnorm} is the estimation error in spectral norm.

\begin{corollary}
\label{thm:spectral_norm}
Suppose the conditions of Theorem \ref{thm:maxnorm} hold, for any $k=1,\ldots,K$, we have 
\begin{equation}
\big\| \widehat{\bOmega}_k -  \bOmega_k^* \big\|_2 = O_P\left(d_k\sqrt{\frac{m_k \log m_k}{n m }}\right). \label{eqn:spectral_norm}
\end{equation}
\end{corollary}

\begin{remark}
\label{remark:spectral}
Now we compare our obtained rate of convergence in spectral norm for $K=2$ with that established in the sparse matrix graphical model literature. In particular, \cite{zhou2014} establishes the rate of $O_P\big(\sqrt{m_k (s_k \vee 1) \log (m_1 \vee  m_2)/(n m_k)}\big)$ for $k=1,2$. Therefore, when $d_k^2 \le (s_k \vee 1)$, which holds for example in the bounded degree graphs, our obtained rate is faster. However, our faster rate comes at the price of assuming the irrepresentability condition. Using recent advance in non-convex regularization \cite{loh2014support}, we can actually eliminate the irrepresentability condition. We leave this to future work. 
\end{remark}

\section{Tensor Inference}
\label{sec:tensor_inference}

This section introduces a statistical inference procedure for sparse tensor graphical models. In particular, built on Tlasso algorithm, a consistent test statistic is constructed for hypothesis
\begin{equation} \label{eqn: H0}
H_{0 k , i j } : \;  [\bOmega_k^*]_{i,j} =0  \quad\quad \textrm{v.s.} \quad\quad H_{1k , ij} : \;  [\bOmega_k^*]_{i,j} \ne  0,
\end{equation}
$\forall  1 \le i < j \le m_k$ and $k=1,\ldots,K$. 
Also, to simultaneously test all off-diagonal entries, a multiple testing procedure is developed with false discovery rate (FDR) control.

\subsection{Construction of Test Statistic}  \label{sec: construct}

Without loss of generality, we focus on testing $\bOmega_1^*$. For a tensor ${\cal T} \in \mathbb R^{m_1\times \cdots \times m_K}$, denote ${\cal T}_{-i_1 , i_2 , \ldots , i_K}  \in \mathbb R^{m_1-1}$ as the vector by removing the $i_1$-th entry of ${\cal T}_{:, i_2 , \ldots , i_K}$. Given that ${\cal T}$ follows a tensor normal distribution $(\ref{eqn:pdf})$, we have, $\forall i_1 \in m_1$, ${\cal T}_{i_1 , i_2 , \ldots , i_K}  |  {\cal T}_{-i_1 , i_2 , \ldots , i_K} \sim $
\begin{equation}
\textrm{N} \left( - [\bOmega_1^*]_{i_1,i_1}^{-1}  [\bOmega_1^*]_{i_1, - i_1}  {\cal T}_{-i_1 , i_2 , \ldots , i_K}  ;  [\bOmega_1^*]_{i_1, i_1 }^{-1} \prod_{k=2}^K[\bSigma_k^*]_{i_k, i_k} \right).
\label{eqn:partial_normal}
\end{equation}
Inspired by \eqref{eqn:partial_normal}, our tensor graphical model can be reformulated into a linear regression. Specifically, for tensor sample ${\cal T}_l$, $l=1,\ldots,n$, $(\ref{eqn:partial_normal})$ implies that,  
\begin{equation}
{\cal T }_{l ; i_1, i_2 , \ldots , i_K} = {\cal T }_{l ; - i_1, i_2 , \ldots , i_K}^\top \btheta_{i_1} + \xi_{l ; i_1, i_2 , \ldots , i_K},   \label{eqn: linear}
\end{equation}
where regression parameter $\btheta_{i_1} = - [\bOmega_1^*]_{i_1,i_1}^{-1}  [\bOmega_1^*]_{i_1, - i_1}$, and noise 
\begin{equation} \label{eq: noise_var}
\xi_{l ;i_1, i_2 , \ldots , i_K} \sim \textrm{N} (0 \, ; [\bOmega_1^*]_{i_1, i_1 }^{-1} \prod_{k=2}^K [\bSigma_k^*]_{i_k, i_k}). 
\end{equation}

Let $\hat{\bOmega}_1$ be an estimate of ${\bOmega}_1$ obtained from Tlasso algorithm. Naturally, a plug-in estimate of ${\btheta}_{i_1}$ follows, i.e., $\hat{\btheta}_{i_1} = (\hat{\theta}_{1, i_1} , \ldots , \hat{\theta}_{m_1-1, i_1})^\top  :=  - [\hat{\bOmega}_1]_{i_1,i_1}^{-1}  [\hat{\bOmega}_1]_{i_1, - i_1}$.
Denote a residual of \eqref{eqn: linear} as $\hat{\xi}_{l ; i_1, i_2 , \ldots , i_K} :=$
$$
 {\cal T }_{l ; i_1, i_2 , \ldots , i_K} - \bar{{\cal T }}_{ i_1, i_2 , \ldots , i_K} - ( {\cal T }_{l ; - i_1, i_2 , \ldots , i_K} - \bar{{\cal T }}_{ - i_1, i_2 , \ldots , i_K} )^\top \hat{\btheta}_{i_1}, 
$$
where $\bar{{\cal T }} = \sum_{l=1}^{n} {\cal T }_{l }/n$. Correspondingly, its sample covariance is, $\forall 1 \le i < j \le m_1$, $\hat{\varrho}_{i,j}=$
$$
 \frac{m_1}{(n-1) m } \sum \limits_{l=1}^n \sum \limits_{i_2=1}^{m_2} \cdots \sum \limits_{i_K=1}^{m_K}  \hat{\xi}_{l ; i, i_2 , \ldots , i_K} \hat{\xi}_{l ; j, i_2 , \ldots , i_K} .
$$

In light of \eqref{eq: noise_var}, information of $[\bOmega_1^*]_{i,j}$ is encoded in $\hat{\varrho}_{i,j}$.
In this sense, a test statistic is proposed, i.e.,
\begin{equation}
\tau_{i,j} = \frac{\hat{\varrho}_{i,j} + \mu_{i,j}}{\varpi}, \forall 1 \le i < j \le m_1. 
\label{eqn: test_statistic}
\end{equation}
Intuition of $\tau_{i,j}$ is extracting knowledge of $[\bOmega_1^*]_{i,j}$ from $\hat{\varrho}_{i,j}$ via two-step correction. 
Notably, bias correction term $\mu_{i,j}:=\hat{\varrho}_{i,i} \hat{\theta}_{i , j} + \hat{\varrho}_{j,j} \hat{\theta}_{j-1, i}$ reduces bias resulting from estimation error of $\hat{\btheta}_{i}$ and $\hat{\btheta}_{j}$. 
In addition, variance correction term 
\begin{equation}
\varpi^2 :=  \frac{m \cdot \|\widehat{\Sbb}_2\|_F^2 \cdots \|\widehat{\Sbb}_K\|_F^2}{m_1 \cdot [\tr(\widehat{\Sbb}_2)]^2 \cdots [\tr(\widehat{\Sbb}_K)]^2},  \label{eqn: var_corr}
\end{equation}
eliminates extra variation introduced by the rest $K-1$ modes (see \eqref{eq: noise_var}). Here $\widehat{\Sbb}_k := \frac{m_k}{nm} \sum_{i=1}^n \widehat{\Vb}_i \widehat{\Vb}_i^{\top}$ is an estimate of $\bSigma_k$, where $\widehat{\Vb}_i := \big[ {\cal T}_i \times \bigl\{\widehat{\bOmega}_1^{1/2},\ldots,\widehat{\bOmega}_{k-1}^{1/2}, \ind_{m_k}, \widehat{\bOmega}_{k+1}^{1/2},\ldots,\widehat{\bOmega}_{K}^{1/2} \bigr\} \big]_{(k)}$ with $\widehat{\bOmega}_{k}$ from Tlasso algorithm. 

Theorem \ref{thm: test_consistency} establishes asymptotic normality of $\tau_{i,j}$. Symmetrically, such normality can be extended to the rest $K-1$ modes. 

\begin{theorem} \label{thm: test_consistency} 
Assume the same assumptions of Theorem \ref{thm:maxnorm}, we have, under null $(\ref{eqn: H0})$,
$$\tilde{\tau}_{i,j}:= \sqrt{\frac{(n-1) m  }{ m_1 \hat{\varrho}_{i,i} \hat{\varrho}_{j,j} }} \tau_{i,j}  \rightarrow \textrm{N} ( 0 ;1 ) $$
in distribution, as $nm/m_1 \rightarrow \infty$. 
\end{theorem}

Theorem \ref{thm: test_consistency} implies that, when $K \ge 2$, asymptotic normality holds even if we have a constant number of observations, which is often the case in practice. For example, let $n=2$ and $m_1 \asymp m_2 $, $nm/m_1$ still goes to infinity as $m_1, m_2 $ diverges . This result reflects an interesting phenomenon specifically in tensor graphical models. Particularly, hypothesis testing for certain mode's precision matrix could take advantage of information from the rest modes in tensor data. As far as we know, this phenomenon has not been discovered by any previous work in tensor graphical models.


\subsection{FDR Control Procedure} \label{sec: FDR_control}
Though our test statistic enjoys consistency on single entry, simultaneously testing all off-diagonal entries is more of practical interest. 
Thus, in this subsection, a multiple testing procedure with false discovery rate (FDR) control is developed.

Given a thresholding level $\varsigma$, denote $\varphi_{\varsigma} (\tilde{\tau}_{i,j}) := \ind \{ |\tilde{\tau}_{i,j}| \ge \varsigma \}$.  
Null is rejected if $\varphi_{\varsigma} (\tilde{\tau}_{i,j})=1$. 
Correspondingly, false discovery proportion (FDP) and FDR are defined as
\begin{equation*}
\textrm{FDP} = \frac{| \{ (i,j) \in {\cal H}_0 : \varphi_{\varsigma} (\tilde{\tau}_{i,j})=1 \} | }{ | \{ (i,j)  : 1 \le i < j \le m_1, \varphi_{\varsigma} (\tilde{\tau}_{i,j})=1 \} |  \vee 1 } , \label{eqn: FDP}
\end{equation*}
and $\textrm{FDR} = \EE (\textrm{FDP})$.
Here ${\cal H}_0 = \{ (i,j) : [\bOmega_1^*]_{i,j} = 0 , 1 \le i < j \le m_1 \}$. 
A sufficient small $\varsigma$ is ideal that significantly enhances power, meanwhile controls FDP under a pre-specific level $\upsilon \in (0,1)$. 
In particular, the ideal thresholding value is 
$$
\varsigma_{*} := \inf \{   \varsigma > 0: \text{FDP} \le \upsilon  \}.
$$ 

However, in practice, $\varsigma_{*}$ is not attainable due to unknown ${\cal H}_0$ in \textrm{FDP}. 
Therefore, we approximate $\varsigma_{*}$ by the following heuristics. 
Firstly, Theorem \ref{thm: test_consistency} implies that $P(\varphi_{\varsigma} (\tilde{\tau}_{i,j})=1)$ is close to $2(1 - \Phi( \varsigma)) $ asymptotically. 
So the numerator of \textrm{FDP} is approximately $2(1 - \Phi( \varsigma)) |{\cal H}_0|$. 
Secondly, sparsity indicates that most entries are zero. Consequently, $|{\cal H}_0|$ is nearly $w:= m_1(m_1 -1 )/2 $. 
Under the above concerns, an approximation of $\varsigma_*$ is $\hat{\varsigma}=$
\begin{equation}
 \inf \bigg \{   \varsigma > 0: \frac{2(1-\Phi( \varsigma )) w}{  | \{ (i,j)  :   i < j  ,  \varphi_{\varsigma} (\tilde{\tau}_{i,j})=1 \} | \vee 1 } \le \upsilon  \bigg \}, \label{eqn: hatu}
\end{equation}
which is a trivial one-dimensional search problem. 

\begin{algorithm}
\caption{Support recovery with FDR control for sparse tensor graphical models}
\begin{algorithmic}[1]
\STATE \textbf{Input:} Tensor samples ${\cal T}_1\ldots, {\cal T}_n$, $\{\widehat{\bOmega}_k\}_{k=1}^{K}$ from Algorithm \ref{alg:tlasso}, and a pre-specific level $\upsilon$.
\STATE \textbf{Initialize:} Support ${\cal S}= \emptyset$.
\STATE Compute test statistic $\tilde{\tau}_{i,j}$, $\forall 1 \le i < j \le m_1$, defined in Theorem \ref{thm: test_consistency}.
\STATE Compute thresholding level $\hat{\varsigma}$ in \eqref{eqn: hatu}.
\STATE If $\tilde{\tau}_{i,j} > \hat{\varsigma}$, $\forall 1 \le i < j \le m_1$, reject null hypothesis and set ${\cal S} = {\cal S } \cup \{(i,j) , (j , i)\}$.
\STATE \textbf{Output:} ${\cal S}  \cup \{(i,i): 1 \le i \le m_1 \}$.
\end{algorithmic}\label{alg: FDP}
\end{algorithm}

Algorithm \ref{alg: FDP} describes our multiple testing procedure with FDR control for support recovery of $\bOmega_1^*$. 
Extension to the rest $K-1$ modes is symmetric. 
Clearly, FDR and FDP for $\bOmega_1^*$ from Algorithm \ref{alg: FDP} are  
$$
\textrm{FDP}_1 = \frac{| \{ (i,j) \in {\cal H}_0 : \varphi_{\hat{\varsigma}} (\tilde{\tau}_{i,j})=1 \} | }{ | \{ (i,j)  : 1 \le i < j \le m_1, \varphi_{\hat{\varsigma}} (\tilde{\tau}_{i,j})=1 \} | \vee 1 },
$$
and $\textrm{FDR}_1 = \EE (\textrm{FDP}_1)$. 
To depict their asymptotic behavior, two additional conditions are imposed related to size of true alternatives and sparsity.
\begin{condition}[({Alternative Size})] 
\label{con: true_alter}
Denote $\varpi_0^2 =  m \cdot \|\bSigma_2^*\|_F^2 \cdots \|\bSigma_K^*\|_F^2 / (m_1 \cdot (\tr(\bSigma_2^*)  \cdots  \tr(\bSigma_K^*))^2)$. 
It holds that $\big | \big\{  (i,j): 1 \le i < j \le m_1, |[\bOmega_1^*]_{i,j}| / \sqrt{[\bOmega_1^*]_{i,i}[\bOmega_1^*]_{j,j}}  \ge 4 \sqrt{\varpi_0 m_1 \log m_1/ ((n-1)m)}  \big\}  \big | \ge  \sqrt{\log \log m_1}.$
\end{condition}
\begin{condition}[({Sparsity})]
\label{con: spar}
For some $\rho < 1/2 $ and $\gamma > 0$, there exists a positive constant $C$ such that 
$\max \limits_{1 \le i \le m_1} \big |  \big\{  j : 1\le j \le m_1, j \ne i, |[\bOmega_1^*]_{i,j}| \ge (\log m_1)^{-2-\gamma} \big\} \big | \le C m_1^\rho .$
\end{condition}
Notably, Condition \ref{con: true_alter} and \ref{con: spar} imply an interesting interplay between sparsity and number of true alternatives. 
In addition, Condition \ref{con: true_alter} is nearly necessary in the sense that FDR control for large-scale multiple testing fails if number of true alternatives is fixed \cite{LS14}. 
Also, if $|{\cal H}_0| = o(w)$ (Condition \ref{con: spar} fails), most hypotheses would be rejected, and $\textrm{FDP}_1 \rightarrow 0$. 
Thus FDR control makes no sense anymore. 


Theorem \ref{thm: FDR_control} characterizes asymptotic properties of $\text{FDP}_1$ and $\text{FDR}_1$. 
For simplicity, we denote $w_0=|{\cal H}_0|$.
\begin{theorem} \label{thm: FDR_control}
Assume the same assumptions of Theorem \ref{thm: test_consistency}, together with Condition \ref{con: true_alter} \& \ref{con: spar}. 
If
$m_1 \le (nm/m_1)^r$ and $w_0 \ge c w$ for some positive constants $r $ and $c $, we have
$$
\textrm{FDP}_1 w / \upsilon w_0 \rightarrow 1 ,  \; \text{and} \; \;  \; \textrm{FDR}_1  w / \upsilon w_0 \rightarrow 1 
$$
in probability as $nm/m_1 \rightarrow \infty$.
\end{theorem} 

Theorem \ref{thm: FDR_control} shows that our FDR control procedure is still valid even when sample size is constant and dimensionality diverges. Similar to Theorem \ref{thm: test_consistency}, this phenomenon is specific to tensor graphical models.


\begin{remark} \label{rem: kro_prod_control}
Theorem \ref{thm: FDR_control} can be utilized to control FDR and FDP of testing Kronecker product $\bOmega_1^* \otimes \cdots \otimes \bOmega_K^*$. Consider a simple example with $K=3$, denote $f_1, f_2, f_3$ as numbers of false discoveries of testing $\bOmega_1^*, \bOmega_2^*, \bOmega_3^*$ respectively, and $d_1, d_2, d_3$ as numbers of corresponding off-diagonal discoveries. 
FDP and FDR of testing $\bOmega_1^* \otimes \bOmega_2^* \otimes \bOmega_3^*$ are 
$$
\textrm{FDP}_{c}= \frac{\alpha_0 (m_3 + d_3)  + (\alpha -\alpha_0 +m_1m_2) f_3 }{ [\prod_{k=1}^3 (d_k + m_k) - m_1m_2m_3 ] \vee 1}, 
$$ 
and $\textrm{FDR}_c = \EE (\textrm{FDP}_c)$, where $\alpha_0 = f_1 (m_2 + d_2) + (d_1 - f_1 + m_1 )f_2$ and $\alpha = (d_1 + m_1 ) (d_2 + m_2 ) - m_1 m_2 $. In practice, values of $f_k$, $k\in \{1,2,3\}$, can be estimated by $\upsilon d_k$ by Theorem \ref{thm: FDR_control}, given that all precision matrices are sparse enough. Therefore, define
$$
\tau=\frac{\alpha_0' (m_3 + d_3)  + (\alpha -\alpha_0' +m_1m_2) \upsilon d_3 }{[ \prod_{k=1}^3 (d_k + m_k) - m_1m_2m_3  ] \vee 1},
$$
where $\alpha_0 '= \upsilon d_1 (m_2 + 2 d_2) + ( m_1 - \upsilon d_1 ) \upsilon d_2$. Similar arguments of Theorem \ref{thm: FDR_control} imply that $\textrm{FDP}_c / \tau\rightarrow 1$ and $\textrm{FDR}_c/\tau \rightarrow 1$.
\end{remark}

\section{Simulations}
\label{sec:simulation}

In this section, we demonstrate superior empirical performance of proposed estimation and inference procedures for sparse tensor graphical models. These procedures are implemented into R package {\bf Tlasso}.

At first, we present numerical study of the Tlasso algorithm with iteration $T=1$ and compare it with two alternative approaches. The first alternative method is graphical lasso (Glasso) approach \cite{friedman2008} that applies to vectorized tensor data. This method ignores tensor structure of observed samples, and estimates Kronecker product of precision matrices $\bOmega_1^*\otimes \cdots \otimes \bOmega_K^*$ directly. The second alternative method is iterative penalized maximum likelihood method (P-MLE) proposed by \cite{he2014}. This method iteratively updates each precision matrix by solving an individual graphical lasso problem while fixing all other precision matrices until a pre-specified termination condition $\sum_{k=1}^K\| \widehat{\bOmega}_k^{(t)} - \widehat{\bOmega}_k^{(t-1)} \|_F/K \le 0.001$ is met. 

In the Tlasso algorithm, the tuning parameter for updating $\widehat{\bOmega}_k$ is set in the form of $C \sqrt{{\log m_k}/(n m m_k)}$ as assumed in Condition \ref{con:tuning_tensor}. 
Throughout all the simulations and real data analysis, we set $C=20$. Sensitivity analysis in \S\ref{sec:sensitivity} of the online supplement shows that the performance of Tlasso is relatively robust to the value of $C$. 
For a fair comparison, the same tuning parameter is applied in P-MLE method for $k=1,\ldots,K$. 
Individual graphical lasso problems in both Tlasso and P-MLE method are computed via $\it{huge}$. 
In the direct Glasso approach, its single tuning parameter is chosen by cross-validation automatically via $\it{huge}$.

In order to measure estimation accuracy of each method, three error criteria are selected. The first one is Frobenius estimation error of Kronecker product of precision matrices, i.e., 
\begin{equation} \label{eq:kro_F}
\frac{1}{m}\bigl\| \widehat{\bOmega}_1 \otimes \cdots \otimes   \widehat{\bOmega}_K - \bOmega^*_1 \otimes \cdots \otimes \bOmega^*_K \bigr\|_F,
\end{equation}
and the rest two are averaged estimation errors in Frobenius norm and max norm, i.e.,
\begin{equation} \label{eq:ave_measure}
 \frac{1}{K} \sum_{k=1}^K  \big\| \widehat{\bOmega}_k - \bOmega^*_k \big\|_F, \ \ \text{and}  \ \ \ \ \frac{1}{K} \sum_{k=1}^K \big\| \widehat{\bOmega}_k - \bOmega^*_k \big\|_{\infty}.
\end{equation}
Note that the last two criteria are only available to P-MLE method and Tlasso. 

%
%

\begin{figure}[h!]
\centering
\includegraphics[scale=0.3]{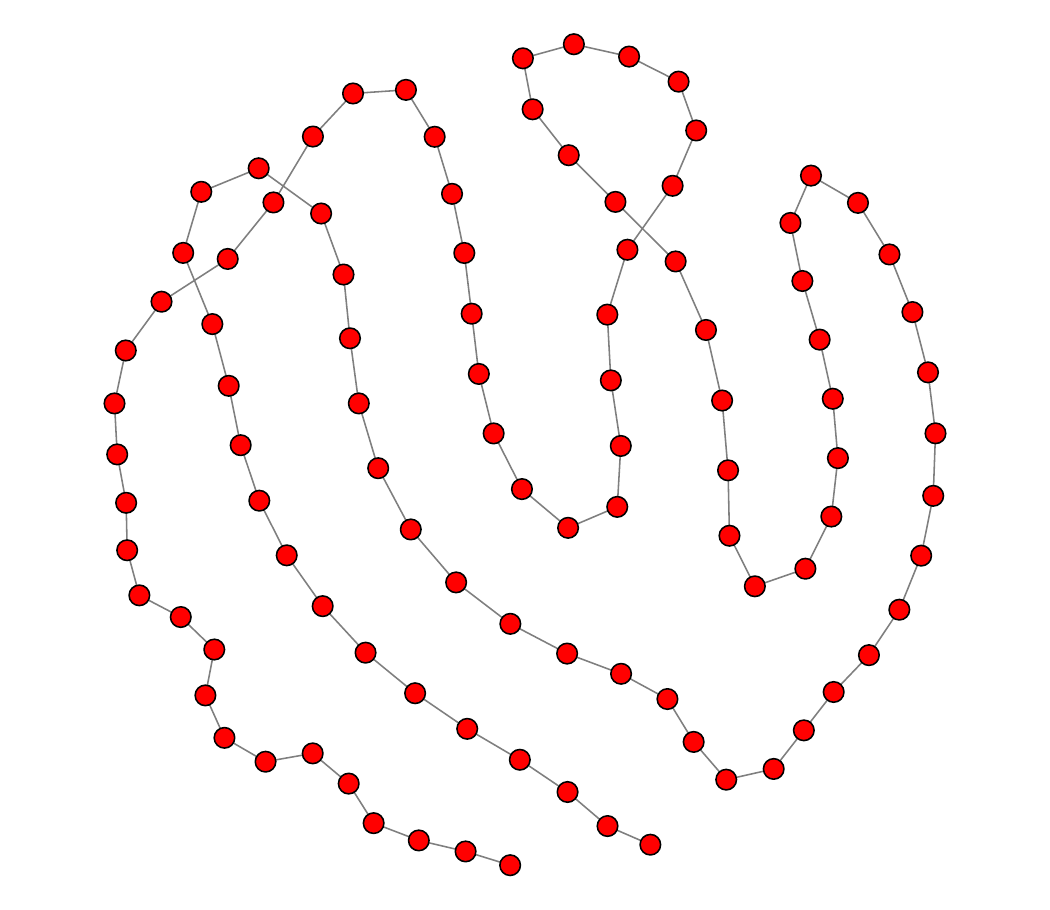}
\includegraphics[scale=0.3]{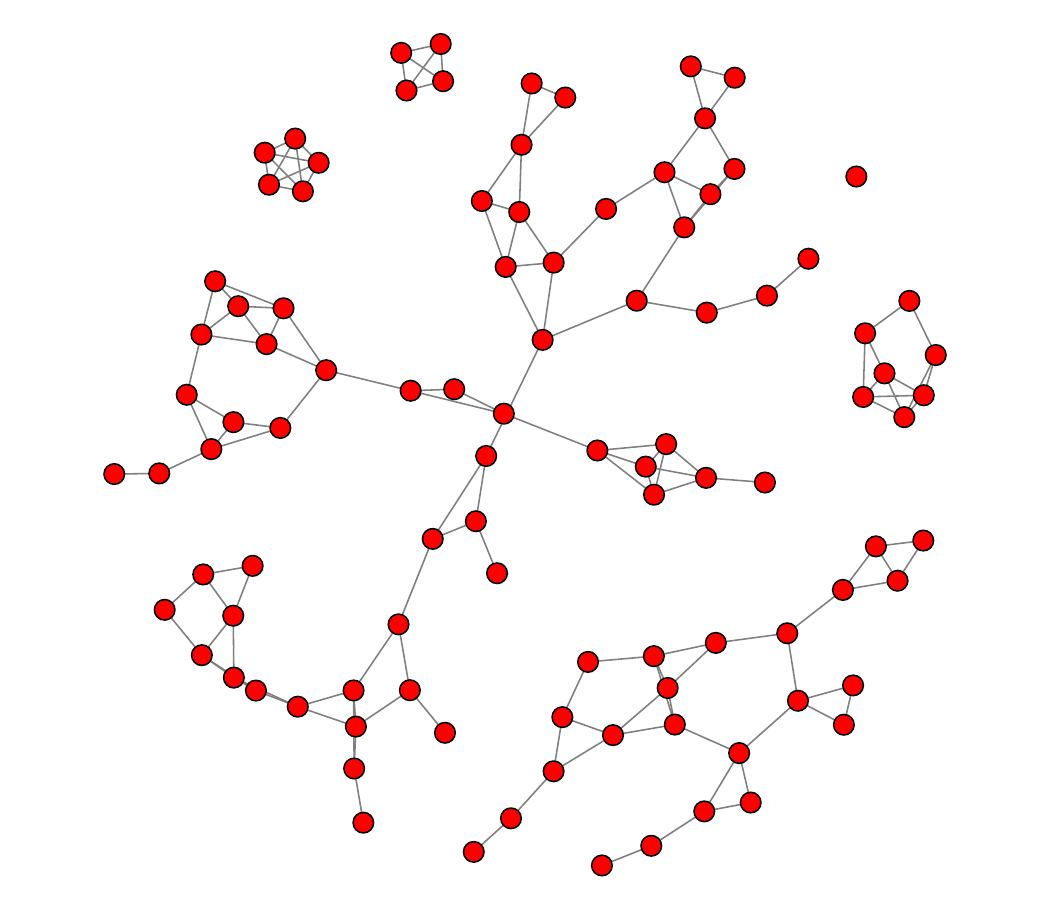}
\caption{An illustration of generated triangle graph (left) in Simulations $1$ and four nearest neighbor graph (right) in Simulations $2$. In this illustration, the dimension is $100$.}
\label{fig:data}
\vskip 0em
\end{figure}

Two simulations are considered for a third order tensor, i.e., $K=3$. In Simulation 1, we construct a triangle graph; in Simulation 2, a four nearest neighbor graph is adopted for each precision matrix. An illustration of generated graphs are shown in Figure \ref{fig:data}. Detailed generation procedures for the two graphs are as follows.

{\bf Triangle:} For each $k=1,\ldots,K$, we construct covariance matrix $\bSigma_k \in \mathbb R^{m_k \times m_k}$ such that its $(i,j)$-th entry is $[\bSigma_k]_{i,j} = \exp(- | h_i - h_j |/2)$ with $h_1 < h_2 < \cdots < h_{m_k}$. The difference $h_i - h_{i-1}$, $i=2,\ldots,m_k$, is generated i.i.d. from $\textrm{Unif}(0.5,1)$. This generated covariance matrix mimics autoregressive process of order one, i.e., $\textrm{AR}(1)$. We set $\bOmega^*_k = \bSigma_k^{-1}$. Similar procedure has also been used by \cite{fan2009}.

{\bf Nearest Neighbor:} For each $k=1,\ldots,K$, we construct precision matrix $\bOmega_k \in \mathbb R^{m_k \times m_k}$ directly from a four nearest-neighbor network. Firstly, $m_k$ points are randomly picked from an unit square and all pairwise distances among them are computed. We then search for the four nearest-neighbors of each point and a pair of symmetric entries in $\bOmega_k$ corresponding to a pair of neighbors that has a randomly chosen value from $[-1,-0.5] \cup [0.5,1]$. To ensure its positive definite property, the final precision matrix is designed as $\bOmega_k^* = \bOmega_k + (|\lambda_{\min}(\bOmega_k) + 0.2| \cdot \ind_{m_k})$, where $\lambda_{\min(\cdot)}$ refers to the smallest eigenvalue. Similar procedure has also been studied by \cite{lee2015}.

In each simulation, we consider three scenarios as follows. Each scenario is repeated 100 times. Averaged computational time, and averaged criteria for estimation accuracy and variable selection consistency are computed.

\begin{itemize}
\item \textbf{Scenario s1:} sample size $n = 50$ and dimension $(m_1,m_2,m_3) = (10,10,10)$.
\item \textbf{Scenario s2:} sample size $n = 80$ and dimension $(m_1,m_2,m_3) = (10,10,10)$.
\item \textbf{Scenario s3:} sample size $n = 50$ and dimension $(m_1,m_2,m_3) = (10,10,20)$.
\end{itemize}

\begin{figure}[h!]
\centering
\includegraphics[scale=0.3]{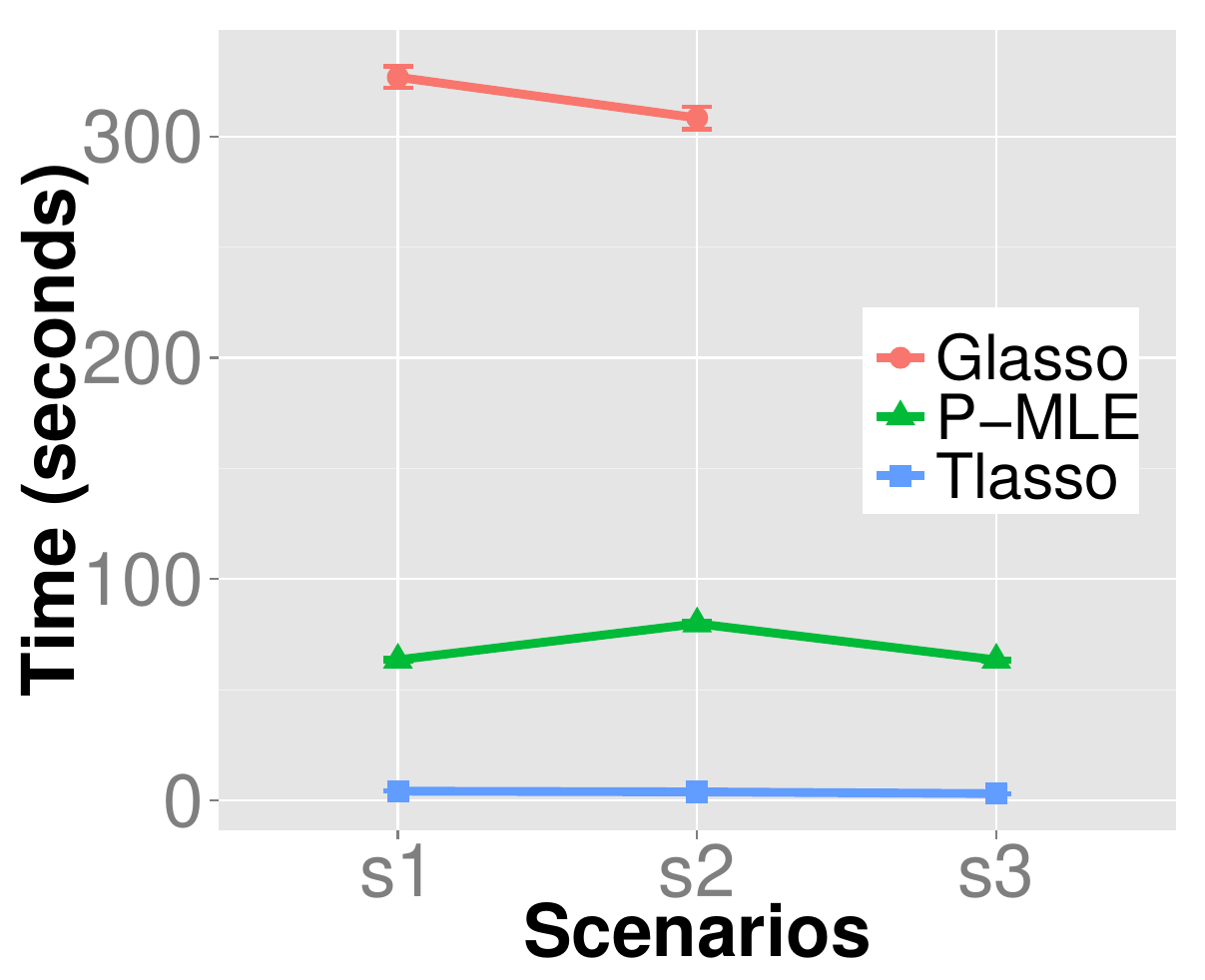}
\includegraphics[scale=0.3]{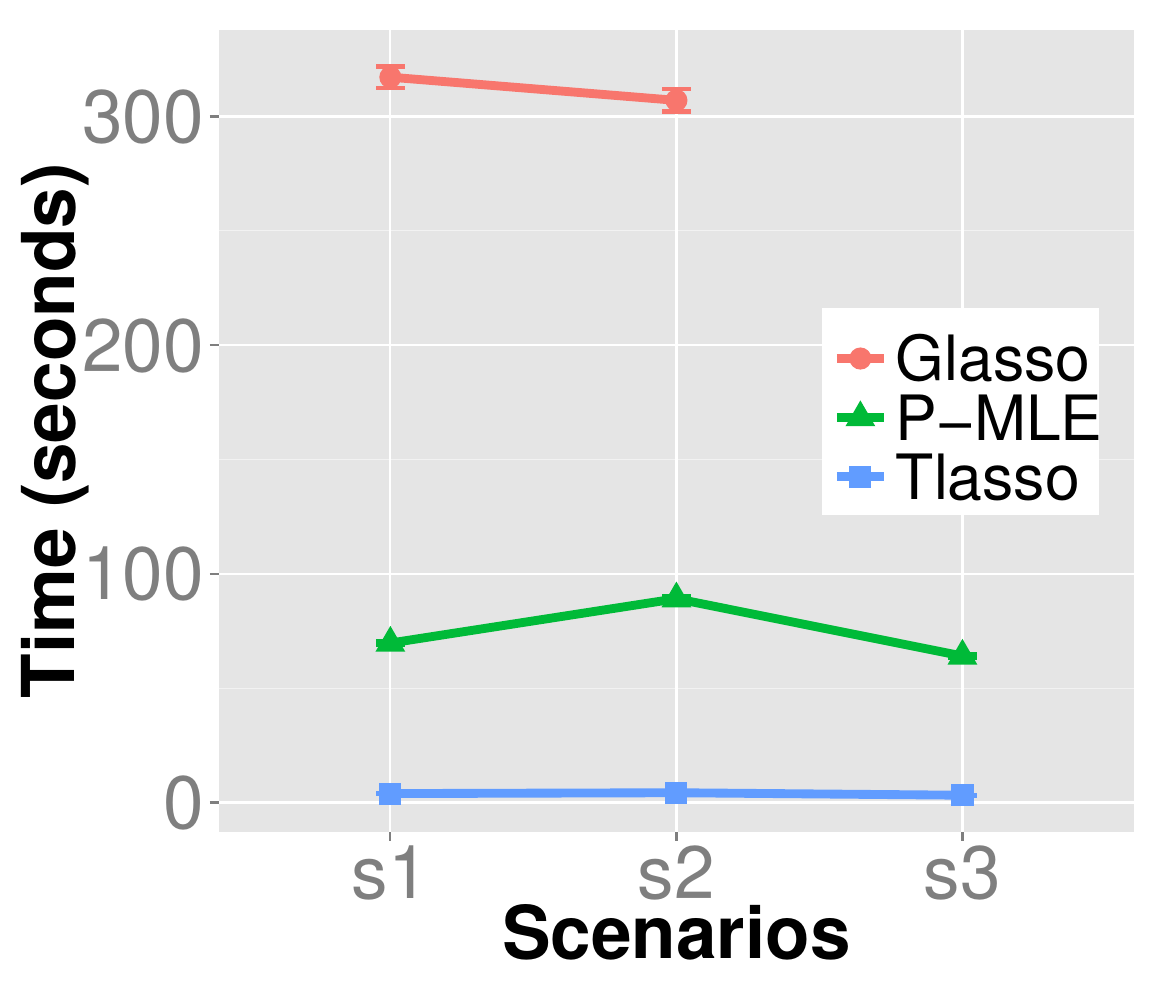}
\includegraphics[scale=0.3]{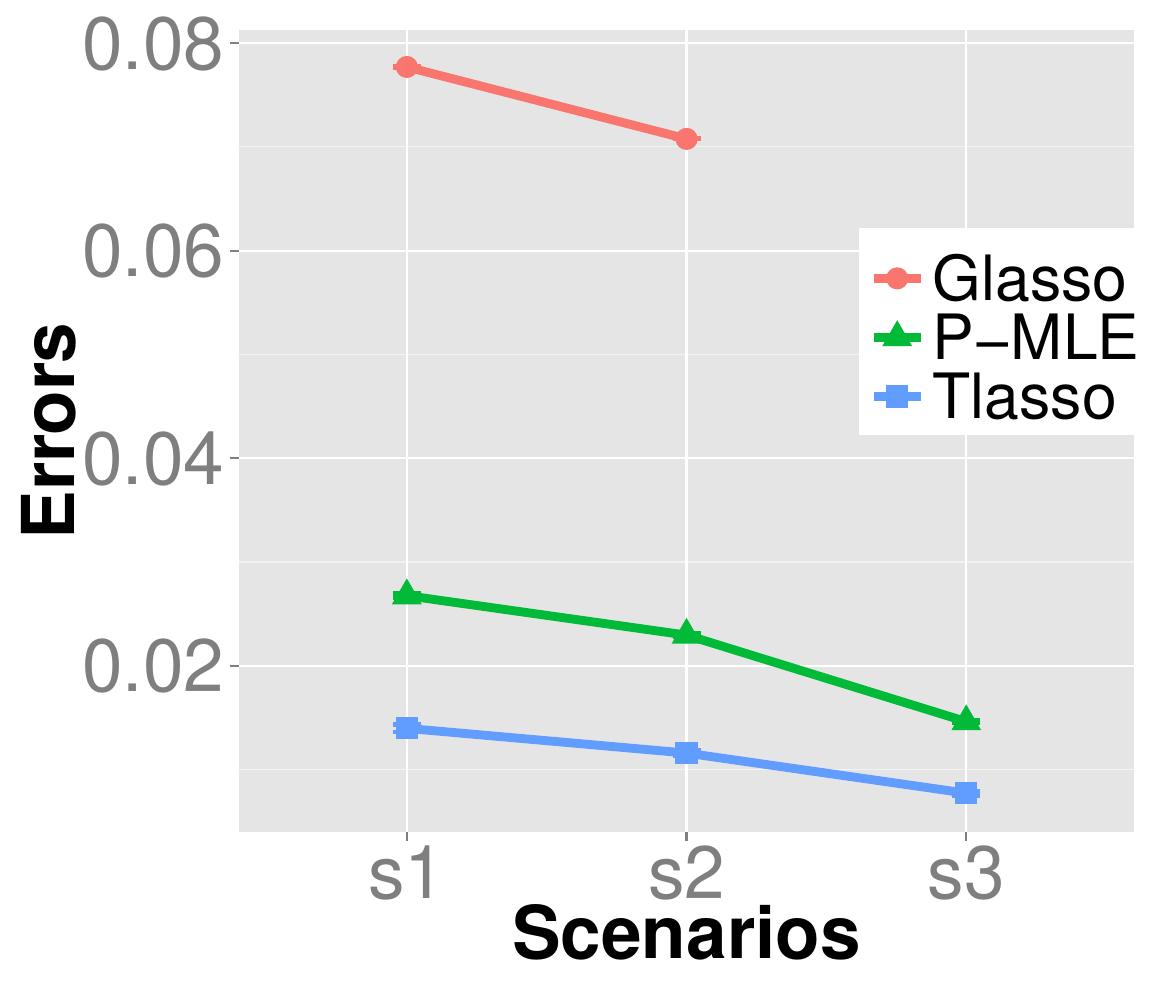}
\includegraphics[scale=0.3]{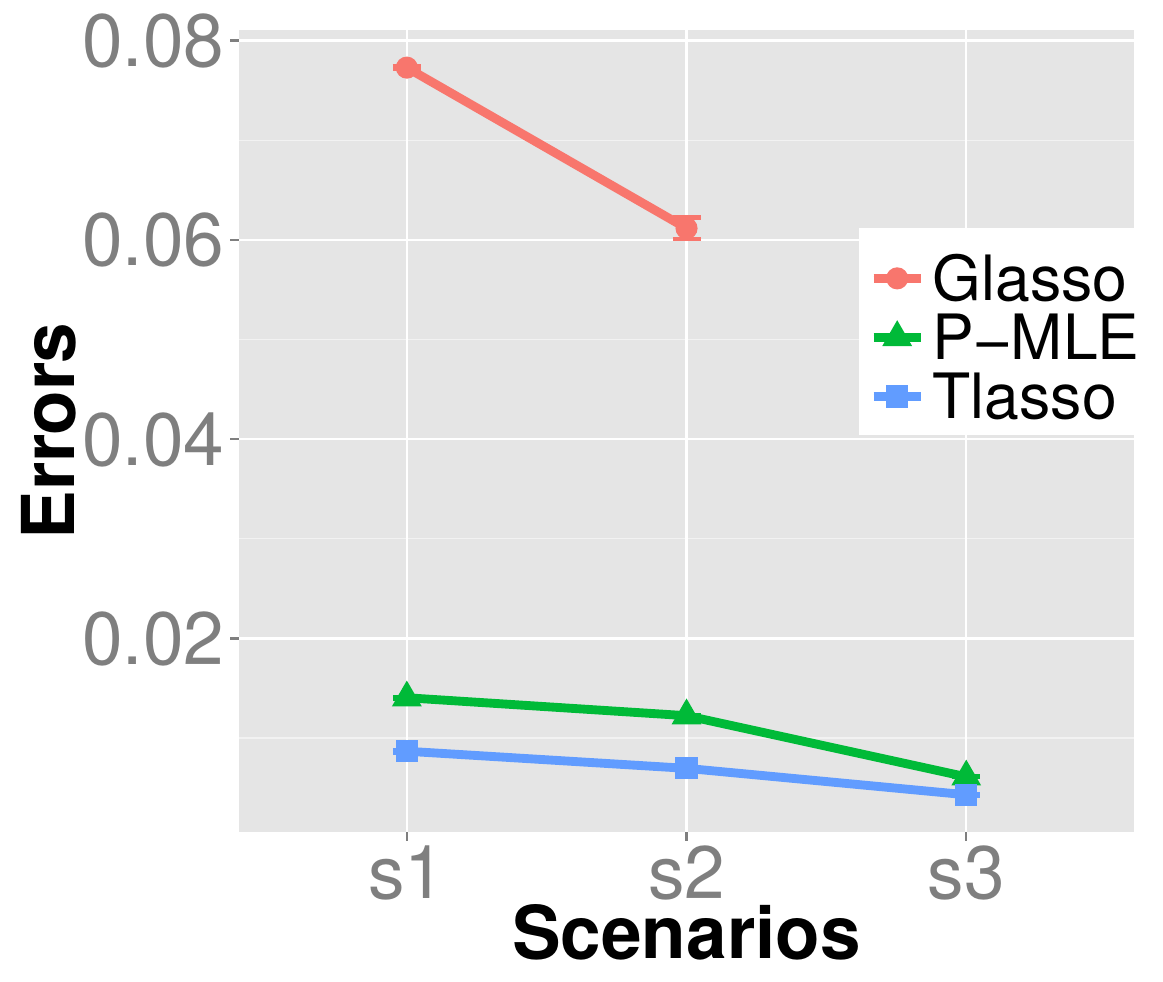}
\caption{The first row: averaged computational time of each method in Simulations $1\&2$, respectively. The second row: averaged estimation error of  Kronecker product of precision matrices of each method in Simulations $1\&2$, respectively. Results for the direct Glasso method in Scenario s3 is omitted due to its extremely slow computation.}
\label{fig:time_error}
\vskip 0em
\end{figure}

We first compare averaged computational time of all methods, see the first row of Figure \ref{fig:time_error}. Clearly, Tlasso is dramatically faster than both competing methods. In particular, in Scenario s3, Tlasso takes about three seconds for each replicate.  P-MLE takes about one minute while the direct Glasso method takes more than half an hour and is omitted in the plot. As we will show below, Tlasso algorithm is not only computationally efficient but also enjoys good estimation accuracy and support recovery performance.

In the second row of Figure \ref{fig:time_error}, we compute  averaged estimation errors of  Kronecker product of precision matrices. Clearly, with respect to tensor graphical structure, the direct Glasso method has significantly larger errors than Tlasso and P-MLE method. Tlasso outperforms P-MLE in Scenarios s1 and s2 and is comparable to P-MLE in Scenario s3. It is worth noting that, in Scenario s3,  P-MLE is $20$ times slower than Tlasso.

\begin{figure}[h!]
\centering
\includegraphics[scale=0.3]{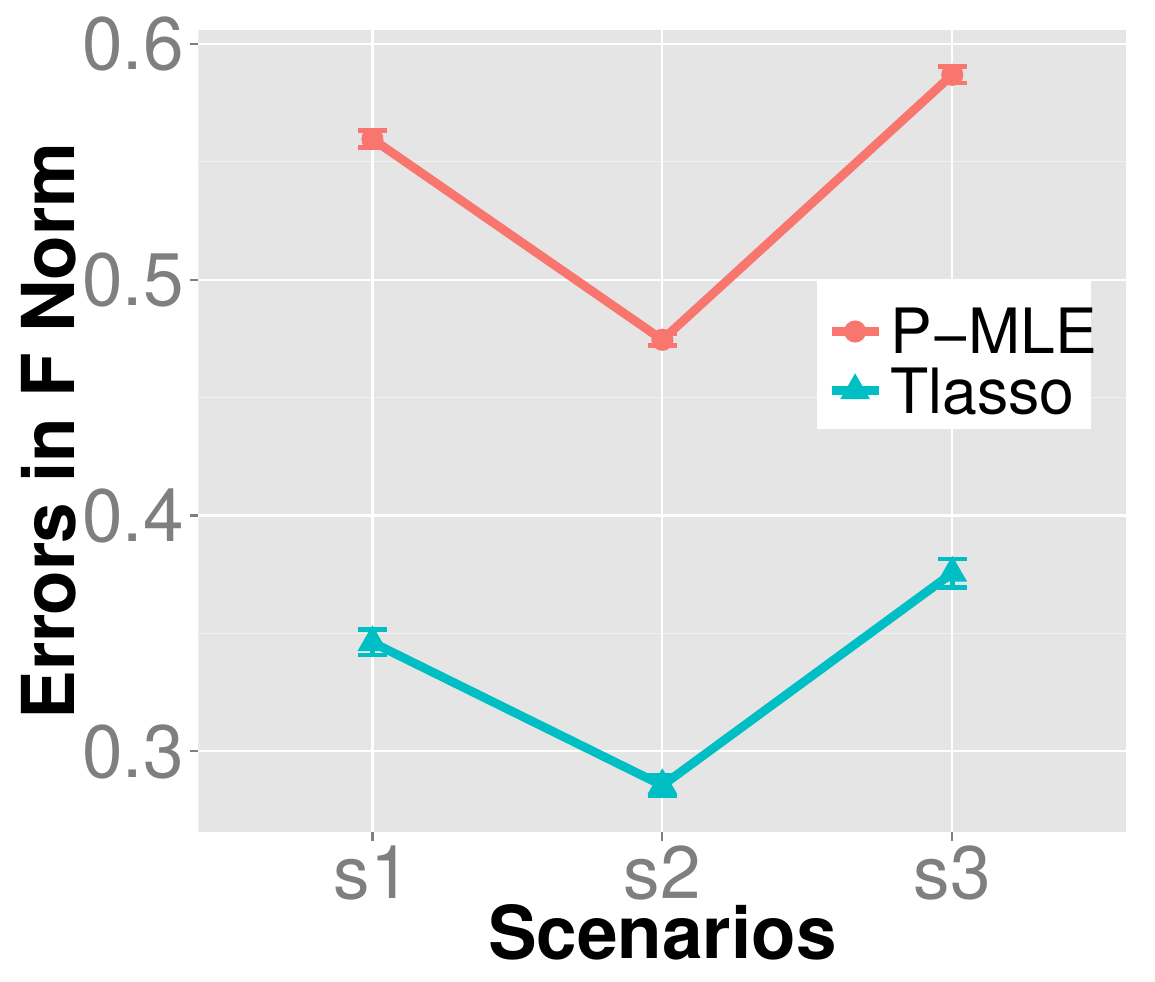}
\includegraphics[scale=0.3]{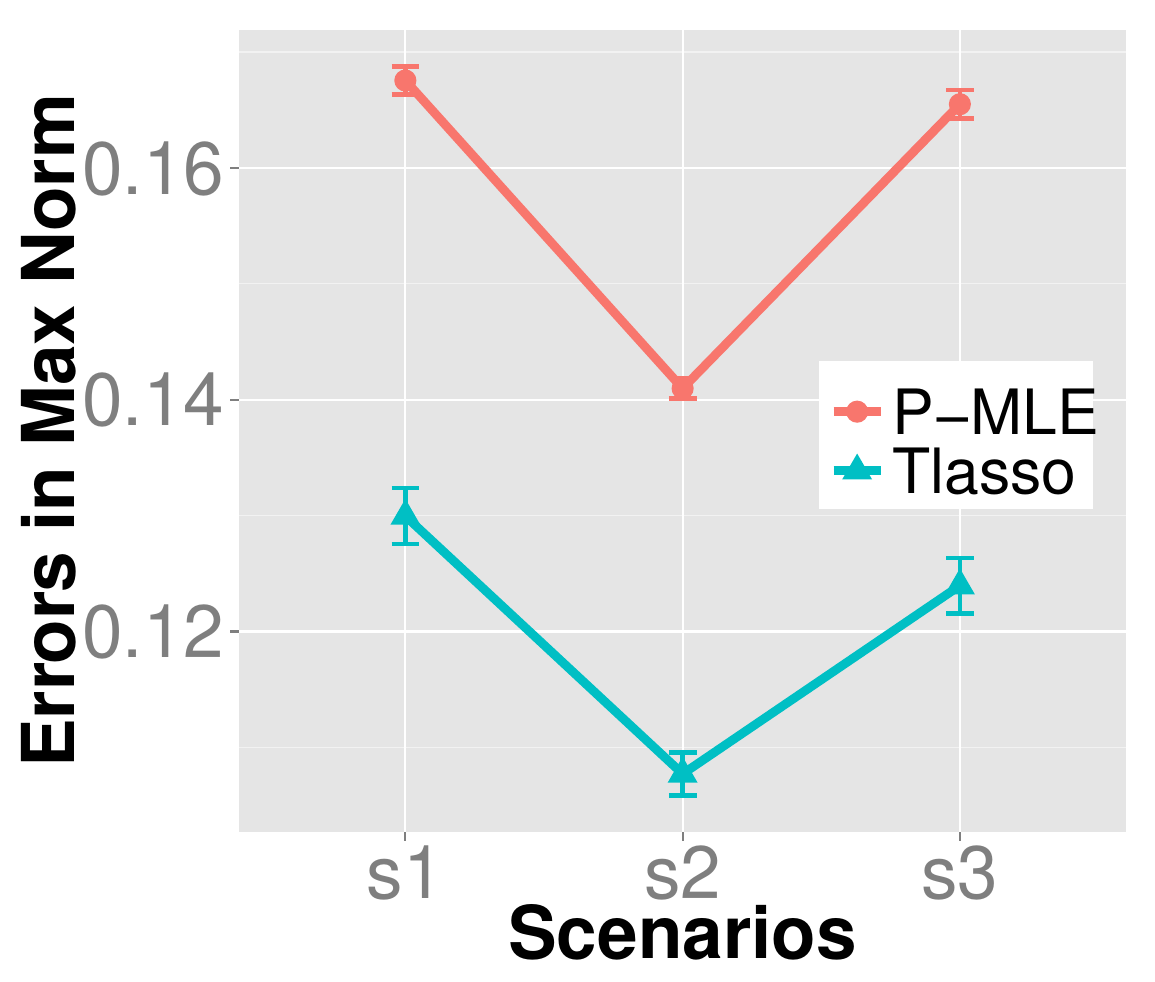}
\includegraphics[scale=0.3]{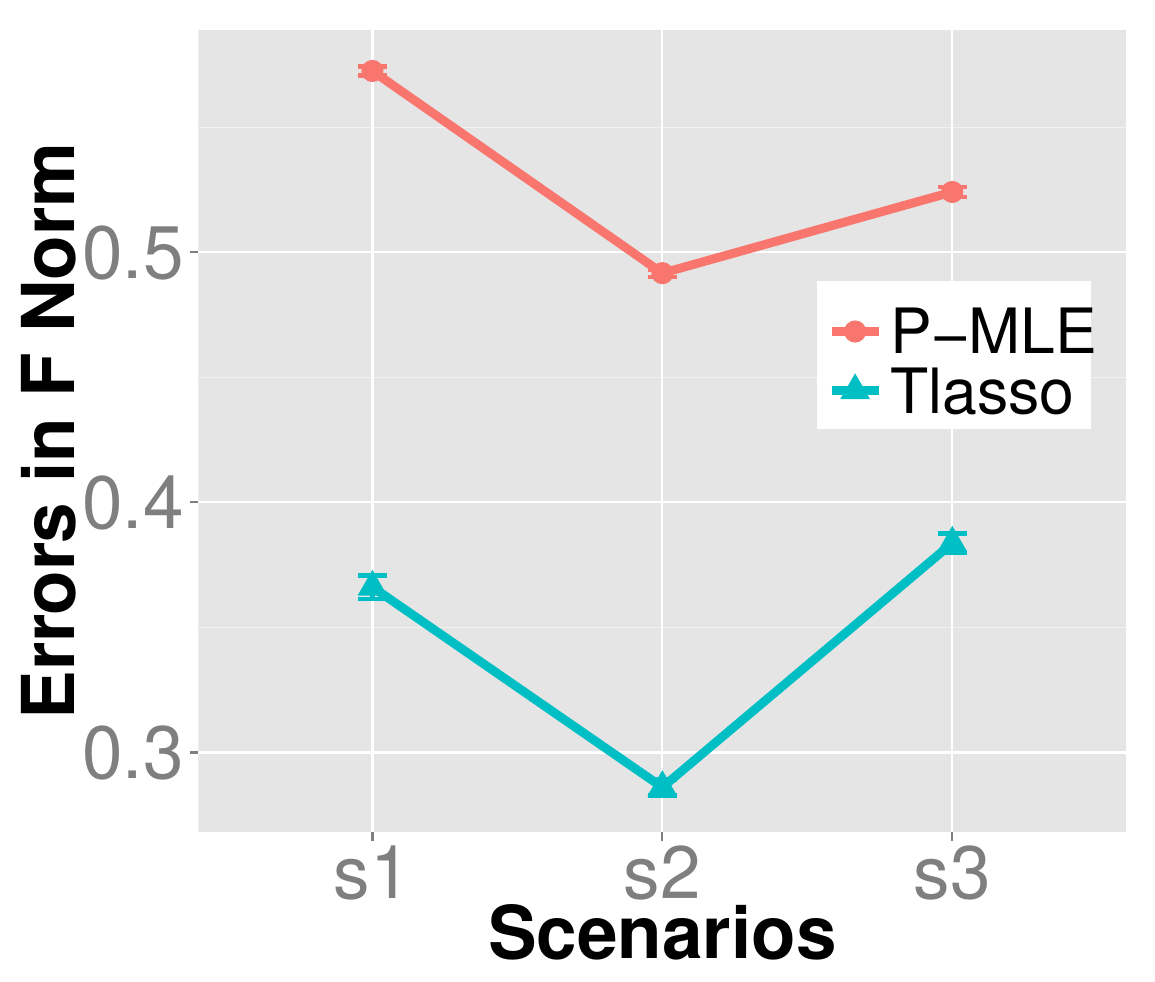}
\includegraphics[scale=0.3]{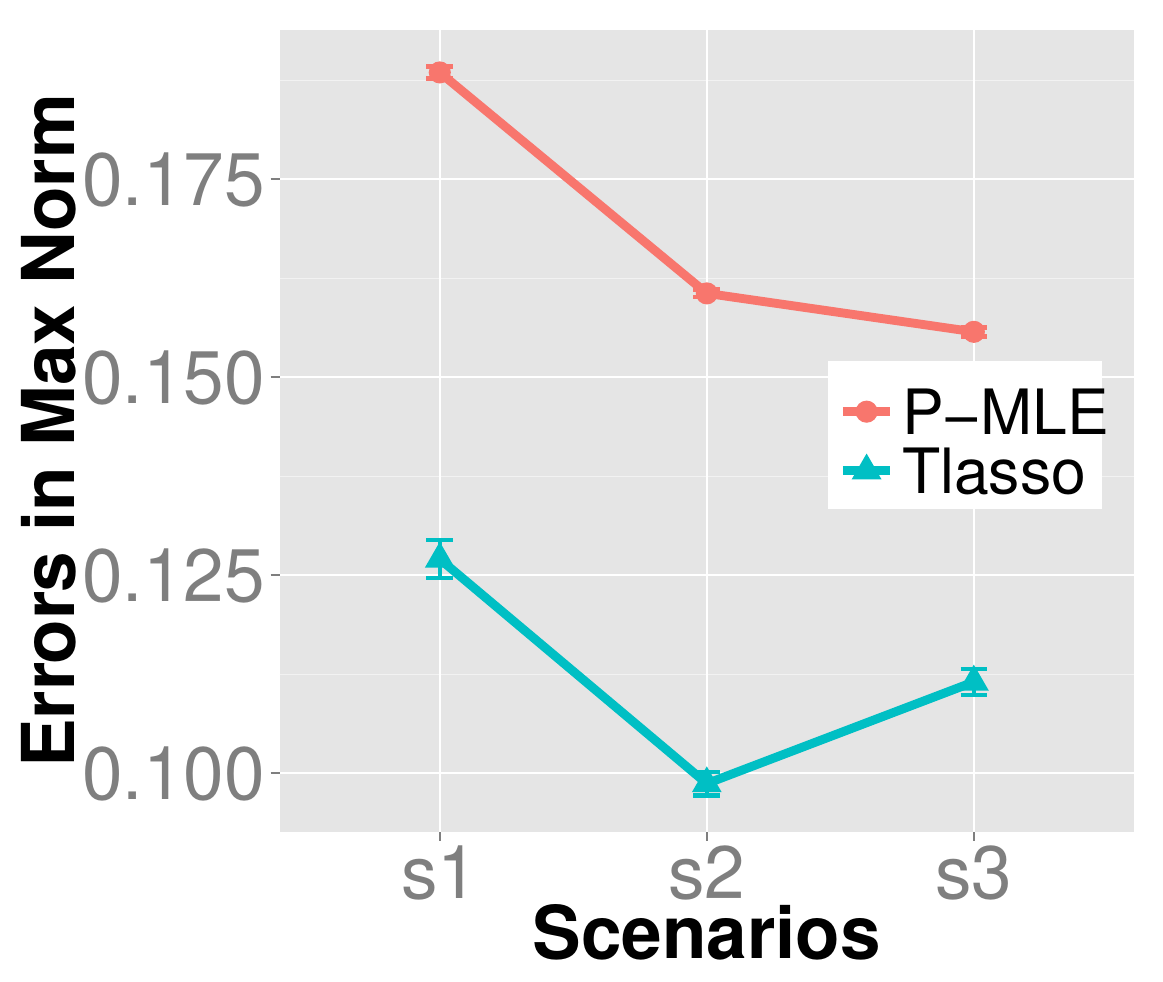}
\caption{Averaged estimation errors of  precision matrices in Frobenius norm and max norm of each method in Simulations $1\&2$, respectively. The first row is for Simulation 1, and the second row is for Simulation 2.}
\label{fig:error_av}
\vskip 0em
\end{figure}

Next, we evaluate  averaged estimation errors of precision matrices in Frobenius norm and max norm for  Tlasso and  P-MLE method. The direct Glasso method only estimate the whole Kronecker product, hence can not produce estimate for each precision matrix. Recall that, as we show in Theorem \ref{thm:final_error} and Theorem \ref{thm:maxnorm},  estimation error for the $k$-th precision matrix is $O_P(\sqrt{ m_k(m_k+s_k)\log m_k /(nm) })$ in Frobenius norm and $O_P(\sqrt{ m_k\log m_k /(nm) })$ in max norm, where $m=m_1m_2m_3$ in this example. These theoretical findings are supported by  numerical results in Figure \ref{fig:error_av}. In particular, as sample size $n$ increases from Scenario s1 to s2,  estimation errors in both Frobenius norm and max norm expectedly decrease. From Scenario s1 to s3, one dimension $m_3$ increases from $10$ to $20$, and other dimensions $m_1, m_2$ keep the same, in which case  averaged estimation error in max norm decreases, while error in Frobenius norm increases due to its additional $\sqrt{m_k+s_k}$ effect. Moreover, compared with  P-MLE method,  Tlasso demonstrates significant better performance in all three scenarios in terms of both Frobenius norm and max norm.

From here, we turn to  numerical study of the proposed inference procedure. Estimation of precision matrices in the inference procedure is conducted under the same setting as the former numerical study of Tlasso algorithm. Similarly,  two simulations are considered, i.e., triangle graph and nearest neighbor graph. In both simulations, third-order tensors are constructed, adopting the same three scenarios as above: Scenario {s1}, {s2}, and {s3}. Each scenario repeats  100 times.



\begin{figure}[h!]
\centering
\includegraphics[scale=0.25]{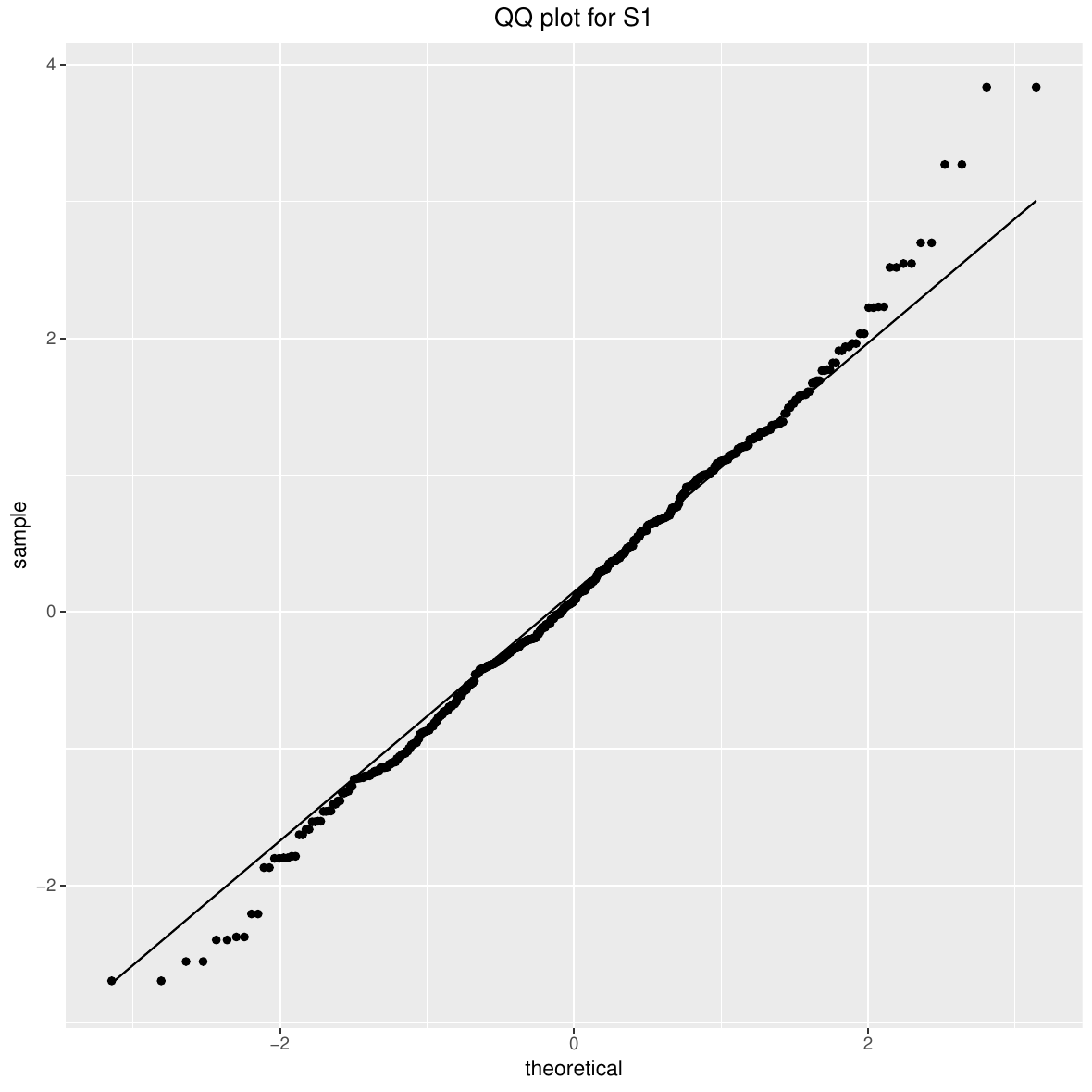}
\includegraphics[scale=0.25]{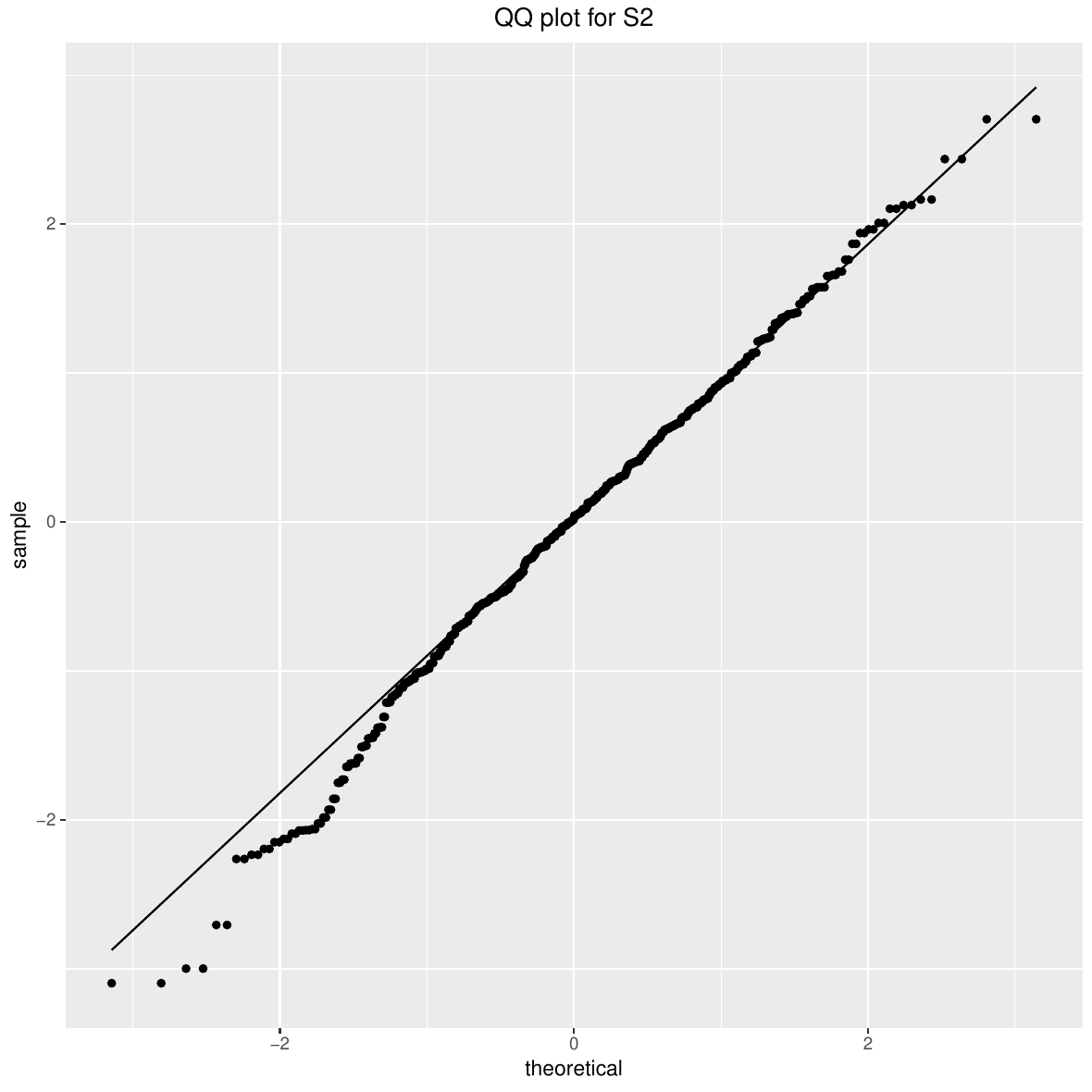}
\includegraphics[scale=0.25]{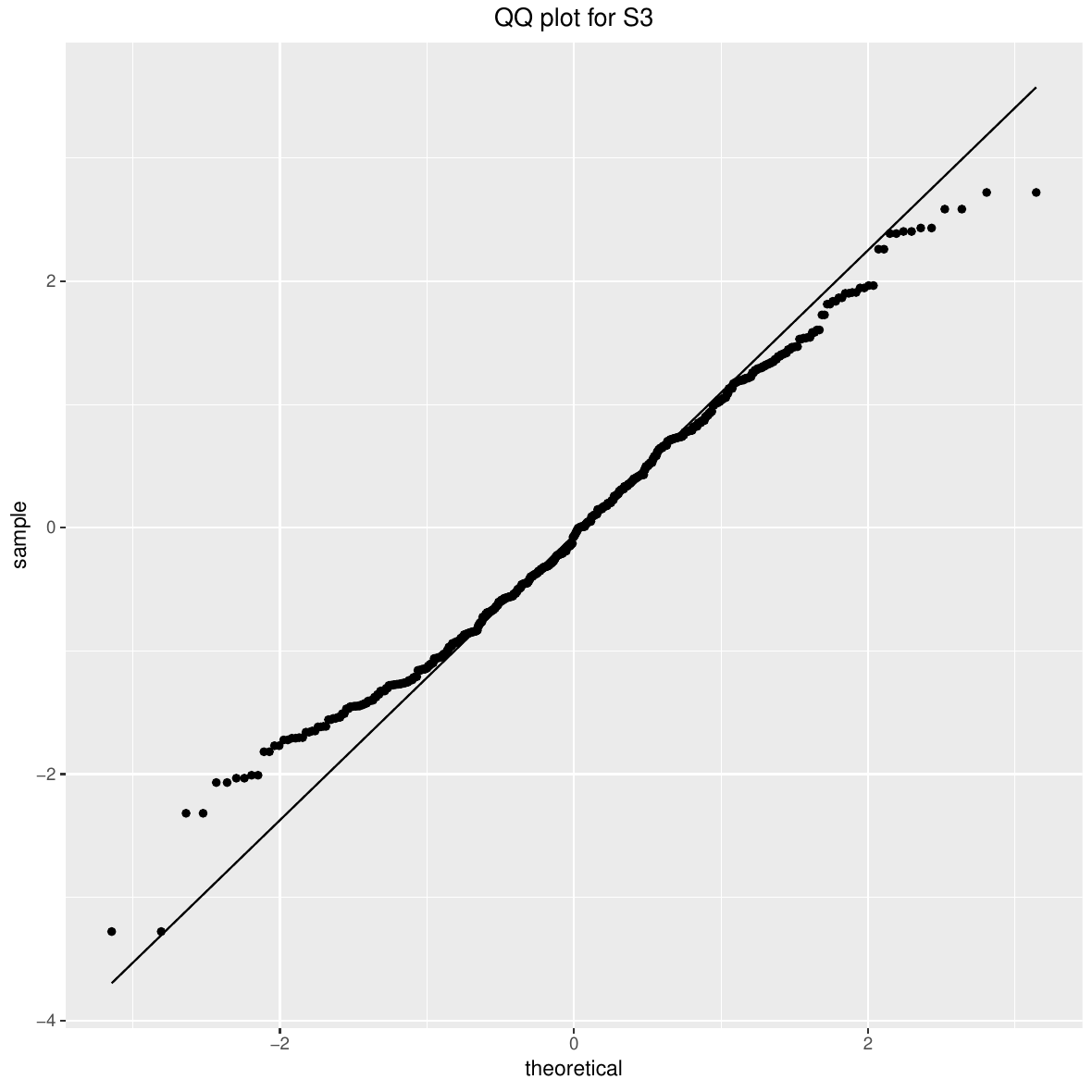}
\includegraphics[scale=0.25]{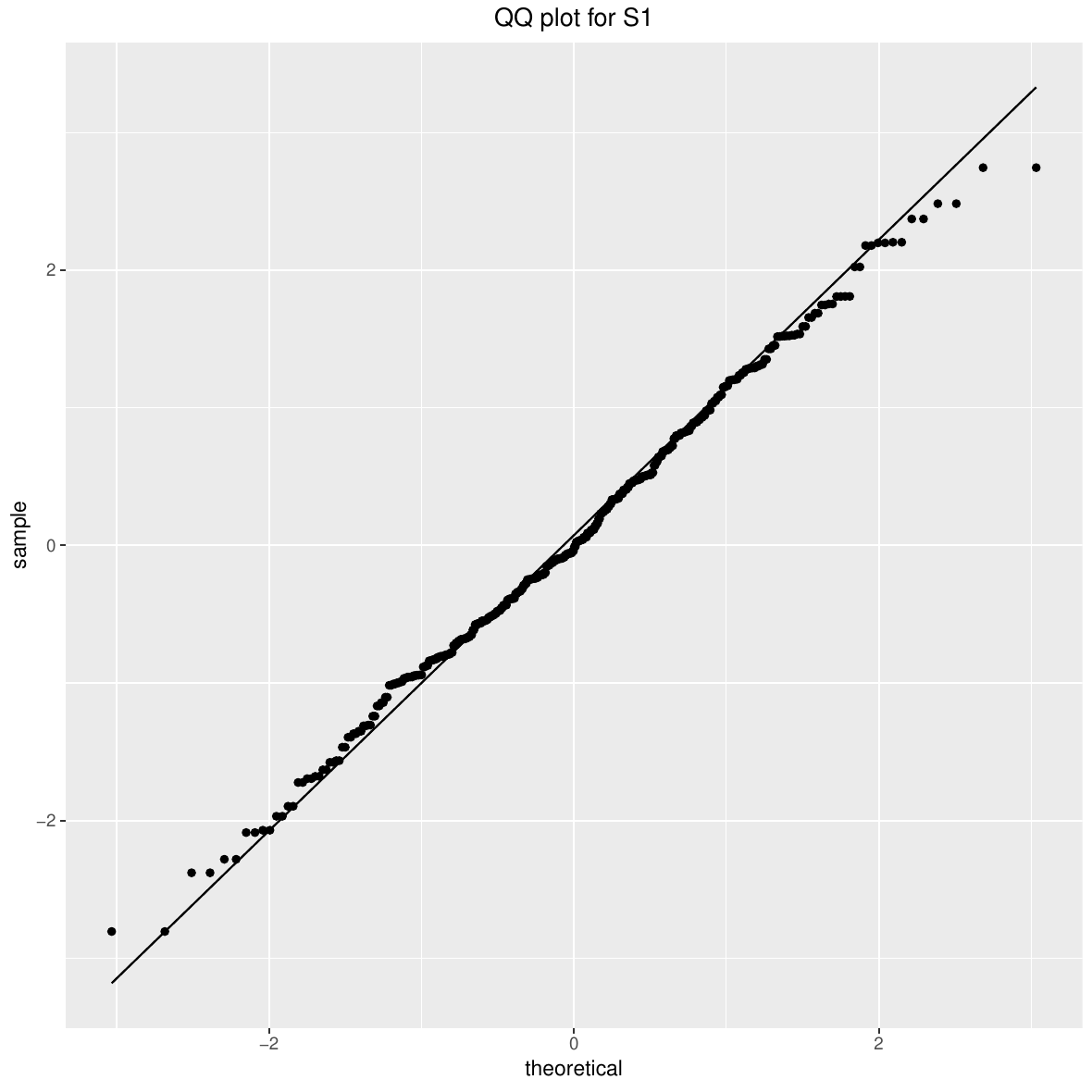}
\includegraphics[scale=0.25]{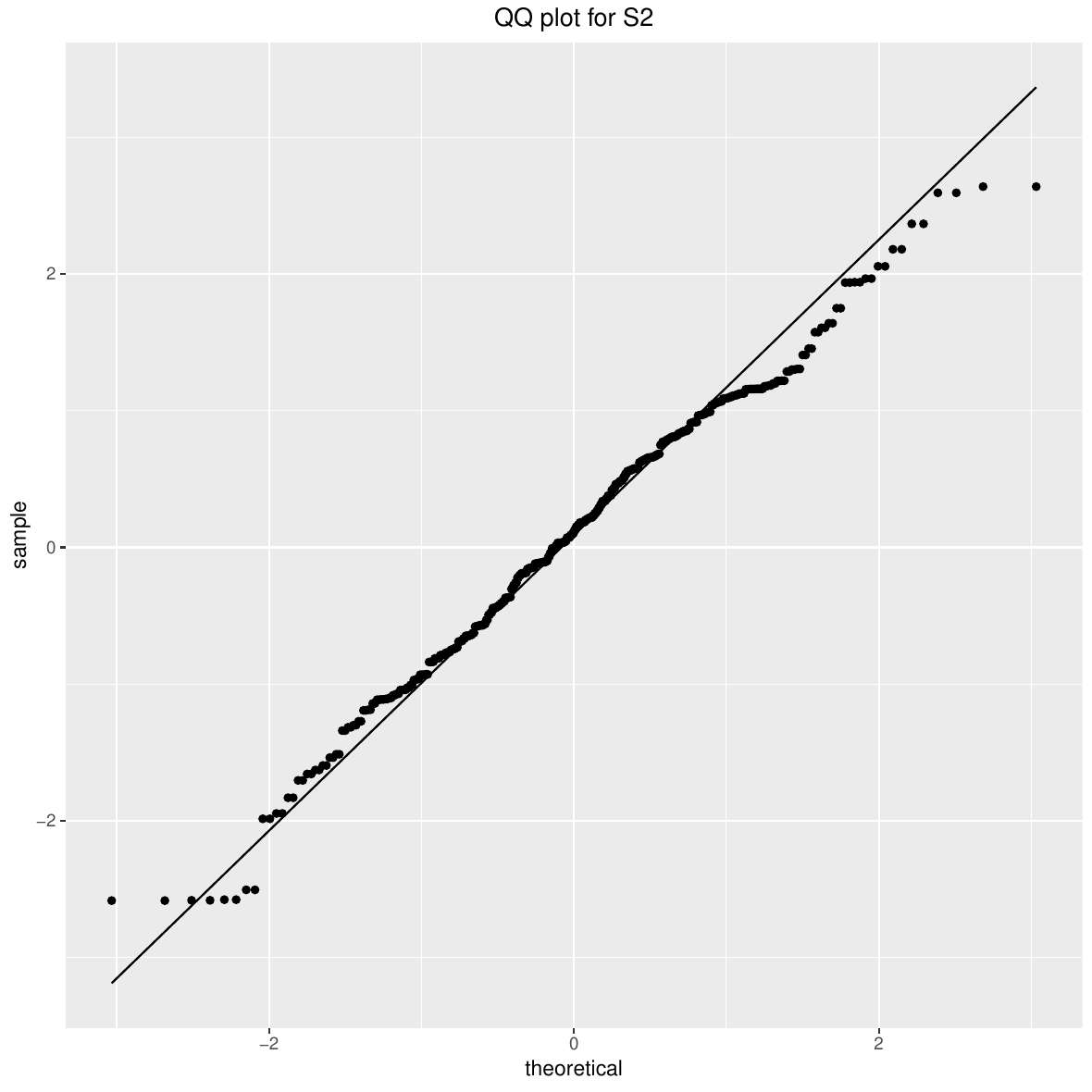}
\includegraphics[scale=0.25]{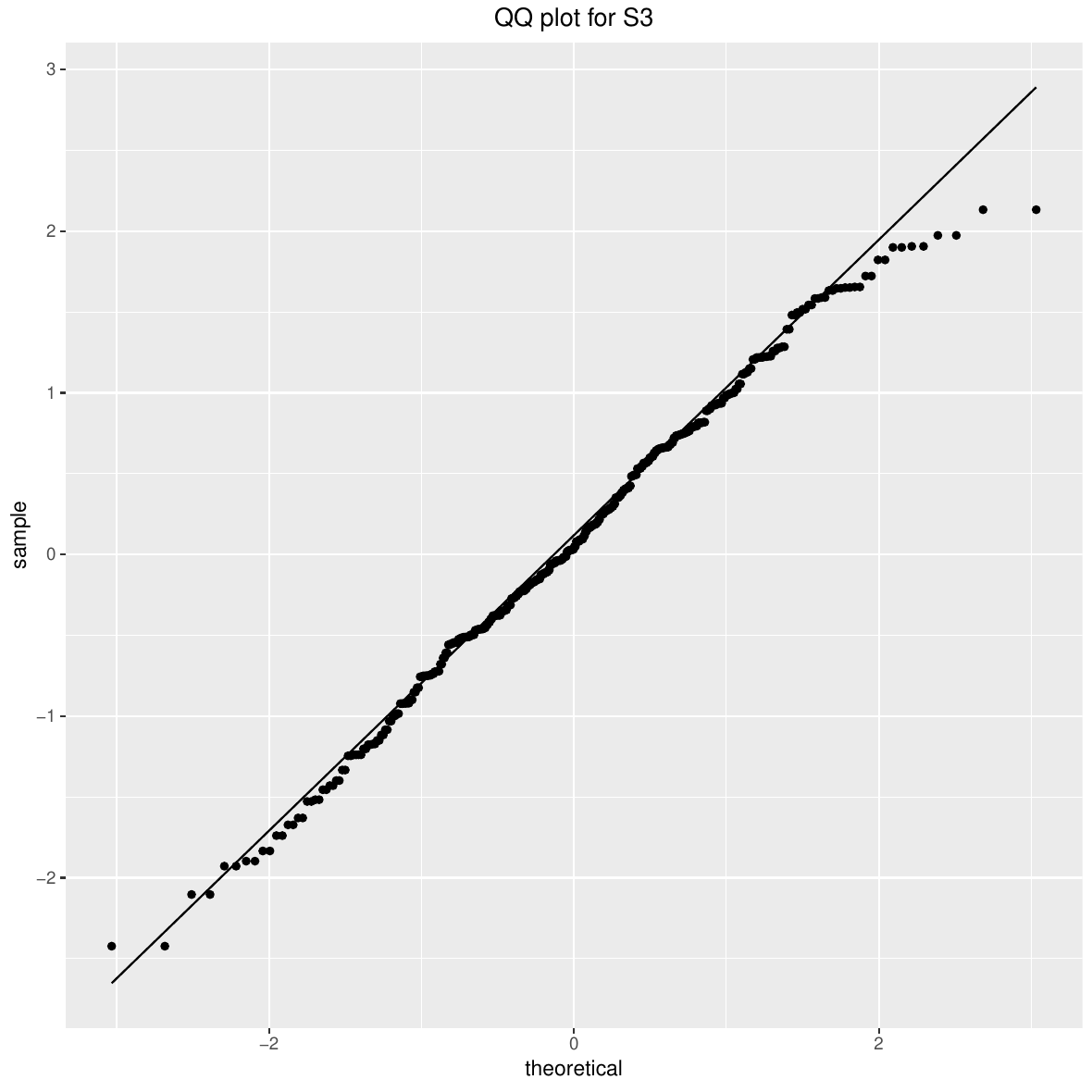}
\caption{QQ plots for fixed zero entry $[\bOmega_1^*]_{6,1}$. From left column to right column is scenario s1, s2 and s3. The first row is simulation 1, and the second  is simulation 2.}
\label{fig: test_consist}
\vskip 0em
\end{figure}

We first evaluate asymptotic normality of our test statistic $\tilde{\tau}_{i,j}$. Figure \ref{fig: test_consist} demonstrates QQ plots of test statistic for fixed zero entry $[\bOmega_1^*]_{6,1}$. Some other zero entries have been selected, and their simulation results are similar. So we only present results of $[\bOmega_1^*]_{6,1}$ in this section. 
As shown in Figure \ref{fig: test_consist}, our test statistic behaves very similar to standard normal even when sample size is small and dimensionality is high. 
It results from the fact that our inference method fully utilizes tensor structure.  


Then we investigate the validity of our FDR control procedure. 
Table \ref{tab:FDP_P} contains FDP, its theoretical limit $\tau$ (see Remark \ref{rem: kro_prod_control}), and power (all in $\%$) for Kronecker product of precision matrices under FDR control. 
Oracle procedure utilizes true covariance and precision matrices to compute test statistic. 
Each mode has the same pre-specific level $\upsilon =5\% $ or $10\%$. 
As show in Table \ref{tab:FDP_P}, powers are almost one and FDPs are small under poor conditions, i.e, small sample size or large dimensionality. 
It implies that our inference method has superior support recovery performance. 
Besides, empirical FDPs get closer to their theoretical limits if either of dimensionality and sample size is larger.
This phenomenon backs up the theoretical justification in Theorem \ref{thm: FDR_control}.
Thanks to fully utilizing tensor structure, difference between oracle FPR and our data-driven FDP decreases as either dimensionality or sample size grows. 

\begin{table}[htb]
\small{
	\caption{Empirical $\textrm{FDP}$, its theoretical limit $\tau$, and $\textrm{power}$ (all in $\%$) of our inference procedure under FDP control for the Kronecker product of precision matrices in scenario s1, s2, and s3.}
	\label{tab:FDP_P}
\begin{center}
	\vskip -1 em
	\begin{tabular}{|p {1em}| p{2.7em}  | p {2em}  p {2em}  p {2em}|p {2em} p {2em} p {2em}|}
	\hline 
	 \multirow{2}{*}{}&  & \multicolumn{3}{c | }{Sim1}   & \multicolumn{3}{c }{Sim2}  \vline  \\
		$\upsilon$ & & s1 & s2 & s3 & s1 & s2 & s3     \\ \hline 
			  &&  \multicolumn{6}{c|}{Empirical FDP ($\tau$)} \\ 	\hline
\multirow{2}{*}{5}  & oracle & 7.8  & 8.7 & 7.3  & 7.6  & 7.3  & 6.9  \\ 
		\hhline{~-------} 
			 & data-driven & 6.7 (9.9) & 7.4 (9.9) & 7.2 (9.9)& 6.9 (11.1) & 7 (11.1) & 7.3 (11.1)  \\ 
		\hline
	\multirow{2}{*}{10} & oracle & 15.7  & 16.2  & 14.9& 15.1  & 14.5  & 14.7  \\ 
		\hhline{~-------} 
 		& data-driven & 13.8 (19.3) & 15.4 (19.3) & 15.1 (19.4) & 13.8 (21.4) & 13.9 (21.4) & 14.9 (21.4)\\  
		\hline
	&&  \multicolumn{6}{c|}{Empirical Power } \\ 	\hline
	\multirow{2}{*}{5}	 & oracle & 100  & 100  & 100 & 99.9  & 100  & 99.9  \\ 	
		\hhline{~-------}
				 & data-driven & 100  & 100  & 100 & 99.8  & 100  & 99.8 \\ 
		\hline
		\multirow{2}{*}{10} & oracle & 100  & 100  & 100 & 100  & 100  & 100 \\ 
		\hhline{~-------}
 			& data-driven & 100  & 100  & 100& 99.9  & 100  & 99.9   \\ 
	\hline
	\end{tabular}
	\end{center} }
	\vskip -0.5em
\end{table}

In the end, we evaluate the true positive rate (TPR) and the true negative rate (TNR) of the Kronecker product of precision matrices for Glasso, P-MLE, and our FDP control procedure to compare their model selection performance. Specifically, let $a^*_{i,j}$ be the $(i,j)$-th entry of $\bOmega^*_1 \otimes \cdots \otimes \bOmega^*_K$ and $\hat{a}_{i,j}$ be the $(i,j)$-th entry of $\hat{\bOmega}_1 \otimes \cdots \otimes \hat{\bOmega}_K$, TPR and TNR of the Kronecker product are
$\sum_{i,j} \ind( \widehat{a}_{i,j} \ne 0, a^*_{i,j} \ne 0)  / \sum_{i,j} \ind(a^*_{i,j} \ne 0)$, and $\sum_{i,j} \ind( \widehat{a}_{i,j} = 0, a^*_{i,j} = 0)  / \sum_{i} \ind(a^*_{i,j} = 0).$
Pre-specific FDP level is $\upsilon=5\%$. 
Table \ref{tab:selection} shows the model selection performance of all three methods. 
A good model selection procedure should produce large TPR and TNR.
Our FDP control procedure has dominating TPR and TNR against the rest methods, i.e., almost all edges are identified and few non-connected edges are included. 

\begin{table}[h!]
\small{
\centering
\vskip -1.5em
\caption{Model selection performance comparison among Glasso, P-MLE, and our FDP control procedure. Here TPR and TNR denote the true positive rate and true negative rate of the Kronecker product of precision matrices.}
\label{tab:selection}
\vskip 0.5em
\begin{tabular}{|c|cc|cc|cc|cc|}
\hline
\multirow{2}{*}{Scenarios} & \multicolumn{2}{c|}{Glasso}  & \multicolumn{2}{c|}{P-MLE}& \multicolumn{2}{c|}{Our FDP control} \\
 & TPR & TNR & TPR & TNR & TPR & TNR  \\
\hline
\quad\quad~ s1&  0.343&  0.930&  1& 0.893&  1& 0.935 \\
Sim1 s2&  0.333&  0.931&  1& 0.894&  1& 0.932 \\
\quad\quad~ s3&  0.146&  0.969&  1& 0.941&   1& 0.929\\
\hline
\quad\quad~ s1&  0.152&  0.917&  1& 0.854&  0.999& 0.926  \\
Sim2  s2&  0.119&  0.938&  1& 0.851&  1& 0.926\\
\quad\quad~ s3&  0.078&  0.962&  1& 0.937& 0.998& 0.928 \\
\hline
\end{tabular}
\vskip -1em
}
\end{table}

In short, the superior numerical performance and cheap computational cost in these simulations suggest that our method could be a competitive estimation and inferential tool for tensor graphical model in real-world applications.

 
\section{Real Data Analysis} \label{sec: real_data}
In this section, we apply our inference procedure on two real data sets. In particular, the first data set is from the Autism Brain Imaging Data Exchange (ABIDE), a study for autism spectrum disorder (ASD); the second set collects users' advertisement clicking behaviors from a major Internet company.
	
\subsection{ABIDE}
\label{sec:ABIDE}

In this subsection, we analyze a real ASD neuroimaging dataset, i.e., ABIDE, to illustrate proposed inference procedure.
As an increasingly prevalent neurodevelopmental disorder, symptoms of ASD are social difficulties, communication deficits, stereotyped behaviors and cognitive delays \cite{RBBHDTBD13}. 
It is of scientific interest to understand how connectivity pattern of brain functional architecture differs between ASD subjects and normal controls. 
After preprocessing, ABIDE consists of the resting-state functional magnetic resonance imaging (fMRI) of 1071 subjects, of which 514 have ASD, and 557 are normal controls.
fMRI image from each subject takes the form of a $30\times36\times30$-dimensional tensor of \textit{fractional amplitude of low-frequency fluctuations} (fALFF), calculated at each brain voxels.
In other words, ABIDE has 514+557 tensor images (each of dimension $30\times36\times30$) from ASD and controls, and these tensor images are 3D scans of human brain, whose entry values are fALFF of brain voxels at corresponding spatial locations.
fALFF is a metric characterizing intensity of spontaneous brain activities, and thus quantifies functional architecture of the brain \cite{SK15}.
Therefore the support of precision matrix of fALFF fMRI images along each mode encodes the connectivity pattern of brain functional architecture.
Dissimilarity in the supports between ASD and controls reveals potentially differential connectivity pattern. 
In this problem, vectorization methods, such as Glasso, will lose track of mode-specific structures, and thus can not be applied.
Due to high dimensionality, false positive becomes a critical issue. However, P-MLE fails to guarantee FDP control as demonstrated in the simulation studies.

We apply the proposed inference procedure to recover the support of mode precision matrices of fALFF fMRI images of ASD group (514 image tensors) and normal control group (557 image tensors), respectively. 
Pre-specific FDP level is set as $0.01\%$. 
The rest setup is the same as in \S\ref{sec:simulation}.
Among the rejected entries of each group, we choose top 60 significant ones (smallest p-values) along each mode. 
All the selected entries show p-values less than 0.01\%. 
Positions of differential entries between ASD and controls are recorded and mapped back to corresponding brain voxels. 
We further locate the voxels in the commonly used Anatomical Automatic Labeling (AAL) atlas \cite{TLPCEDMJ02}, which consists of 116 brain regions of interest.
Brain regions including the voxels, listed in Table \ref{tab:brain_regions}, are suspected to have differential connectivity patterns between ASD and normal controls.  

Our results in general match the established literature. 
For example, postcentral gyrus agrees with \cite{HSEM10}, which identifies postcentral gyrus as a key region where brain structure differs in autism.
Also, \cite{NTSSM13} suggests that thalamus plays a role in motor abnormalities reported in autism studies. 
Moreover, temporal lobe demonstrates differential brain activity and brain volume in autism subjects \cite{HSKSC15}. 

\begin{table}[htb]
\small{
	\caption{Brain regions of potentially differential connectivity pattern identified by our inference procedure.}
	\label{tab:brain_regions}
\begin{center}
	\vskip -1 em
	\begin{tabular}{| p {8em}  p {8.2em}  p {8.2em}|}
	\hline 
Hippocampus\_L     &  ParaHippocampal\_R &  Hippocampus\_R     \\
Temporal\_Sup\_L    &  Amygdala\_L        &  Temporal\_Sup\_R    \\ 
Insula\_L          &  Amygdala\_R        &  Insula\_R          \\
Frontal\_Mid\_R&  Thalamus\_L        &  Thalamus\_R       \\
Pallidum\_L        &  Putamen\_L         &  Caudate\_R          \\
Precentral\_L      &  Frontal\_Inf\_Oper\_L&  Frontal\_Inf\_Oper\_R \\
Precentral\_R      &  Postcentral\_L    &  Postcentral\_R     \\
Temporal\_Pole\_Sup\_R     &  & \\
	\hline
	\end{tabular}
	\end{center} }
	\vskip -0.5em
\end{table} 

%
%
%

\subsection{Advertisement Click Data}
In this subsection, we apply the proposed inference method to an online advertising data set from a major Internet company. 
This dataset consists of click-through rates (CTR), i.e., the number of times a user has clicked on an advertisement from a certain device divided by the number of times the user has seen that advertisement from the  device, for advertisements displayed on the company's webpages from May 19, 2016 to June 15, 2016. 
It tracks clicking behaviors of 814 users for 16 groups of advertisement from 19 publishers on each day of weeks, conditional on two devices, i.e., PC and mobile.
Thus, two $16\times 19 \times 7 \times 814$ tensors are formed by computing CTR corresponds to each (\texttt{advertisement}, \texttt{publisher}, \texttt{dayofweek}, \texttt{users}) quadruplet, conditional on PC and mobile respectively. 
However, more than 95\% entries of either CTR tensor are missing. 
Hence, an alternating minimization tensor completion algorithm \cite{JO14} is first conducted on the two tensors.
Differential dependence structures within advertisements, publishers, and days of weeks between PC and mobile are of particular business interest. 
Therefore, we apply the proposed inference procedure to \texttt{advertisement}, \texttt{publisher}, and \texttt{dayofweek} modes of completed PC and mobile tensors respectively. 
Setup is the same as in \S\ref{sec:ABIDE}.
Among the rejected entries of each device, top $(30, 12, 10)$ significant ones in mode (\texttt{advertisement}, \texttt{publisher}, \texttt{dayofweek}) are selected.
All the selected entries show p-values less than 0.01\%.
Pairs of entities, represented by the positions of differential entries between PC and mobile, are suspected to display dissimilar dependence when switching device. 

\begin{figure}[h!]
\centering
\caption{Analysis of the advertisement clicking data. Shown are differential dependence patterns between PC (red lines) and mobile (black lines) identified by our inference procedure. From left to right are advertisements, publishers, days of weeks.}
\includegraphics[scale=0.26]{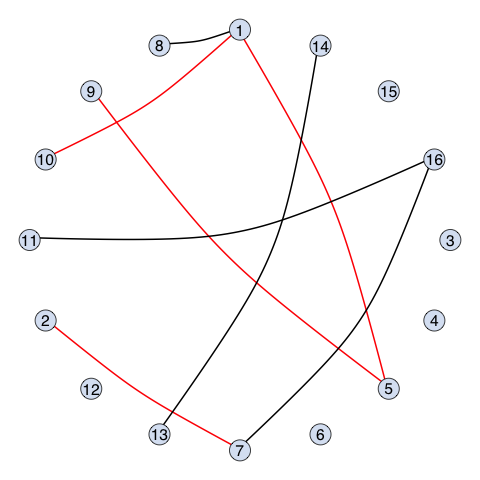}
\includegraphics[scale=0.26]{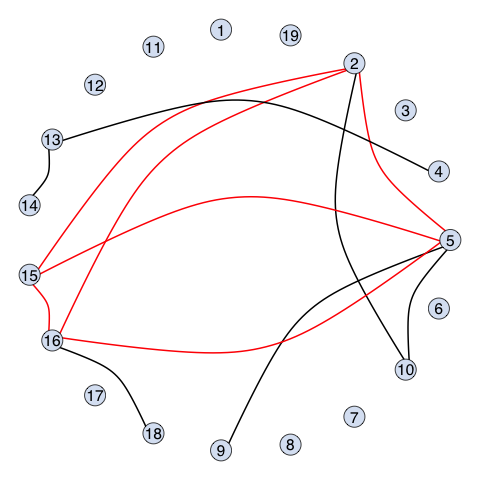}
\includegraphics[scale=0.26]{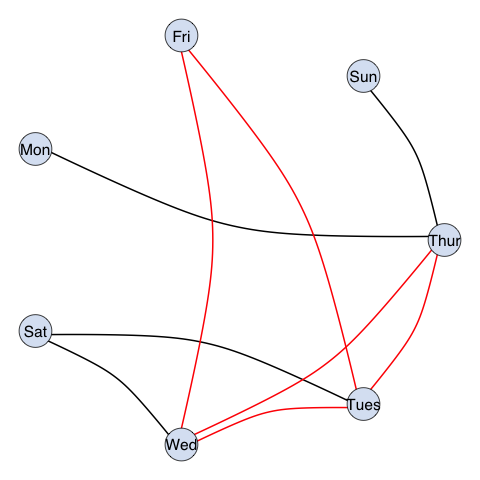}
\label{fig: differ_dependence}
\vskip 0em
\end{figure}

Figure \ref{fig: differ_dependence} demonstrates differential dependence patterns between PC and mobile in terms of advertisement, publishers, and days of weeks.
Note that red lines indicate dependence only on PC, and black lines stand for those only on mobile.
Due to confidential reason, description on specific entity of \texttt{advertisement} and \texttt{publisher} is not presented. 
We only provide general interpretations on the identified differential dependence patterns as follows.
In \texttt{advertisement} mode, credit card ads and mortgage ads are linked on mobile.
Such dependence is reasonable that people involved in mortgage would be more interested in credit card ads. 
On PC, uber share and solar energy are interchained. 
It can be interpreted in the sense that both uber share and solar energy are attractive for customers with energy-saving awareness.
As for \texttt{publisher} mode, sport news publisher and weather news publisher are shown to be dependent on PC. 
This phenomenon can be accounted by the fact that sports and weather are the two most popular news choices when browsing websites. 
Also, magazine publishers (e.g., beauty magazines, tech magazines, and TV magazines) are connected on mobile.  
It is reasonable in the sense that people tend to read several casual magazines on mobiles for relaxing or during waiting. 
In \texttt{dayofweek} mode, strong dependence is demonstrated among weekdays, say from Tuesday to Friday, on PC. 
However, no clear pattern is showed on mobile. 
It can be explained that employees operate PC mostly at work on weekdays but use mobile every day.

\section{Discussion}
\label{sec:discussion}

In this paper, we propose a novel sparse tensor graphical model to analyze graphical structure of high-dimensional tensor data. 
An efficient Tlasso algorithm is developed, which attains an estimator with minimax-optimal convergence rate in estimation. 
Tlasso algorithm not only is much faster than alternative approaches but also demonstrates superior estimation accuracy.
In order to recover graph connectivity, we further develop an inference procedure with FDP control.
Its asymptotic normality and validity of FDP control is rigorously justified.
Numerical studies demonstrate its superior model selection performance. 
The above evidences motivate our methods more practically useful in comparison to other alternatives on real-life applications.

In Tlasso algorithm, graphical lasso penalty is applied for updating each precision matrix of tensor data. Lasso penalty is conceptually simple and computationally efficient. However, it is known to induce additional bias in estimation. In practice, other non-convex penalties, like SCAD \cite{fan2001}, MCP \cite{zhang2010}, or Truncated $\ell_1$ \cite{shen2012}, are able to correct such bias. Optimization properties of non-convex penalized high-dimensional models have recently been studied by \cite{wang2014b}, which enables theoretical analysis of sparse tensor graphical model with non-convex penalties.

\section*{Acknowledgement}
 
Han Liu is grateful for the support of NSF CAREER Award DMS1454377, NSF IIS1408910, NSF IIS1332109, NIH R01MH102339, NIH R01GM083084, and NIH R01HG06841. 
Guang Cheng's research is sponsored by NSF CAREER Award DMS-1151692, NSF DMS-1418042, DMS-1712907, Simons Fellowship in Mathematics, Office of Naval Research (ONR N00014-15-1-2331).
Will Wei Sun was visiting Princeton and Guang Cheng was on sabbatical at Princeton while this work was carried out; Will Wei Sun and Guang Cheng would like to thank Princeton ORFE department for their hospitality and support.

{{\small

}
}

\newpage

\setcounter{footnote}{0}
\setcounter{section}{0}
\renewcommand{\thesection}{S.\arabic{section}}
\setcounter{equation}{0}
\counterwithout{equation}{section}
\renewcommand{\theequation}{S.\arabic{equation}}
\setcounter{theorem}{0}
\counterwithout{theorem}{section}
\renewcommand{\thetheorem}{S.\arabic{theorem}}
\renewcommand{\thelemma}{S.\arabic{theorem}}

\setcounter{page}{1}
\begin{center}
\textit{\large Supplementary Material to}
\end{center}
\begin{center}
\title{\LARGE Tensor Graphical Model: Non-convex Optimization and Statistical Inference}
\vskip10pt
\end{center}

The supplementary material is organized as follows:

\begin{itemize}
\item In Section \ref{sec:proof_theorem}, proofs of main theorems are provided. 
\item In Section \ref{sec:key_lemma}, proofs of key lemmas are provided.
\item In Section \ref{sec:other_lemma}, auxiliary lemmas are listed.
\item In Section \ref{sec:sensitivity}, additional numerical results of sensitivity analysis for tuning parameter are provided.
\item In Section \ref{sec:add_simulation}, additional numerical results on the effects of sample size and dimensionality are presented.
\end{itemize}

 
\section{Proof of Main Theorems}
\label{sec:proof_theorem}

\noindent{\bf Proof of Theorem \ref{thm:population_tensor}:} To ease the presentation, we show that Theorem \ref{thm:population_tensor} holds when $K=3$. The proof can be easily generalized to the case with $K>3$. 

We first simplify the population log-likelihood function. Note that when ${\cal T} \sim \textrm{TN}({\bf0}; \bSigma_1^*, \bSigma_2^*, \bSigma_3^*)$, Lemma 1 of \cite{he2014} implies that $\textrm{vec}({\cal T}) \sim \textrm{N}(\textrm{vec}({\bf0}); \bSigma_3^* \otimes \bSigma_2^* \otimes \bSigma_1^*)$. Therefore, 
\begin{align*}
& \mathbb E \big\{   \textrm{tr} \big[ \textrm{vec}({\cal T}) \textrm{vec}({\cal T})^{\top} (\bOmega_3 \otimes \bOmega_2 \otimes \bOmega_1) \big]   \big\} \\
=&  \textrm{tr} \big[ (\bSigma_3^* \otimes \bSigma_2^* \otimes \bSigma_1^*)(\bOmega_3 \otimes \bOmega_2 \otimes \bOmega_1)  \big]\\
 =&   \textrm{tr}(\bSigma_3^*\bOmega_3)\textrm{tr}(\bSigma_2^*\bOmega_2)\textrm{tr}(\bSigma_1^*\bOmega_1),
\end{align*}
where the second equality is due to the properties of Kronecker product that $(\Ab \otimes \Bb)(\Cb \otimes \Db) = (\Ab \Cb) \otimes (\Bb \Db)$ and $\textrm{tr}(\Ab \otimes \Bb) = \textrm{tr}(\Ab) \textrm{tr}(\Bb)$. Therefore, the population log-likelihood function can be rewritten as
\begin{align*}
q(\bOmega_1, \bOmega_2, \bOmega_3) = &\frac{ \textrm{tr}(\bSigma_3^*\bOmega_3)\textrm{tr}(\bSigma_2^*\bOmega_2)\textrm{tr}(\bSigma_1^*\bOmega_1) }{m_1 m_2 m_3} - \frac{1}{m_1} \log |\bOmega_1| \\
&-  \frac{1}{m_2} \log |\bOmega_2| -  \frac{1}{m_3} \log |\bOmega_3|.
\end{align*}

Taking derivative of $q(\bOmega_1, \bOmega_2, \bOmega_3)$ with respect to $\bOmega_1$ while fixing $\bOmega_2$ and $\bOmega_3$, we have
$$
\nabla_1q(\bOmega_1, \bOmega_2, \bOmega_3) = \frac{\textrm{tr}(\bSigma_3^*\bOmega_3)\textrm{tr}(\bSigma_2^*\bOmega_2)}{m_1m_2m_3} \bSigma_1^* - \frac{1}{m_1} \bOmega_1^{-1}.
$$
Setting it as zero leads to $\bOmega_1 = {m_2m_3}[\textrm{tr}(\bSigma_3^*\bOmega_3)\textrm{tr}(\bSigma_2^*\bOmega_2)]^{-1}\bOmega_1^*$. This is indeed a minimizer of $q(\bOmega_1, \bOmega_2, \bOmega_3)$ when fixing $\bOmega_2$ and $\bOmega_3$, since the second derivative $\nabla_1^2 q(\bOmega_1, \bOmega_2, \bOmega_3) = m_1^{-1} \bOmega_1^{-1} \otimes \bOmega_1^{-1}$ is positive definite. Therefore, we have 
\begin{equation}
M_{1}(\bOmega_2, \bOmega_3) =  \frac{m_2m_3}{ \textrm{tr}(\bSigma_3^*\bOmega_3)\textrm{tr}(\bSigma_2^*\bOmega_2) } \bOmega_1^*.
\label{eqn:M1}
\end{equation} 
Therefore, $M_{1}(\bOmega_2, \bOmega_3)$ equals to the true parameter $\bOmega_1^*$ up to a constant. The computations of $M_{2}(\bOmega_1, \bOmega_3)$ and $M_{3}(\bOmega_1, \bOmega_2)$ follow from the same argument. This ends the proof of Theorem \ref{thm:population_tensor}. \hfill $\blacksquare$\\

\noindent{\bf Proof of Theorem \ref{thm:statistical_error}:} To ease the presentation, we show that \eqref{eqn:error_tensor} holds when $K=3$. The proof of the case when $K>3$ is similar. We focus on the proof of the statistical error for the sample minimization function $\hat{M}_1(\bOmega_2, \bOmega_3)$. 

By definition, $\hat{M}_1(\bOmega_2, \bOmega_3) = \argmin_{\bOmega_1} q_n(\bOmega_1, \bOmega_2, \bOmega_3) = \argmin_{\bOmega_1} L(\bOmega_1)$, where
$$
L(\bOmega_1) = \frac{1}{m_1}\textrm{tr}(\Sbb_1 \bOmega_1) - \frac{1}{m_1} \log |\bOmega_1| + \lambda_1 \|\bOmega_1\|_{1,\textrm{off}},
$$
with the sample covariance matrix 
$$
\Sbb_1 = \frac{1}{m_2m_3 n}\sum_{i=1}^n \Vb_i \Vb_i^{\top} \textrm{~with~} \Vb_i = \big[ {\cal T}_i \times \bigl\{\ind_{m_1}, \bOmega_{2}^{1/2},\bOmega_{3}^{1/2}    \bigr\} \big]_{(1)}.
$$

For some constant $H>0$, we define the set of convergence
\begin{align*} 
\mathbb{A}:= \Bigg \{\bDelta \in \mathbb R^{m_1 \times m_1}: &\bDelta  = \bDelta^{\top},   \|\bDelta\|_F = H \sqrt{\frac{(m_1+s_1)\log m_1 }{nm_2m_3 }}  \Bigg\}.
\end{align*}
The key idea is to show that 
\begin{equation}
\inf_{\bDelta \in \mathbb{A}} \left\{ L\big(M_1(\bOmega_2, \bOmega_3) + \bDelta\big) - L\big(M_1(\bOmega_2, \bOmega_3)\big)  \right\} >0,
\label{eqn:goal}
\end{equation}
with high probability. To understand it, note that the function $L\big(M_1(\bOmega_2, \bOmega_3) + \bDelta\big) - L\big(M_1(\bOmega_2, \bOmega_3)\big)$ is convex in $\bDelta$. In addition, since $\hat{M}_1(\bOmega_2, \bOmega_3)$ minimizes $L(\bOmega_1)$, we have
\begin{align*}
 L\big(\hat{M}_1  (\bOmega_2, \bOmega_3)\big) - L\big(M_1(\bOmega_2, \bOmega_3)\big)  
\le L\big(M_1(\bOmega_2, \bOmega_3)\big) - L\big(M_1(\bOmega_2, \bOmega_3)\big) = 0.
\end{align*}
If we can show \eqref{eqn:goal}, then the minimizer $\widehat{\bDelta} = \hat{M}_1(\bOmega_2, \bOmega_3) - M_{1}(\bOmega_2, \bOmega_3)$ must be within the interior of the ball defined by $\mathbb{A}$, and hence $\|\widehat{\bDelta}\|_F \le  H \sqrt{(m_1+s_1)\log m_1/(nm_2m_3 )}$. Similar technique is applied in vector-valued graphical model literature \cite{fan2009}.

To show \eqref{eqn:goal}, we first decompose $L\big(M_1(\bOmega_2, \bOmega_3) + \bDelta\big) - L\big(M_1(\bOmega_2, \bOmega_3)\big) = I_1 + I_2 + I_3$, where
\begin{eqnarray*}
I_1 &:=& \frac{1}{m_1} \textrm{tr}( \bDelta \Sbb_1  )-\frac{1}{m_1} \big\{  \log |M_1(\bOmega_2, \bOmega_3) + \bDelta |   - \log |M_1(\bOmega_2, \bOmega_3)|    \big\}, \\
I_2 &:=& \lambda_1\big \{   \|  [ M_1(\bOmega_2, \bOmega_3) + \bDelta ]_{\SSS_1} \|_1 -   \| [M_1(\bOmega_2, \bOmega_3)]_{\SSS_1}  \|_1 \big\}, \\
I_3 &:=& \lambda_1 \big\{  \| [M_1(\bOmega_2, \bOmega_3) + \bDelta]_{\SSS_1^c} \|_1 -   \| [M_1(\bOmega_2, \bOmega_3)]_{\SSS_1^c}  \|_1 \big\}.
\end{eqnarray*}
It is sufficient to show $I_1 + I_2 + I_3>0$ with high probability. To simplify the term $I_1$, we employ the Taylor expansion of $f(t) = \log |M_1(\bOmega_2, \bOmega_3) + t \bDelta|$ at $t=0$ to obtain 
\begin{eqnarray*}
&&\log |M_1(\bOmega_2, \bOmega_3) + \bDelta | -    \log |M_1(\bOmega_2, \bOmega_3)|    \\
&=&  \textrm{tr}\big\{ [M_1(\bOmega_2, \bOmega_3)]^{-1} \bDelta \big\}    - [\textrm{vec}(\bDelta)]^{\top} \left [ \int_0^1 (1-\nu) \Mb_{\nu}^{-1} \otimes \Mb_{\nu}^{-1}  \mathrm{d} \nu   \right] \textrm{vec}(\bDelta),
\end{eqnarray*}
where $\Mb_{\nu}:= M_1(\bOmega_2, \bOmega_3) + \nu \bDelta \in \mathbb R^{m_1\times m_1}$. This leads to
\begin{align*}
I_1 = &  \underbrace{\frac{1}{m_1} \textrm{tr}\big( \{\Sbb_1 - [M_1(\bOmega_2, \bOmega_3)]^{-1} \}  \bDelta  \big)}_{I_{11}}  +  \underbrace{\frac{1}{m_1} [\textrm{vec}(\bDelta)]^{\top} \left [ \int_0^1 (1-\nu) \Mb_{\nu}^{-1} \otimes  \Mb_{\nu}^{-1}  \mathrm{d} \nu   \right] \textrm{vec}(\bDelta)}_{I_{12}}.
\end{align*}

For two symmetric matrices $\Ab,\Bb$, it is easy to see that $|\textrm{tr}(\Ab\Bb)| = |\sum_{i,j} \Ab_{i,j} \Bb_{i,j}|$. Based on this observation, we decompose $I_{11}$ into two parts: those in the set $\SSS_1 = \{(i,j): [\bOmega^*_1]_{i,j} \ne 0\}$ and those not in $\SSS_1$. That is, $| I_{11} | \le I_{111} + I_{112}$, where
\begin{eqnarray*}
I_{111} &:=&  \frac{1}{m_1} \Big| \sum_{(i,j)\in \SSS_1} \big\{ \Sbb_1- [M_1(\bOmega_2, \bOmega_3)]^{-1}  \big\}_{i,j}  \bDelta_{i,j}   \Big|,\\
I_{112} &:=&  \frac{1}{m_1} \Big| \sum_{(i,j)\notin \SSS_1} \big\{ \Sbb_1- [M_1(\bOmega_2, \bOmega_3)]^{-1} \big\}_{i,j}  \bDelta_{i,j}   \Big|.
\end{eqnarray*}

{\bf Bound $I_{111}$:} For two matrices $\Ab,\Bb$ and a set $\SSS$, we have 
\begin{align*}
\Big|\sum_{(i,j) \in \SSS} \Ab_{i,j} \Bb_{i,j}\Big|  &\le \max_{i,j}|\Ab_{i,j}| \Big|\sum_{(i,j) \in \SSS}\Bb_{i,j} \Big|   \le \sqrt{|\SSS|}\max_{i,j}|\Ab_{i,j}|\|\Bb\|_F,
\end{align*}
where the second inequality is due to the Cauchy-Schwarz inequality and the fact that $\sum_{(i,j) \in \SSS}\Bb_{i,j}^2 \le \|\Bb\|_F^2$. Therefore, we have
\begin{align} \notag
I_{111} &\le \frac{\sqrt{s_1+m_1}}{m_1} \cdot \max_{i,j} \left| \left\{ \Sbb_1- [M_1(\bOmega_2, \bOmega_3)]^{-1}  \right\}_{ij} \right|  \|\bDelta\|_F \nonumber\\
&\le  C \sqrt{ \frac{(m_1+s_1)\log m_1 }{nm_1^2m_2m_3} } \|\bDelta\|_F = \frac{C H \cdot (m_1+s_1)\log m_1 }{n m_1 m_2 m_3}, \label{eqn:I111}
\end{align}
where \eqref{eqn:I111} is from Lemma \ref{lemma:sample_cov_tensor}, the definition of $M_1(\bOmega_2, \bOmega_3)$ in \eqref{eqn:M1}, and the fact that $\bDelta\in\mathbb{A}$.

{\bf Bound $I_{12}$:} For any vector $\vb \in \mathbb R^p$ and any matrix $\Ab\in \mathbb R^{p\times p}$, the variational form of Rayleigh quotients implies $\lambda_{\min}(\Ab) = \min_{\|\xb\|=1} \xb^{\top}\Ab \xb$ and hence $\lambda_{\min}(\Ab) \|\vb\|^2 \le \vb^{\top} \Ab \vb$. Setting $\vb = \textrm{vec}(\bDelta)$ and $\Ab = \int_0^1 (1-\nu) \Mb_{\nu}^{-1} \otimes  \Mb_{\nu}^{-1}  \mathrm{d} \nu$ leads to 
$$
I_{12} \ge \frac{1}{m_1}  \|\textrm{vec}(\bDelta)\|_2^2 \int_0^1 (1-\nu) \lambda_{\min} \left (\Mb_{\nu}^{-1} \otimes  \Mb_{\nu}^{-1} \right)  \mathrm{d} \nu.
$$
Moreover, by the property of Kronecker product, we have 
$$
\lambda_{\min} \left (\Mb_{\nu}^{-1} \otimes  \Mb_{\nu}^{-1} \right) = [\lambda_{\min}(\Mb_{\nu}^{-1})]^2 =  [\lambda_{\max}(\Mb_{\nu})]^{-2}.
$$
In addition, by definition, $\Mb_{\nu}= M_1(\bOmega_2, \bOmega_3) + \nu \bDelta$, and hence we have
$$
\lambda_{\max}[M_1(\bOmega_2, \bOmega_3) + \nu \bDelta] \le \lambda_{\max}[M_1(\bOmega_2, \bOmega_3)]  + \lambda_{\max}(\nu \bDelta).
$$
Therefore, we can bound $I_{12}$ from below, that is,
\begin{eqnarray*}
I_{12} &\ge& \frac{\|\textrm{vec}(\bDelta)\|_2^2}{ 2m_1}  \min_{0\le \nu \le 1} \big [\lambda_{\max}[M_1(\bOmega_2, \bOmega_3)] +     \lambda_{\max}(\nu \bDelta)   \big]^{-2} \\
&\ge& \frac{\|\textrm{vec}(\bDelta)\|_2^2}{ 2m_1} \big[\|M_1(\bOmega_2, \bOmega_3)\|_2 + \|\bDelta\|_2\big]^{-2}.
\end{eqnarray*}
On the boundary of $\mathbb{A}$, it holds that $\|\bDelta\|_2 \le \|\bDelta\|_F = o(1)$. Moreover, according to \eqref{eqn:M1}, we have 
\begin{align} 
\|M_1(\bOmega_2, \bOmega_3)\|_2 = &  \left|\frac{m_2m_3}{ \textrm{tr}(\bSigma_3^*\bOmega_3)\textrm{tr}(\bSigma_2^*\bOmega_2) } \right| \|\bOmega_1^*\|_2  
\le  \frac{100}{81} \|\bSigma_1^*\|_2 \le \frac{1.5}{C_1},
\label{eqn:M1_norm}
\end{align}
where the first inequality is due to 
\begin{eqnarray*}
\textrm{tr}(\bSigma_3^*\bOmega_3) &=& \textrm{tr}[\bSigma_3^*(\bOmega_3 - \bOmega_3^*) + \ind_{m_3}] \\ 
 &\ge& m_3 - |\textrm{tr}[\bSigma_3^*(\bOmega_3 - \bOmega_3^*)]\\
 & \ge & m_3 - \|\bSigma_3^*\|_F \|\bOmega_3 - \bOmega_3^*\|_F \\
 &\ge & m_3(1- {\alpha} \|\bSigma^*_3\|_2/{\sqrt{m_3}}) \ge 0.9 m_3,
\end{eqnarray*}
for sufficiently large $m_3$. Similarly, it holds that $\textrm{tr}(\bSigma_2^*\bOmega_2) \ge 0.9 m_2$. The second inequality in \eqref{eqn:M1_norm} is due to Condition \ref{con:eigenvalue_tensor}. This together with the fact that $\|\textrm{vec}(\bDelta)\|_2 = \|\bDelta\|_F = o(1) \le {0.5}/{C_1}$ for sufficiently large $n$ imply that
\begin{equation}
I_{12} \ge \frac{\|\textrm{vec}(\bDelta)\|_2^2}{ 2m_1} \left(\frac{C_1}{2} \right)^2 = \frac{C_1^2 H^2}{8} \cdot \frac{(m_1 + s_1) \log m_1}{n m_1 m_2 m_3},
\label{eqn:bound_I12}
\end{equation}
which dominates the term $I_{111}$ for sufficiently large $H$.

{\bf Bound $I_{2}$:} To bound $I_2$, we apply the triangle inequality and then connect the $\ell_1$ matrix norm with its Frobenius norm to obtain the final bound. Specifically, we have
\begin{align*}
|I_2| \le \lambda_1  \left \|  [ \bDelta ]_{\SSS_1} \right \|_1 & = \lambda_1  \sum_{(i,j)\in \SSS_1} |  \bDelta_{i,j} |  \\
& \le \lambda_1  \sqrt{ (s_1+m_1) \sum_{(i,j)\in \SSS_1} \bDelta_{i,j} ^2}  \\
 & \le   \lambda_1  \sqrt{ s_1+m_1}\|\bDelta\|_F, 
\end{align*}
where the first inequality is from triangle inequality, the second inequality is due to the Cauchy-Schwarz inequality by noting that $s_1 = |\SSS_1| - m_1$, and the last inequality is due to the definition of Frobenius norm. By Condition \ref{con:tuning_tensor}, $\lambda_1 \le C_2 \sqrt{\log m_1 / (nm_1^2 m_2 m_3)}$. Therefore,
$$
|I_2| \le C_2 H \cdot \frac{(m_1+s_1)\log m_1}{n m_1m_2m_3},
$$
which is dominated by $I_{12}$ for sufficiently large $H$ according to \eqref{eqn:bound_I12}.

{\bf Bound $I_{3} - |I_{112}|$:} We show $I_{3} - |I_{112}|>0$. According to \eqref{eqn:M1}, we have that $M_1(\bOmega_2, \bOmega_3)$ equals $\bOmega_1^*$ up to a non-zero coefficient. Therefore, for any entry $(i,j)\in \SSS_1^c$, we have $[M_1(\bOmega_2, \bOmega_3)]_{i,j} = 0$. This implies that
\begin{align*}
I_3 & = \lambda_1 \sum_{(i,j)\in \SSS_1^c} \big\{ \big | [M_1(\bOmega_2, \bOmega_3)]_{i,j} + \bDelta_{i,j} \big | -   |[M_1(\bOmega_2, \bOmega_3)]_{i,j} | \big\}\\
& = \lambda_1 \sum_{(i,j)\in \SSS_1^c} |\bDelta_{i,j} |.
\end{align*}
This together with the expression of $I_{112}$ and the bound in Lemma \ref{lemma:sample_cov_tensor} leads to
\begin{eqnarray*}
& & I_{3} - I_{112} \\ 
&=& \sum_{(i,j)\in \SSS_1^c}  \left\{   \lambda_1  -  m_1^{-1}\big\{ \Sbb_1- [M_1(\bOmega_2, \bOmega_3)]^{-1}  \big\}_{i,j}  \right\}   |\bDelta_{i,j} |  \\
&\ge&  \Bigg( \lambda_1  - C\sqrt{\frac{\log m_1}{nm_1^2 m_2 m_3}}\Bigg)   \sum_{(i,j)\in \SSS_1^c} |\bDelta_{i,j} | > 0,
\end{eqnarray*}
as long as $1/C_2 > C$ for some constant $C$, which is valid for sufficient small $C_2$ in Condition \ref{con:tuning_tensor}.

Combining all these bounds together, we have, for any $\bDelta \in \mathbb{A}$, with high probability, 
\begin{align*}
& L\big(M_1(\bOmega_2, \bOmega_3) + \bDelta\big) - L\big(M_1(\bOmega_2, \bOmega_3)\big)  
 \ge   I_{12} - I_{111} - |I_2| + I_3 - I_{112} > 0,
\end{align*}
which ends the proof Theorem \ref{thm:statistical_error}. \hfill $\blacksquare$\\

\noindent{\bf Proof of Theorem \ref{thm:final_error}:} We show it by connecting the one-step convergence result in Theorem \ref{thm:population_tensor} and the statistical error result in Theorem \ref{thm:statistical_error}. We show the case when $K=3$. The proof of the $K>3$ case is similar. We focus on the proof of the estimation error $\bigl\|\widehat{\bOmega}_1 - \bOmega^*_1\bigr\|_F$. 

To ease the presentation, in the following derivation we remove the superscript in the initializations $\bOmega_2^{(0)}$ and $\bOmega_3^{(0)}$ and use $\bOmega_2$ and $\bOmega_3$ instead. According to the procedure in Algorithm 1, we have
\small{ 
\begin{eqnarray*}
& & \bigl\|\widehat{\bOmega}_1 - \bOmega^*_1\bigr\|_F  = \left\| \frac{\hat{M}_1(\bOmega_2, \bOmega_3)}{\big\|\hat{M}_1(\bOmega_2, \bOmega_3)\big\|_F}  -  \frac{M_1(\bOmega_2, \bOmega_3)}{\big\| M_1(\bOmega_2, \bOmega_3)\big\|_F}   \right\|_F \\
&\le & \left\| \frac{\hat{M}_1(\bOmega_2, \bOmega_3)}{\big\|\hat{M}_1(\bOmega_2, \bOmega_3)\big\|_F}  -  \frac{M_{1}(\bOmega_2, \bOmega_3)}{\big\|\hat{M}_1(\bOmega_2, \bOmega_3)\big\|_F}   \right\|_F     +  \left\| \frac{M_{1}(\bOmega_2, \bOmega_3)}{\big\|\hat{M}_1(\bOmega_2, \bOmega_3)\big\|_F}  -  \frac{M_{1}(\bOmega_2, \bOmega_3)}{\big\|M_{1}(\bOmega_2, \bOmega_3)\big\|_F}   \right\|_F\\
&\le & \frac{2}{\big\|\hat{M}_1(\bOmega_2, \bOmega_3)\big\|_F} \big\| \hat{M}_1(\bOmega_2, \bOmega_3) - M_{1}(\bOmega_2, \bOmega_3) \big\|_F,
\end{eqnarray*}  
} 
\normalsize
where the last inequality is due to the triangle inequality $||a|-|b||\le |a-b|$ and the summation of two parts. We next bound $\big\|\hat{M}_1(\bOmega_2, \bOmega_3)\big\|_F$. By triangle inequality, 
\begin{align*}
 \big\|\hat{M}_1(\bOmega_2, \bOmega_3) \big\|_F  & \ge \|M_{1}(\bOmega_2, \bOmega_3) \|_F  - \big\|M_{1}(\bOmega_2, \bOmega_3)  - \hat{M}_1(\bOmega_2, \bOmega_3)\big\|_F \\
&  \ge 2^{-1} \|M_{1}(\bOmega_2, \bOmega_3) \|_F,
\end{align*}
since $\big\|M_{1}(\bOmega_2, \bOmega_3)  - \hat{M}_1(\bOmega_2, \bOmega_3)\big\|_F = o_P(1)$ as shown in Theorem \ref{thm:statistical_error}. Moreover, by the Cauchy-Schwarz inequality, we have 
$$
\textrm{tr}(\bSigma_2^*\bOmega_2) \le \|\bSigma_2^*\|_F \|\bOmega_2\|_F \le  m_2 \|\bSigma_2^*\|_2 \|\bOmega_2\|_2 \le 2m_2/C_1,
$$ 
due to Condition \ref{con:eigenvalue_tensor} and the fact that $\bOmega_2 \in \BB(\bOmega_2^*)$. Similarly, we have $\textrm{tr}(\bSigma_3^*\bOmega_3) \le 2m_3/C_1$. This together with the expression of $M_{1}(\bOmega_2, \bOmega_3)$ in \eqref{eqn:M1} imply that $\big\|\hat{M}_1(\bOmega_2, \bOmega_3) \big\|_F \ge C_1^2 / 4$ and hence 
\begin{align*}
\|\widehat{\bOmega}_1 - \bOmega^*_1\|_F & \le \frac{8}{C_1^2} \big\| \hat{M}_1(\bOmega_2, \bOmega_3) - M_{1}(\bOmega_2, \bOmega_3) \big\|_F   = O_P\Biggl(\sqrt{ \frac{m_1(m_1+s_1)\log m_1 }{nm_1m_2m_3} } \Biggr),
\end{align*}
according to Theorem \ref{thm:statistical_error}. This ends the proof Theorem \ref{thm:final_error}. \hfill $\blacksquare$\\

\noindent{\bf Proof of Theorem \ref{thm:maxnorm}:} We prove it by transferring the optimization problem to an equivalent primal-dual problem and then applying the convergence results of \cite{ravikumar2011} to obtain the desirable rate of convergence. 

Given the sample covariance matrix $\widehat{\Sbb}_k$ defined in Lemma \ref{lemma:sample_cov_max}, according to \eqref{eqn:one_qn}, for each $k = 1,\ldots, K$, the optimization problem has an unique solution $\widehat{\bOmega}_k$ which satisfies the following Karush-Kuhn-Tucker (KKT) conditions  
\begin{equation}
\widehat{\Sbb}_k - \widehat{\bOmega}_k  + m_k\lambda_k \widehat{\Zb}_k = 0,
\label{eqn:kkt}
\end{equation}
where $\widehat{\Zb}_k \in \mathbb R^{m_k\times m_k}$ belongs to the sub-differential of $\|\bOmega_k\|_{1,\textrm{off}}$ evaluated at $\widehat{\bOmega}_k$, that is, 
\[
    [\widehat{\Zb}_k]_{i,j}  :=
\begin{cases}
	0,& \text{if } i=j\\
	\textrm{sign}([\widehat{\bOmega}_k]_{i,j}) &  \text{if $i\ne j$ and $[\widehat{\bOmega}_k]_{i,j}\ne 0$}\\
	\in [-1,+1] &  \text{if $i\ne j$ and $[\widehat{\bOmega}_k]_{i,j} = 0$}.
\end{cases}
\]

Following \cite{ravikumar2011}, we construct the primary-dual witness solution $(\tilde{\bOmega}_k,\tilde{\Zb}_k)$ such that
$$
\tilde{\bOmega}_k := \argmin_{ \substack{ \bOmega_k \succ 0, \bOmega_k = \bOmega_k^{\top}, \\ [\bOmega_k]_{\SSS_k^c} = 0} }\left\{  \textrm{tr}\big(\widehat{\Sbb}_k \bOmega_k\big) -  \log |\bOmega_k| + m_k\lambda_k \|\bOmega_k\|_{1,\textrm{off}}   \right\},
$$
where the set $\SSS_k$ refers to the set of true non-zero edges of $\bOmega^*_k$. Therefore, by construction, the support of the dual estimator $\tilde{\bOmega}_k$ is a subset of the true support, i.e., $\textrm{supp}(\tilde{\bOmega}_k) \subseteq  \textrm{supp}(\bOmega^*_k)$. We then construct $\tilde{\Zb}_k$ as the sub-differential $\widehat{\Zb}_k$ and then for each $(i,j)\in  \SSS_k^c$, we replace $[\tilde{\Zb}_{k}]_{i,j}$ with $( [\tilde{\bOmega}_k^{-1}]_{i,j} - [\widehat{\Sbb}_k]_{i,j} ) / (m_k\lambda_k)$ to ensure that $(\tilde{\bOmega}_k,\tilde{\Zb}_k)$ satisfies the optimality condition \eqref{eqn:kkt}.

Denote $\bDelta := \tilde{\bOmega}_k - \bOmega_k^*$ and $R(\bDelta) := \tilde{\bOmega}_k^{-1} - \bOmega_k^{*-1} + \bOmega_k^{*-1}\bDelta \tilde{\bOmega}_k^{-1}$. According to Lemma 4 of \cite{ravikumar2011}, in order to show the strict dual feasibility $\tilde{\bOmega}_k = \widehat{\bOmega}_k$, it is sufficient to prove
$$
\max \bigl\{ \big\|\widehat{\Sbb}_k - \bSigma^*_k\big\|_{\infty}, \| R(\bDelta) \|_{\infty}  \bigr\} \le \frac{\alpha_k m_k\lambda_k}{8},
$$
with $\alpha_k$ defined in Condition \ref{con:IR}. As assumed in Condition \ref{con:tuning_tensor}, the tuning parameter satisfies $1/C_2 \sqrt{{\log m_k}/( n m m_k)} \le \lambda_{k} \le C_2 \sqrt{{\log m_k}/(n m m_k)}$ for some constant $C_2>0$ and hence ${\alpha_k m_k\lambda_k}/{8} \ge C_3 \sqrt{m_k \log m_k/( n m)}$ for some constant $C_3>0$.

In addition, according to Lemma \ref{lemma:sample_cov_max}, we have
$$
 \big\| \widehat{\Sbb}_k -  \bSigma_k^*  \big\|_{\infty} = O_P\left( \max_{j=1,\ldots, K}  \sqrt{ \frac{(m_j+s_j) \log m_j }{nm} } \right).
$$
Under the assumption that $s_j = O(m_j)$ for $j=1,\ldots, K$ and $m_1 \asymp m_2 \asymp \cdots \asymp m_K$, we have 
$$
 \big\| \widehat{\Sbb}_k -  \bSigma_k^* \big\|_{\infty} = O_P\left( \sqrt{\frac{m_k \log m_k}{ n m}}\right).
$$
Therefore, there exists a sufficiently small constant $C_2$ such that $\big\| \widehat{\Sbb}_k -  \bSigma_k^*  \big\|_{\infty} \le {\alpha_k m_k\lambda_k}/{8}$.

Moreover, according to Lemma 5 of \cite{ravikumar2011}, $\| R(\bDelta) \|_{\infty} \le 1.5 d_k \|\bDelta\|_{\infty}^2 \kappa_{\bSigma_k^*}^3$ as long as $\| \bDelta\|_{\infty} \le (3  \kappa_{\bSigma_k^*} d_k)^{-1}$. According to Lemma 6 of \cite{ravikumar2011}, if we can show 
\begin{align*}
r  &:= 2 \kappa_{\bGamma_k^*} \bigl(\big\| \widehat{\Sbb}_k -  \bSigma_k^*  \big\|_{\infty} + m_k\lambda_k\bigr)  \\
 & \le \min \bigg\{  \frac{1}{3 \kappa_{\bSigma_k^*} d_k },  \frac{1}{\kappa_{\bSigma_k^*}^3 \kappa_{\bGamma_k^*} d_k}  \bigg\},
\end{align*}
then we have $\|\bDelta \|_{\infty} \le r $. By Condition \ref{con:bound_kappa}, $ \kappa_{\bGamma_k^*}$ and $\kappa_{\bSigma_k^*}$ are bounded. Therefore, $\big\| \widehat{\Sbb}_k -  \bSigma_k^*  \big\|_{\infty} + m_k\lambda_k$ is in the same order of $\sqrt{m_k \log m_k/(nm)}$, which is in a smaller order of $d_k^{-1}$ by the assumption of $d_k$ in Condition \ref{con:bound_kappa}. Therefore, we have shown that $\| R(\bDelta) \|_{\infty} \le m_k\lambda_k$ for a sufficiently small constant $C_2$.

Combining above two bounds, we achieve the strict dual feasibility $\tilde{\bOmega}_k = \widehat{\bOmega}_k$. Therefore, we have $\textrm{supp}\big(\widehat{\bOmega}_k\big) \subseteq \textrm{supp}(\bOmega_k^*)$ and moreover,
$$
\big\| \widehat{\bOmega}_k -  \bOmega_k^* \big\|_{\infty} = \| \bDelta\|_{\infty}  = O_P\left(\sqrt{\frac{m_k \log m_k}{n m }}\right).
$$
This ends the proof of Theorem \ref{thm:maxnorm}.  \hfill $\blacksquare$\\

\noindent{{\bf Proof of Theorem \ref{thm: test_consistency}}:}
To ease the presentation, we show that Theorem \ref{thm: test_consistency} holds when $K=3$. The proof can easily be generalized to the case with $K>3$.

We prove it by first deriving the limiting distribution of bias-corrected sample covariance, then applying the convergence result of variance correction term in Lemma \ref{lem: Ahat_consistency} to scale the distribution into standard normal.

Lemma \ref{lem: test_expansion} gives the expression of bias-corrected sample covariance  
\begin{align} \notag
	\hat{\varrho}_{i,j} + \mu_{i,j}& = -b_{ij} \frac{ [ \bOmega_1^*]_{i,j}}{ [ \bOmega_1^*]_{i,i} [ \bOmega_1^*]_{j,j}}  \\ \notag 
	& +\frac{1}{(n-1)m_2 m_3}   \sum \limits_{l=1}^{n} \sum  \limits_{i_2=1}^{m_2}  \sum \limits_{i_3=1}^{m_3} \big ( \tilde{\xi}_{l ; i, i_2 , i_3} \tilde{\xi}_{l ; j,i_2 , i_3}  \\\notag 
	& \quad \quad \quad \quad \quad \quad\quad \quad\quad \quad\quad \quad   - \EE \tilde{\xi}_{l ; i, i_2,i_3} \tilde{\xi}_{l ; j, i_2 , i_3} \big ) \\ \label{eqn: dist_debias}
  	 &\quad  + O_P \big(  (a_{n1}^2 + a_{n1}+1) \sqrt{\frac{\log m_1}{nm_2m_3}}  +a_{n2}^2 \big) 
\end{align}
Lemma \ref{lem: lem3} (i) implies the limiting distribution of the second term in \eqref{eqn: dist_debias}
\begin{align*}
& \frac{\sum \limits_{l=1}^{n} \sum \limits_{i_2=1}^{m_2}  \sum \limits_{i_3=1}^{m_3} \big (\tilde{\xi}_{l;i, i_2, i_3} \tilde{\xi}_{l; j, i_2, i_3} - \EE \tilde{\xi}_{l;i, i_2, i_3} \tilde{\xi}_{l; j, i_2, i_3} \big ) }{\sqrt{(n-1) \| \bSigma_2^*\|_F^2 \| \bSigma_3^*\|_F^2 }} \\
&\quad\quad\quad\quad\quad\quad \rightarrow \textrm{N} \left ( 0 ;  \frac{1}{[\bOmega_1^*]_{i,i} [\bOmega_1^*]_{j,j}} + \frac{([\bOmega_1^*]_{i,j})^2}{([\bOmega_1^*]_{i,i}[\bOmega_1^*]_{j,j})^2} \right   ),
\end{align*}
in distribution as $nm_2 m_3 \rightarrow \infty$. Therefore, the limiting distribution of $\hat{\varrho}_{i,j} + \mu_{i,j}$ is
\begin{align} 
& \frac{\sqrt{n-1} m_2 m_3 ( \hat{\varrho}_{i,j} + \mu_{i,j} + b_{ij} \frac{ [ \bOmega_1^*]_{i,j}}{ [ \bOmega_1^*]_{i,i} [ \bOmega_1^*]_{j,j}} ) }{\sqrt{ \| \bSigma_2^*\|_F^2   \| \bSigma_3^*\|_F^2  }}  \rightarrow \textrm{N} \bigg ( 0 ; \frac{1}{[\bOmega_1^*]_{i,i} [\bOmega_1^*]_{j,j}} + \frac{([\bOmega_1^*]_{i,j})^2}{([\bOmega_1^*]_{i,i}[\bOmega_1^*]_{j,j})^2}   \bigg ) ,  \label{eqn: bias_dist}
\end{align}
in distribution as $nm_2 m_3 \rightarrow \infty$. 

To scale \eqref{eqn: bias_dist} into standard normal distribution under null hypothesis, an approximation of $\{[\bOmega_1^*]_{i,i}\}_{i=1}^{m_1}$ is required. 
\eqref{eqn: ind2} implies that 
\begin{align}  
\hat{\varrho}_{i,i}  = &    \frac{1}{(n-1)m_2m_3}    \sum \limits_{l=1}^{n} \sum  \limits_{i_2=1}^{m_2}  \sum \limits_{i_3=1}^{m_3}(\tilde{\xi}_{l ; j, i_2 , i_3} )^2    + O_P \big(  (a_{n1}^2 + a_{n1}) \sqrt{\frac{\log m_1}{nm_2m_3}} +a_{n2}^2 \big). \label{eqn: trans_varrho}
\end{align}
Lemma \ref{lem: lem3} (ii) implies that 
\begin{align} 
\max \limits_{i,j \in \{ 1, \ldots, m_1\}} \bigg |  &\frac{1}{(n-1)m_2m_3}   \sum \limits_{l=1}^{n} \sum  \limits_{i_2=1}^{m_2}  \sum \limits_{i_3=1}^{m_3}(\tilde{\xi}_{l ; j, i_2 , i_3} )^2   - \frac{ \tr (\bSigma_2^*)\tr (\bSigma_3^*) }{m_2m_3 [\bOmega_1^*]_{i,i}} \bigg | = O_p  \big (  \sqrt{\frac{\log m_1 }{nm/m_1}} \big ) .\label{eqn: trans_xi_tilde}
\end{align}
Combining \eqref{eqn: trans_varrho} and \eqref{eqn: trans_xi_tilde}, we have 
\begin{align*}
& \hat{\varrho}_{i,i} - \frac{ \tr (\bSigma_2^*)\tr (\bSigma_3^*) }{m_2m_3 [\bOmega_1^*]_{i,i}}  
=   O_P  \bigg ( (a_{n1}^2 + a_{n1}+1) \sqrt{\frac{\log m_1}{nm/m_1}}  +a_{n2}^2 + \sqrt{\frac{\log m_1}{nm/m_1}} \bigg ).
\end{align*}
Lemma \ref{lem: beta_consistency} ensures that $(a_{n1} , a_{n2} ) \rightarrow 0$. Therefore, we can use $ \tr (\bSigma_2^*)\tr (\bSigma_3^*)  / (m_2m_3 \hat{\varrho}_{i,i})$ to approximate $ [\bOmega_1^*]_{i,i}$. Under $H_{01, ij}$, $[\bOmega_1^*]_{i,j} = 0$ for $ i \ne j$. \eqref{eqn: bias_dist} becomes
\begin{equation} \label{eqn: bias_dist_null}
\frac{\sqrt{n-1} m_2 m_3 ( \hat{\varrho}_{i,j} + \mu_{i,j}  ) }{\sqrt{ \| \bSigma_2^*\|_F^2   \| \bSigma_3^*\|_F^2  }} \rightarrow \textrm{N} \bigg ( 0 ; \frac{1}{[\bOmega_1^*]_{i,i} [\bOmega_1^*]_{j,j} }   \bigg ) ,  
\end{equation}
Then we substitute $[\bOmega_1^*]_{i,i}$ with $ \tr (\bSigma_2^*)\tr (\bSigma_3^*)  / (m_2m_3 \hat{\varrho}_{i,i})$ and get  
\begin{equation} \label{eqn: bias_dist_sub}
 \frac{\sqrt{n-1}\tr (\bSigma_2^*)\tr (\bSigma_3^*) (\hat{\varrho}_{i,j} + \mu_{i,j})}{ \sqrt{ \| \bSigma_2^*\|_F^2   \| \bSigma_3^*\|_F^2   \hat{\varrho}_{i,i} \hat{\varrho}_{j,j}} }  \rightarrow \textrm{N} ( 0 ;1 ) 
 \end{equation}
in distribution. 
Lemma \ref{lem: Ahat_consistency} implies the convergency of variance correction term that 
\begin{equation} \label{eqn: bias_dist_var}
 \frac{\varpi^2(\tr(\bSigma_2^*))^2 (\tr(\bSigma_3^*))^2}{m_2 m_3   \|\bSigma_2^*\|_F^2  \|\bSigma_3^*\|_F^2 } \rightarrow 1 
 \end{equation}
in probability. Combining \eqref{eqn: bias_dist_sub} and \eqref{eqn: bias_dist_var}, we have
$$ \sqrt{\frac{(n-1) m_2 m_3   }{ \hat{\varrho}_{i,i} \hat{\varrho}_{j,j}}} \tau_{i,j}  \rightarrow \textrm{N} ( 0 ;1 ). $$
in distribution as $ nm_2 m_3  \rightarrow \infty$. The proof is complete.   \hfill $\blacksquare$\\

\noindent{{\bf Proof of Theorem \ref{thm: FDR_control}}:}
To ease the presentation, we show that Theorem \ref{thm: FDR_control} holds when $K=3$. The proof of the case when $K>3$ is similar.

We prove it by first construct an approximation to test statistic, then applying similar strategy in Theorem 3.1 \cite{liu2013gaussian}.

From the proof of  Lemma \ref{lem: lem3}, we have
\begin{align} \notag
 &\sum \limits_{l=1}^{n} \sum \limits_{i_2=1}^{m_2}  \sum \limits_{i_3=1}^{m_3} \tilde{\xi}_{l;i, i_2, i_3} \tilde{\xi}_{l; j, i_2, i_3} \\\label{eqn: xi_zeta}
 = & \sum \limits_{l=1}^{n-1} \sum \limits_{i_2=1}^{m_2}  \sum \limits_{i_3=1}^{m_3} \lambda_{i_2}^{(2)}\lambda_{i_3}^{(3)} \zeta_{l; i , i_2 ,i_3}\zeta_{l; j , i_2 ,i_3},
\end{align}
where 
 \[
(\zeta_{l; i , i_2 ,i_3} , \zeta_{l; j , i_2 ,i_3} )   \sim  \textrm{TN} \bigg{\{} \bm{0} ; 
  \begin{pmatrix}
    [\bOmega^*_1]_{i , i }^{-1} &    \frac{[\bOmega^*_1]_{i , j }}{[\bOmega^*_1]_{i,i }[\bOmega^*_1]_{j , j }}\\
    \frac{[\bOmega^*_1]_{i , j }}{[\bOmega^*_1]_{i,i }[\bOmega^*_1]_{j , j } } &      [\bOmega^*_1]_{j , j }^{-1}
  \end{pmatrix}  \bigg{\}}
\]
i.i.d. for $1 \le l \le n, \; 1 \le i_2 \le m_2 , \; 1 \le i_3 \le m_3$. Then, let
\begin{align}  
U_{i,j}  & = \bigg \{ \sum \limits_{l=1}^{n-1} \sum \limits_{i_2=1}^{m_2}  \sum \limits_{i_3=1}^{m_3} \lambda_{i_2}^{(2)}\lambda_{i_3}^{(3)} ( \zeta_{l; i , i_2 ,i_3}\zeta_{l; j , i_2 ,i_3}   - \EE \zeta_{l; i , i_2 ,i_3}\zeta_{l; j , i_2 ,i_3} ) \bigg \}   \frac{  ([\bOmega_1^*]_{i,i}[\bOmega_1^*]_{j,j})^{1/2} }{\sqrt{(n-1) \|\bSigma_2^*\|_F^2\|\bSigma_3^*\|_F^2}} .\label{eqn: U_ij}
\end{align}
The variance of $U_{i,j}$ is 
\begin{align} \notag
\Var (U_{i,j}) = & \bigg \{\sum \limits_{l=1}^{n-1} \sum \limits_{i_2=1}^{m_2}  \sum \limits_{i_3=1}^{m_3} (\lambda_{i_2}^{(2)})^2 (\lambda_{i_3}^{(3)} )^2 \Var \big ( \zeta_{l; i , i_2 ,i_3}\zeta_{l; j , i_2 ,i_3}   - \EE \zeta_{l; i , i_2 ,i_3}\zeta_{l; j , i_2 ,i_3} \big ) \bigg \} \frac{ ([\bOmega_1^*]_{i,i}[\bOmega_1^*]_{j,j})}{(n-1) \|\bSigma_2^*\|_F^2\|\bSigma_3^*\|_F^2}\\ \notag
= & \bigg \{ \sum \limits_{l=1}^{n-1} \sum \limits_{i_2=1}^{m_2}  \sum \limits_{i_3=1}^{m_3} (\lambda_{i_2}^{(2)})^2 (\lambda_{i_3}^{(3)} )^2 \Big (  \frac{1}{[\bOmega_1^*]_{i,i} [\bOmega_1^*]_{j,j}}    + \frac{([\bOmega_1^*]_{i,j})^2}{([\bOmega_1^*]_{i,i}[\bOmega_1^*]_{j,j})^2}      \Big )  \bigg \}\frac{ ([\bOmega_1^*]_{i,i}[\bOmega_1^*]_{j,j})}{(n-1) \|\bSigma_2^*\|_F^2\|\bSigma_3^*\|_F^2} \\ 
= & 1 + [\bOmega_1^*]_{i,j}^2([\bOmega_1^*]_{i,i}[\bOmega_1^*]_{j,j})^{-1} \label{eqn: var_Uij}
\end{align}
We next prove that $U_{i,j}$ is an approximation to test statistic. From Lemma \ref{lem: liu2013lem1} with $d=1$, together with \eqref{eqn: var_Uij}, 
\begin{align*}
& \max \limits_{1 \le i , j \le m_1} \sup \limits_{0 \le t \le 4 \sqrt{\log m_1}} \Bigg |   \frac{P \Big (|U_{i,j} |\ge t \sqrt{1 + \frac{[\bOmega_1^*]_{i,j}^2}{([\bOmega_1^*]_{i,i}[\bOmega_1^*]_{j,j}) }   } \Big )}{G(t)} -1  \Bigg |  \\
 &\quad \quad \quad \quad \quad \quad \quad \quad \quad  \quad \quad \quad \quad \quad \quad \quad \quad \quad  \le C(\log m_1 )^{-1-\epsilon},
\end{align*}
for some $\epsilon > 0$ and $G(t) = 2 - 2 \Phi (t)$. Setting $t= \sqrt {\log m_1}$ gives
\begin{equation}
\max \limits_{1 \le i,j \le m_1} |U_{i,j}| = O_P(\sqrt{\log m_1}). \label{eqn: U}
\end{equation}
Combining \eqref{eqn: xi_zeta} and \eqref{eqn: U_ij}, we have 
\begin{align*}
U_{i,j} = & \sum \limits_{l=1}^{n-1} \sum \limits_{i_2=1}^{m_2}  \sum \limits_{i_3=1}^{m_3} \bigg ( \tilde{\xi}_{l;i, i_2, i_3} \tilde{\xi}_{l; j, i_2, i_3} - \EE \tilde{\xi}_{l;i, i_2, i_3} \tilde{\xi}_{l; j, i_2, i_3} \bigg )\frac{  ([\bOmega_1^*]_{i,i}[\bOmega_1^*]_{j,j})^{1/2}}{\sqrt{(n-1) \|\bSigma_2^*\|_F^2\|\bSigma_3^*\|_F^2}}.
\end{align*}
Lemma \ref{lem: lem3} (i) implies that 
\begin{align*}
& \frac{\sum \limits_{l=1}^{n} \sum \limits_{i_2=1}^{m_2}  \sum \limits_{i_3=1}^{m_3} \big (\tilde{\xi}_{l;i, i_2, i_3} \tilde{\xi}_{l; j, i_2 , i_3} - \EE \tilde{\xi}_{l;i, i_2, i_3} \tilde{\xi}_{l; j, i_2 , i_3} \big ) }{\sqrt{(n-1) \| \bSigma_2^*\|_F^2 \| \bSigma_3^*\|_F^2 }}   \rightarrow \textrm{N} \bigg ( 0 ; \frac{1}{[\bOmega_1^*]_{i,i} [\bOmega_1^*]_{j,j}} + \frac{([\bOmega_1^*]_{i,j})^2}{([\bOmega_1^*]_{i,i}[\bOmega_1^*]_{j,j})^2}   \bigg ),
\end{align*}
in distribution as $nm_2m_3 \rightarrow \infty$. 
Therefore, the limiting distribution of $U_{i,j}$ is
\begin{equation} \label{eqn: U_dist}
U_{i,j}\rightarrow {\rm{N}} \bigg ( 0 ; 1 + \frac{([\bOmega_{1}^*]_{i,j})^2}{[\bOmega_{1}^*]_{i,i}[\bOmega_{1}^*]_{j,j}} \bigg ) .
\end{equation}
in distribution as $nm_2m_3 \rightarrow \infty$. 
Similar arguments in Theorem \ref{thm: test_consistency} indicates that 
\begin{align} 
& \sqrt\frac{(n-1) m_2m_3}{\hat{\varrho}_{i,i} \hat{\varrho}_{j,j} \varpi^2} \bigg  (\hat{\varrho}_{i,j} + \mu_{i,j} + b_{i,j} \frac{[\bOmega_{1}^*]_{i,j}}{[\bOmega_{1}^*]_{i,i}[\bOmega_{1}^*]_{j,j}} \bigg )  \rightarrow \textrm{N} \bigg ( 0 ; 1 + \frac{([\bOmega_{1}^*]_{i,j})^2}{[\bOmega_{1}^*]_{i,i}[\bOmega_{1}^*]_{j,j}} \bigg ). \label{eqn: approx_dist}
\end{align}
in distribution as $nm_2m_3 \rightarrow \infty$. 
From \eqref{eqn: U} and \eqref{eqn: approx_dist}, we can see that $U_{ij}$ and test statistic converges to the same limiting distribution. Following \eqref{eqn: U}, we have 
\begin{align*}
& \max \limits_{1 \le i < j \le m_1} \bigg |  \sqrt\frac{(n-1) m_2m_3}{\hat{\varrho}_{i,i} \hat{\varrho}_{j,j} \varpi^2} \bigg (\hat{\varrho}_{i,j} + \mu_{i,j}+ b_{i,j} \frac{[\bOmega_{1}^*]_{i,j}}{[\bOmega_{1}^*]_{i,i}[\bOmega_{1}^*]_{j,j}} \bigg )  - U_{i,j} \bigg|    = O_P (\sqrt{\log m_1})
\end{align*}
as $ n m_2 m_3 \rightarrow \infty$. The rest of the proof exactly follows Theorem 3.1 in \cite{liu2013gaussian} step by step. We skip the details. The proof is complete.    \hfill $\blacksquare$\

\section{Proof of key lemmas}
\label{sec:key_lemma}

The first key lemma establishes the rate of convergence of the difference between a sample-based quadratic form and its expectation. This new concentration result is also of independent interest.

\begin{lemma}
\label{lemma:sample_cov}
Assume i.i.d. data $\Xb, \Xb_1,\ldots, \Xb_n \in \mathbb R^{p\times q}$ follows the matrix-variate normal distribution such that $\textrm{vec}(\Xb_i) \sim \textrm{N}(\textbf{0}; \bPsi^* \otimes \bSigma^*)$ with $\bPsi^* \in \mathbb R^{q\times q}$ and $\bSigma^* \in \mathbb R^{p\times p}$. Assume that $0 < C_1 \le \lambda_{\min}(\bSigma^*) \le \lambda_{\max}(\bSigma^*) \le 1/C_1 < \infty$ and $0 < C_2 \le \lambda_{\min}(\bPsi^*) \le \lambda_{\max}(\bPsi^*) \le 1/C_2 < \infty$ for some positive constants $C_1,C_2$. For any symmetric and positive definite matrix $\bOmega \in \mathbb R^{p\times p}$, we have 
$$
\max_{i,j} \bigg\{ \frac{1}{np} \sum_{i=1}^n \Xb_i^{\top} \bOmega \Xb_i - \frac{1}{p} \EE(\Xb^{\top} \bOmega \Xb) \bigg\}_{i,j} = O_P\Bigg(\sqrt{ \frac{\log q }{np} } \Bigg).
$$
\end{lemma}

\noindent{\bf Proof of Lemma \ref{lemma:sample_cov}:} Consider a random matrix $\Xb$ following the matrix normal distribution such that $\textrm{vec}(\Xb) \sim \textrm{N}(\textbf{0}; \bPsi^* \otimes \bSigma^*)$. Let $\bLambda^* = \bPsi^{*-1}$ and $\bOmega^* = \bSigma^{*-1}$. Let $\Yb := (\bOmega^*)^{1/2} \Xb (\bLambda^*)^{1/2}$. According to the properties of matrix normal distribution \cite{gupta2000},  $\Yb$ follows a matrix normal distribution such that $\textrm{vec}(\Yb) \sim \textrm{N}(\textbf{0}; \ind_{q} \otimes \ind_{p})$, that is, all the entries of $\Yb$ are i.i.d. standard Gaussian random variables. Next we rewrite the term $\Xb^{\top} \bOmega \Xb$ by $\Yb$ and then simplify it. Simple algebra implies that 
$$
\Xb^{\top} \bOmega \Xb = (\bLambda^*)^{-1/2} \Yb^{\top} (\bOmega^*)^{-1/2} \bOmega (\bOmega^*)^{-1/2} \Yb  (\bLambda^*)^{-1/2}.
$$ 
When $\bOmega$ is symmetric and positive definite, the matrix $\Mb:=(\bOmega^*)^{-1/2} \bOmega (\bOmega^*)^{-1/2}\in \mathbb R^{p\times p}$ is also symmetric and positive definite with Cholesky decomposition $\Ub^{\top}\Ub$, where $\Ub \in \mathbb R^{p\times p}$. Therefore,
$$
\Xb^{\top} \bOmega \Xb = (\bLambda^*)^{-1/2} \Yb^{\top} \Ub^{\top}\Ub  \Yb  (\bLambda^*)^{-1/2}.
$$
Moreover, denote the column of the matrix $(\bLambda^*)^{-1/2}$ as $(\bLambda^*)^{-1/2}_{(j)}$ and denote its row as $(\bLambda^*)^{-1/2}_{i}$ for $i,j=1,\ldots,q$. Define the standard basis $\eb_i \in \mathbb R^q$ as the vector with $1$ in its $i$-th entry and $0$ in all the rest entries. The $(s,t)$-th entry of matrix $\Xb^{\top} \bOmega \Xb$ can be written as
$$
\left\{ \Xb^{\top} \bOmega \Xb \right\}_{s,t} = \eb_s^{\top}  \Xb^{\top} \bOmega \Xb \eb_{t} = (\bLambda^*)^{-1/2}_{s} \Yb^{\top} \Ub^{\top}\Ub  \Yb  (\bLambda^*)^{-1/2}_{(t)}.
$$

For the sample matrices $\Xb_1,\ldots, \Xb_n$, we apply similar transformation that $\Yb_i = (\bOmega^*)^{1/2} \Xb_i (\bLambda^*)^{1/2}$. We apply the above derivation to the sample-based quadratic term $\Xb_i^{\top} \bOmega \Xb_i$. Let $\Ab = (\ab_1,\ldots,\ab_n) \in \mathbb R^{p\times n}$ with $\ab_i =   \Ub \Yb_i (\bLambda^*)^{-1/2}_{s}  \in \mathbb R^{p}$ and $\Bb = (\bbb_1,\ldots,\bbb_n) \in \mathbb R^{p\times n}$ with $\bbb_i =   \Ub \Yb_i (\bLambda^*)^{-1/2}_{t}  \in \mathbb R^{p}$. Then we have
\begin{eqnarray}
\Big\{ \frac{1}{np} \sum_{i=1}^n \Xb_i^{\top} \bOmega \Xb_i \Big\}_{s,t}  &=& \frac{1}{np} \sum_{i=1}^n \ab_i^{\top} \bbb_i = \frac{1}{np} \sum_{i=1}^n\sum_{j=1}^p \Ab_{i,j} \Bb_{i,j} \nonumber \\
& = & \frac{1}{4np} \sum_{i=1}^n\sum_{j=1}^p \big\{ (\Ab_{i,j} + \Bb_{i,j})^2 -   (\Ab_{i,j} - \Bb_{i,j})^2 \big\}  \nonumber \\
&=& \frac{1}{4np} \left\{ \| \textrm{vec}(\Ab)+ \textrm{vec}(\Bb) \|_2^2 +  \| \textrm{vec}(\Ab)- \textrm{vec}(\Bb) \|_2^2 \right\}. \label{eqn:vecAB}
\end{eqnarray}
Next we derive the explicit form of $\textrm{vec}(\Ab)$ and $\textrm{vec}(\Bb)$ in \eqref{eqn:vecAB}. Remind that $(\bLambda^*)^{-1/2}_{s}$ is a vector of length $q$. By the property of matrix products, we can rewrite $\ab_i = [(\bLambda^*)^{-1/2}_{s} \otimes \Ub] \textrm{vec}(\Yb_i)$, where $\otimes$ is the Kronecker product. Therefore, we have 
\begin{eqnarray*}
\textrm{vec}(\Ab)  &=& \big[ \ind_n \otimes (\bLambda^*)^{-1/2}_{s} \otimes \Ub  \big] \tb :=  \Qb_1 \tb,\\
\textrm{vec}(\Bb)  &=& \big[ \ind_n \otimes (\bLambda^*)^{-1/2}_{t} \otimes \Ub  \big] \tb :=  \Qb_2 \tb,
\end{eqnarray*}
where $\tb = \big\{ [\textrm{vec}(\Yb_1)]^{\top}, \ldots, [\textrm{vec}(\Yb_n)]^{\top} \big\}^{\top} \in \mathbb R^{npq}$ is a vector with $npq$ i.i.d. standard normal entries. Here $\Qb_1 := \ind_n \otimes (\bLambda^*)^{-1/2}_{s} \otimes \Ub$ and $\Qb_2 := \ind_n \otimes (\bLambda^*)^{-1/2}_{t} \otimes \Ub$ with $\Qb_1,\Qb_2 \in \mathbb R^{np \times npq}$. By the property of multivariate normal distribution, we have
\begin{eqnarray*}
\textrm{vec}(\Ab)+ \textrm{vec}(\Bb) &\sim& \textrm{N}\big(0; (\Qb_1 + \Qb_2)(\Qb_1+\Qb_2)^{\top}\big):= \textrm{N}(0;\Hb_1),\\
\textrm{vec}(\Ab)- \textrm{vec}(\Bb) &\sim& \textrm{N}\big(0; (\Qb_1 - \Qb_2)(\Qb_1-\Qb_2)^{\top}\big):= \textrm{N}(0;\Hb_2).
\end{eqnarray*}
Next, we bound the spectral norm of two matrices $\Hb_1$ and $\Hb_2$. By the property of matrix norm and the fact that one matrix and its transpose matrix have the same spectral norm, we have
$$
\|\Hb_1\|_2 \le \|\Qb_1\Qb_1^{\top}\|_2 + 2\|\Qb_1\Qb_2^{\top}\|_2 + \|\Qb_2\Qb_2^{\top}\|_2,
$$
then we bound each of these three terms individually. According to the definition of $\Qb_1$ and the property of matrix Kronecker products, we have
\begin{eqnarray*}
\Qb_1\Qb_1^{\top} &=&  \big[ \ind_n \otimes (\bLambda^*)^{-1/2}_{s} \otimes \Ub \big]  \big[  \ind_n \otimes (\bLambda^*)^{-1/2}_{s} \otimes \Ub \big]^{\top}\\
&=&  \ind_n \otimes (\bLambda^*)^{-1/2}_{s}[(\bLambda^*)^{-1/2}_{s}]^{\top} \otimes \Mb,
\end{eqnarray*}
where the last equality is due to the fact that $(\Cb_1 \otimes \Cb_2)^{\top} = \Cb_1^{\top} \otimes \Cb_2^{\top}$ and $(\Cb_1 \otimes \Cb_2)(\Cb_3 \otimes \Cb_4) = (\Cb_1\Cb_3) \otimes (\Cb_2\Cb_4)$ for any matrices $\Cb_1,\ldots, \Cb_4$ such that the matrix multiplications $\Cb_1\Cb_3$ and $\Cb_2\Cb_4$ are valid. Moreover, we also use the Cholesky decomposition of $\Mb$, i.e., $\Mb = \Ub^{\top}\Ub$. Remind that $(\bLambda^*)^{-1/2}_{s}[(\bLambda^*)^{-1/2}_{s}]^{\top} \in \mathbb R$, therefore, the spectral norm $\Qb_1\Qb_1^{\top}$ can be written as
\begin{eqnarray*}
\|\Qb_1\Qb_1^{\top} \|_2 &=& \big|(\bLambda^*)^{-1/2}_{s}[(\bLambda^*)^{-1/2}_{s}]^{\top}\big| \cdot \|\ind_n\|_2 \|\Mb\|_2\\
&\le & \|\bPsi^*\|_2 \|\Mb\|_2 \le \left(1+\alpha/C_1\right)/C_2.
\end{eqnarray*}
Here the first inequality is because $\| \ind_n \|_2 = 1$ and 
\begin{eqnarray*}
\big|(\bLambda^*)^{-1/2}_{s}[(\bLambda^*)^{-1/2}_{s}]^{\top} \big| &=& \big\| [(\bLambda^*)^{-1/2}_{s}]^{\top}(\bLambda^*)^{-1/2}_{s} \big\|_2 \le \max_{j} \big\|[(\bPsi^*)^{1/2}_{j}]^{\top}(\bPsi^*)^{1/2}_{j}\big\|_2\\
&\le&  \Big\|\sum_{j=1}^q [(\bPsi^*)^{1/2}_{j}]^{\top}(\bPsi^*)^{1/2}_{j}\Big\|_2 = \|\bPsi^*\|_2,
\end{eqnarray*}
and the second inequality is because $\|\bPsi^*\|_2 \le 1/C_2$ and 
\begin{eqnarray*}
\|\Mb\|_2 &=& \left\|(\bOmega^*)^{-1/2} \bOmega (\bOmega^*)^{-1/2}\right\|_2 = \|(\bOmega^*)^{-1/2} (\bOmega - \bOmega^*) (\bOmega^*)^{-1/2}  + \ind_p \|_2 \\
&\le& \|(\bOmega^*)^{-1/2} \|_2^2 \|\bOmega - \bOmega^*\|_2 + 1 \le \|\bSigma^*\|_2  \|\bOmega - \bOmega^*\|_F + 1 \le 1+\alpha/C_1.
\end{eqnarray*}

Similarly, we have $\|\Qb_2\Qb_2^{\top} \|_2\le \left(1+\alpha/C_1\right)/C_2$. For $\|\Qb_1\Qb_2^{\top}\|_2$, similar arguments imply that 
\begin{eqnarray*}
\Qb_1\Qb_2^{\top} = \ind_n \otimes (\bLambda^*)^{-1/2}_{s}[(\bLambda^*)^{-1/2}_{t}]^{\top} \otimes \Mb,
\end{eqnarray*}
and hence its spectral norm is bounded as
\begin{eqnarray*}
\|\Qb_1\Qb_2^{\top} \|_2 &=& |(\bLambda^*)^{-1/2}_{s}[(\bLambda^*)^{-1/2}_{t}]^{\top}| \cdot \|\ind_n\|_2 \|\Mb\|_2\\
&\le & \|\bPsi^*\|_2 \|\Mb\|_2 \le \left(1+\alpha/C_1\right)/C_2,
\end{eqnarray*}
where the first inequality is because the above derivation and the Cauchy-Schwarz inequality. Specifically, let $\bPsi^* = (\bPsi^*_{i,j})$, we have 
\begin{eqnarray*}
&&|(\bLambda^*)^{-1/2}_{s}[(\bLambda^*)^{-1/2}_{t}]^{\top}|  =  \sqrt{(\bPsi^*)_s [(\bPsi^*)_t]^{\top}} = \big[\sum_{j=1}^q \bPsi^*_{s,j}\bPsi^*_{t,j} \big]^{1/2} \\ 
&\le& \Big\{ (\sum_{j=1}^q \bPsi_{s,j}^{*2})(\sum_{j=1}^q \bPsi_{t,j}^{*2}) \Big\}^{1/4}  \le \sqrt{\|\bPsi^*\|_2 \|\bPsi^*\|_2} \le  C_2^{-1}.
\end{eqnarray*}
Applying the same techniques to $\|\Hb_2\|_2$, we have
\begin{eqnarray}
\|\Hb_1\|_2 &\le& 4\left(1+\alpha/C_1\right)/C_2, \label{eqn:boundH1}\\
\|\Hb_2\|_2 &\le& 4\left(1+\alpha/C_1\right)/C_2.  \label{eqn:boundH2}
\end{eqnarray}

Next, we apply Lemma \ref{lemma:NW11} to bound the $(s,t)$-th entry of the differential matrix between the sample-based term and its expectation. Denote $\rho_{s,t} := [ p^{-1} \EE(\Xb^{\top} \bOmega \Xb)]_{s,t}$. According to the derivation in \eqref{eqn:vecAB}, we have
\begin{eqnarray}
&& \left\{ \frac{1}{np} \sum_{i=1}^n \Xb_i^{\top} \bOmega \Xb_i - \frac{1}{p} \EE(\Xb^{\top} \bOmega \Xb) \right\}_{s,t} \nonumber \\
& = &  \bigg[ \frac{1}{4np} \sum_{i,j} (a_{ij} + b_{ij})^2 - \frac{\bDelta_{s,t} + \rho_{s,t}}{2} \bigg] - \bigg[ \frac{1}{4np} \sum_{i,j} (a_{ij} - b_{ij})^2 - \frac{\bDelta_{s,t} - \rho_{s,t}}{2} \bigg], \label{eqn:twobound}
\end{eqnarray}
where $\bDelta_{s,t}$ is defined as
$$
\bDelta_{s,t} :=\EE \Big\{ (4np)^{-1} \sum_{i,j} [ (a_{ij} + b_{ij})^2 +   (a_{ij} - b_{ij})^2 ] \Big\}.
$$
Moreover, according to the definition of $\rho_{s,t}$ and the fact in \eqref{eqn:vecAB}, we have $\EE\{(4np)^{-1} \sum_{i=1}^n\sum_{j=1}^p [ (a_{ij} + b_{ij})^2 -   (a_{ij} - b_{ij})^2 ]\} = \rho_{s,t}$. Therefore, we have  
\begin{eqnarray}
\EE\big\{(4np)^{-1} \sum_{i,j} (a_{ij} + b_{ij})^2\big\} &=& \frac{\bDelta_{s,t} + \rho_{s,t}}{2},\label{eqn:expect1}\\
\EE\big\{(4np)^{-1} \sum_{i,j} (a_{ij} - b_{ij})^2\big\} &=& \frac{\bDelta_{s,t} - \rho_{s,t}}{2}. \label{eqn:expect2}
\end{eqnarray}
Therefore, \eqref{eqn:twobound} implies that, for any $\delta>0$,
\small{
\begin{eqnarray*}
&&\mathbb P \Big [ \Big| \Big\{ \frac{1}{np} \sum_{i=1}^n \Xb_i^{\top} \bOmega \Xb_i - \frac{1}{p} \EE(\Xb^{\top} \bOmega \Xb) \Big\}_{s,t} \Big| \ge \delta  \Big] \le \\
&&\underbrace{  \mathbb P \Big[  \Big| \frac{1}{np} \sum_{i,j} (a_{ij} + b_{ij})^2 - 2(\bDelta_{s,t} + \rho_{s,t}) \Big| > 2\delta\Big] }_{I_1}+ \underbrace{  \mathbb P \Big[  \Big| \frac{1}{np} \sum_{i,j} (a_{ij} - b_{ij})^2 - 2(\bDelta_{s,t} - \rho_{s,t}) \Big| > 2\delta\Big] }_{I_2}.
\end{eqnarray*}
}
\normalsize
Remind that $\sum_{i=1}^n\sum_{j=1}^p (a_{ij} + b_{ij})^2 = \textrm{vec}(\Ab)+ \textrm{vec}(\Bb) \sim \textrm{N}(0;\Hb_1)$ and $\sum_{i=1}^n\sum_{j=1}^p (a_{ij} - b_{ij})^2 = \textrm{vec}(\Ab)- \textrm{vec}(\Bb) \sim \textrm{N}(0;\Hb_2)$. According to \eqref{eqn:expect1} and \eqref{eqn:expect2}, we apply Lemma \ref{lemma:NW11} to obtain
\begin{eqnarray*}
I_1 &\le& 2 \exp\bigg\{ -\frac{np}{2} \bigg(\frac{\delta}{2\|\Hb_1\|_2} - \frac{2}{\sqrt{np}} \bigg)^2 \bigg\}   + 2 \exp(-np/2),\\
I_2 &\le& 2 \exp\bigg\{ -\frac{np}{2} \bigg(\frac{\delta}{2\|\Hb_2\|_2} - \frac{2}{\sqrt{np}} \bigg)^2 \bigg\}   + 2 \exp(-np/2).
\end{eqnarray*}

Finally, in order to derive the convergence rate of the maximal difference over all index $(s,t)$, we employ the max sum inequality. That is, for random variables $x_1,\ldots,x_n$, we have $\mathbb P(\max_{i} x_i \ge t) \le \sum_{i=1}^n \mathbb P(x_i \ge t) \le n  \max_{i} \mathbb P(x_i \ge t)$. This together with \eqref{eqn:boundH1} and \eqref{eqn:boundH2} imply that 
\begin{eqnarray}
&&\mathbb P \bigg [ \max_{(s,t)}\bigg\{ \frac{1}{np} \sum_{i=1}^n \Xb_i^{\top} \bOmega \Xb_i - \frac{1}{p} \EE(\Xb^{\top} \bOmega \Xb) \bigg\}_{s,t}  \ge \delta  \bigg] \nonumber\\
&\le& 4 q^2 \exp\bigg\{ -\frac{np}{2} \bigg[\frac{\delta C_1C_2}{8(C_1+\alpha)} - \frac{2}{\sqrt{np}} \bigg]^2 \bigg\}   + 4q^2 \exp(-np/2). \label{eqn:bound_final}
\end{eqnarray}
Let $\delta = 8(C_1+\alpha)(C_1C_2)^{-1} [ 4\sqrt{\log q/(np)} + 3(np)^{-1/2}]$ in \eqref{eqn:bound_final} which satisfies the condition in Lemma \ref{lemma:NW11} since $\delta > 2(np)^{-1/2}$ when $q$ is sufficiently large. Therefore, we obtain the desirable conclusion that, with high probability,
$$
\max_{(s,t)}\left\{ \frac{1}{np} \sum_{i=1}^n \Xb_i^{\top} \bOmega \Xb_i - \frac{1}{p} \EE(\Xb^{\top} \bOmega \Xb) \right\}_{s,t} =O_P\left(\sqrt{ \frac{\log q }{np} } \right).
$$
This ends the proof of Lemma \ref{lemma:sample_cov}. \hfill $\blacksquare$\\

\begin{lemma}
\label{lemma:sample_cov_tensor}
Assume i.i.d. tensor data ${\cal T}, {\cal T}_1,\ldots, {\cal T}_n \in \mathbb R^{m_1 \times m_2 \times \cdots \times m_K}$ follows the tensor normal distribution $\textrm{TN}({\bf0}; \bSigma_1^*, \ldots, \bSigma_K^*)$. Assume Condition \ref{con:eigenvalue_tensor} holds. For any symmetric and positive definite matrices $\bOmega_j \in \mathbb R^{m_j\times m_j}, j\ne k$, we have
$$
\mathbb E [\Sbb_k] =  \frac{ m_k [\prod_{j\ne k} \textrm{tr}(\bSigma_j^*\bOmega_j)] }{m}\bSigma_k^*,
$$
for $\Sbb_k = \frac{m_k}{nm} \sum_{i=1}^n \Vb_i \Vb_i^{\top}$ with $\Vb_i = \big[ {\cal T}_i \times \{\bOmega_1^{1/2},\ldots,\bOmega_{k-1}^{1/2}, \ind_{m_k}, \bOmega_{k+1}^{1/2},\ldots,\bOmega_{K}^{1/2}\} \big]_{(k)}$ and $m = \prod_{k=1}^K m_k$. Moreover, we have
\begin{equation}
\max_{s,t} \left\{ \Sbb_k -  \frac{ m_k [\prod_{j\ne k} \textrm{tr}(\bSigma_j^*\bOmega_j)] }{m}\bSigma_k^* \right\}_{s,t} = O_P\left(\sqrt{ \frac{m_k \log m_k }{nm} } \right).\label{eqn:sampletensor}
\end{equation}
\end{lemma}

\noindent{\bf Proof of Lemma \ref{lemma:sample_cov_tensor}:} The proof follows by carefully examining the distribution of $\Vb_i$ and then applying Lemma \ref{lemma:sample_cov}. We only show the case with $K=3$ and $k=1$. The extension to a general $K$ follows similarly. 

According to the property of mode-$k$ tensor multiplication, we have $\Vb_i = \left[ {\cal T}_i \right]_{(1)}  (\bOmega_{3}^{1/2} \otimes \bOmega_2^{1/2})$, and hence 
\begin{eqnarray*}
\Sbb_1 &=& \frac{1}{nm_2m_3} \sum_{i=1}^n   \left[ {\cal T}_i \right]_{(1)} (\bOmega_{3}^{1/2} \otimes \bOmega_2^{1/2}) (\bOmega_{3}^{1/2} \otimes \bOmega_2^{1/2})   \left[ {\cal T}_i \right]_{(1)}^{\top} \\
&=& \frac{1}{nm_2m_3} \sum_{i=1}^n   \left[ {\cal T}_i \right]_{(1)} (\bOmega_{3} \otimes \bOmega_2)  \left[ {\cal T}_i \right]_{(1)}^{\top}.
\end{eqnarray*}
When tensor ${\cal T}_i \sim \textrm{TN}({\bf0}; \bSigma_1^*, \bSigma_2^*, \bSigma_3^*)$, the property of mode-$k$ tensor multiplication shown in Proposition 2.1 in \cite{hoff2011} implies that
$$
[{\cal T}_i]_{(1)} \in \mathbb R^{m_1 \times (m_2m_3)} \sim \textrm{MN}({\bf 0}; \bSigma_1^*,  \bSigma_3^* \otimes \bSigma_2^*),
$$
where $\textrm{MN}({\bf 0}; \bSigma_1^*,  \bSigma_3^* \otimes \bSigma_2^*)$ is the matrix-variate normal \cite{dawid1981} such that the row covariance matrix of $[{\cal T}_i]_{(1)}$ is $\bSigma_1^*$ and the column covariance matrix of $[{\cal T}_i]_{(1)}$ is $\bSigma_3^* \otimes \bSigma_2^*$. Therefore, in order to show \eqref{eqn:sampletensor}, according to Lemma \ref{lemma:sample_cov}, it is sufficient to show
\begin{equation}
\mathbb E [\Sbb_1] =  \frac{ \textrm{tr}(\bSigma_3^*\bOmega_3)\textrm{tr}(\bSigma_2^*\bOmega_2) }{m_2m_3}\bSigma_1^*.
\label{eqn:toshow}
\end{equation}

According to the distribution of $[{\cal T}_i]_{(1)}$, we have 
$$
\Vb_i \sim \textrm{MN}\left({\bf 0}; \bSigma_1^*,  (\bOmega_{3}^{1/2} \otimes \bOmega_2^{1/2}) (\bSigma_3^* \otimes \bSigma_2^*) (\bOmega_{3}^{1/2} \otimes \bOmega_2^{1/2})\right),
$$ 
and hence
$$
\Vb_i^{\top} \sim \textrm{MN}\left({\bf 0}; (\bOmega_{3}^{1/2} \otimes \bOmega_2^{1/2}) (\bSigma_3^* \otimes \bSigma_2^*) (\bOmega_{3}^{1/2} \otimes \bOmega_2^{1/2}), \bSigma_1^*\right).
$$
Therefore, according to Lemma \ref{lemma:compute_cov}, we have
$$
\mathbb E[\Vb_i \Vb_i^{\top}] = \bSigma_1^* \textrm{tr}\left[  (\bOmega_{3} \otimes \bOmega_2) (\bSigma_3^* \otimes \bSigma_2^*)  \right] = \bSigma_1^* \textrm{tr}(\bSigma_3^*\bOmega_{3}) \textrm{tr}(\bSigma_2^*\bOmega_2),
$$
which implies \eqref{eqn:toshow} according to the definition of $\Sbb_1$. Finally, applying Lemma \ref{lemma:sample_cov} to $\Sbb_1$ leads to the desirable result. This ends the proof of Lemma \ref{lemma:sample_cov_tensor}. \hfill $\blacksquare$\\

The following lemma establishes the rate of convergence of the sample covariance matrix in max norm.

\begin{lemma}
\label{lemma:sample_cov_max}
Assume i.i.d. tensor data ${\cal T}, {\cal T}_1,\ldots, {\cal T}_n \in \mathbb R^{m_1 \times m_2 \times \cdots \times m_K}$ follows the tensor normal distribution $\textrm{TN}({\bf0}; \bSigma_1^*,\cdots,\bSigma_K^*)$, and assume Condition \ref{con:eigenvalue_tensor} holds. Let $\widehat{\bOmega}_j \in \mathbb R^{m_j\times m_j}, j\ne k$, be the estimated precision matrix from Algorithm \ref{alg:tlasso} with iteration number $T=1$. Denote the $k$-th sample covariance matrix as
$$
\widehat{\Sbb}_k = \frac{m_k}{nm} \sum_{i=1}^n \widehat{\Vb}_i \widehat{\Vb}_i^{\top},
$$
with $m = \prod_{k=1}^K m_k$ and $\widehat{\Vb}_i := \big[ {\cal T}_i \times \bigl\{\widehat{\bOmega}_1^{1/2},\ldots,\widehat{\bOmega}_{k-1}^{1/2}, \ind_{m_k}, \widehat{\bOmega}_{k+1}^{1/2},\ldots,\widehat{\bOmega}_{K}^{1/2} \bigr\} \big]_{(k)}$.
We have
\begin{equation}
\max_{s,t} \big[ \widehat{\Sbb}_k -  \bSigma_k^* \big]_{s,t} = O_P\left( \max_{j=1,\ldots, K}  \sqrt{ \frac{(m_j+s_j) \log m_j }{nm} } \right).\label{eqn:max_norm_sample_cov}
\end{equation}
\end{lemma}

\noindent{\bf Proof of Lemma \ref{lemma:sample_cov_max}:} The proof follows by decomposing the $\widehat{\Sbb}_k -  \bSigma_k^*$ into two parts and then applying Lemma \ref{lemma:sample_cov_tensor} and Theorem \ref{thm:final_error} for each part to bound the final error.

Note that the triangle inequality implies that
$$
\big\| \widehat{\Sbb}_k -  \bSigma_k^* \big\|_{\infty} \le   \underbrace{\bigg\| \widehat{\Sbb}_k - \frac{ m_k [\prod_{j\ne k} \textrm{tr}(\bSigma_j^* \widehat{\bOmega}_j)] }{m}\bSigma_k^*  \bigg\|_{\infty}}_{I_1}  + \underbrace{ \bigg \| \frac{ m_k [\prod_{j\ne k} \textrm{tr}(\bSigma_j^* \widehat{\bOmega}_j)] }{m}\bSigma_k^*  - \bSigma_k^* \bigg\|_{\infty}}_{I_2}.
$$
Note that here the covariance matrix $\widehat{\Sbb}_k$ is constructed based on the estimators $\widehat{\bOmega}_j, j\ne k$. According to \eqref{eqn:sampletensor} in Lemma \ref{lemma:sample_cov_tensor}, we have
$$
I_1 = O_P\left(\sqrt{ \frac{m_k \log m_k }{nm} } \right).
$$
The remainder part is to bound the error $I_2$. Note that $\textrm{tr}(\bSigma_j^* \bOmega_j^*) = \textrm{tr}(\ind_{m_j}) = m_j$. Therefore,
$$
I_2 = \underbrace{\Big| \frac{m_k}{m} \Big[   \prod_{j\ne k} \textrm{tr}(\bSigma_j^* \widehat{\bOmega}_j) - \prod_{j\ne k} \textrm{tr}(\bSigma_j^* \bOmega_j^*)   \Big]     \Big| }_{I_3} \|\bSigma_k^*\|_{\infty}.
$$
Given that $\|\bSigma_k^*\|_{\infty} = O_P(1)$, it is sufficient to bound the coefficient $I_3$. We only demonstrate the proofs with $K=3$ and $k=1$. The extension to a general $K$ follows similarly. In this case, we have
\begin{eqnarray*}
I_3 &=&  \frac{m_1}{m} \left|  \textrm{tr}(\bSigma_2^* \widehat{\bOmega}_2) \textrm{tr}(\bSigma_3^* \widehat{\bOmega}_3) -  \textrm{tr}(\bSigma_2^* \bOmega^*_2) \textrm{tr}(\bSigma_3^* \bOmega^*_3)  \right|\\
&\le&   \bigg|  \frac{ \textrm{tr}(\bSigma_2^* \widehat{\bOmega}_2) \textrm{tr}[\bSigma_3^* ( \widehat{\bOmega}_3 - \bOmega^*_3)]}{m_2m_3} \bigg| + \bigg|  \frac{ \textrm{tr}[\bSigma_2^* ( \widehat{\bOmega}_2 - \bOmega^*_2)] \textrm{tr}(\bSigma_3^*\bOmega^*_3)}{m_2m_3} \bigg|.
\end{eqnarray*}
According to the proof of Theorem \ref{thm:final_error}, we have $C_1 \le \textrm{tr}(\bSigma_j^*\bOmega_j) /m_j \le 1/C_1$ for any $j=1,\ldots,K$ and some constant $C_1>0$. Moreover, we have $\textrm{tr}(\bSigma_3^*\bOmega^*_3) = m_3$. Therefore, we have
$$
I_3 \le \bigg|  \frac{  \textrm{tr}[\bSigma_3^* ( \widehat{\bOmega}_3 - \bOmega^*_3)]}{m_3} \bigg| + \bigg|  \frac{ \textrm{tr}[\bSigma_2^* ( \widehat{\bOmega}_2 - \bOmega^*_2)] }{m_2} \bigg|.
$$
Here $\textrm{tr}[\bSigma_j^* ( \widehat{\bOmega}_j - \bOmega^*_j)] \le \|\bSigma_j^*\|_F \big\| \widehat{\bOmega}_j - \bOmega^*_j\big\|_F \le \sqrt{m_j} \|\bSigma_j^*\|_2 \big\| \widehat{\bOmega}_j - \bOmega^*_j\big\|_F$. According to Condition \ref{con:eigenvalue_tensor}, $\|\bSigma_j^*\|_2 = O_P(1)$. This together with Theorem \ref{thm:final_error} implies that
$$
I_3 = O_P \left( \sqrt{\frac{(m_3+s_3) \log m_3}{ nm}} + \sqrt{\frac{(m_2+s_2) \log m_2}{ nm}} \right). 
$$
By generalizing it to a general $K$ and $k$, we have that
$$
I_3 = O_P \left( \max_{j\ne k}\sqrt{\frac{(m_j+s_j) \log m_j}{ nm}} \right),
$$
and hence
$$
\big\| \widehat{\Sbb}_k -  \bSigma_k^* \big\|_{\infty} = O_P\left( \sqrt{ \frac{m_k \log m_k }{nm} }  +  \max_{j\ne k}\sqrt{\frac{(m_j+s_j) \log m_j}{ nm}} \right),
$$
which leads to the desirable result. This ends the proof of Lemma \ref{lemma:sample_cov_max}.  \hfill $\blacksquare$\\

\begin{lemma} \label{lem: test_expansion}
Assume i.i.d. tensor data ${\cal T}, {\cal T}_1,\ldots, {\cal T}_n \in \mathbb R^{m_1 \times m_2 \times \cdots \times m_K}$ follows the tensor normal distribution $\textrm{TN}({\bf0}; \bSigma_1^*, \ldots, \bSigma_K^*)$. $\bOmega_k^*$ is the inverse of $\bSigma_k^*$. Let 
\begin{equation} \label{eqn: xi_tilde}
\tilde{\xi}_{l ; i_1, \ldots , i_K} = {\cal T }_{l ; i_1, \ldots , i_K} - \bar{{\cal T }}_{ i_1, \ldots  , i_K} - ( {\cal T }_{l ; - i_1, \ldots  , i_K} - \bar{{\cal T }}_{ - i_1, \ldots , i_K} )^\top \btheta_{i_1}
\end{equation} 
and 
$$
[\Xi_1]_{ii ; \xi} = \frac{m_1}{(n-1) m} \sum \limits_{l=1}^{n} \sum \limits_{i_2=1}^{m_2}  \cdots \sum \limits_{i_K=1}^{m_K} ( \hat{\xi}_{l;i, \ldots, i_K} )^2,
$$
where $\bm{\theta}_{i_1}$ and $\hat{\xi}_{l;i, \ldots, i_K}$ are defined in \S\ref{sec: construct}.
Assume Condition \ref{con:eigenvalue_tensor} and $m_1 \log m_1 = o(nm)$ hold. 
Furthermore, suppose 
\begin{equation} \label{eqn: xi_approx}
\max \limits_{i \in \{1 ,\ldots , m_1\}}\|\hat{\btheta}_{i} - \btheta_{i} \|_1 =O(a_{n1}) \text{ , and  } \max \limits_{i \in \{1 ,\ldots , m_1\}} \|\hat{\btheta}_{i} - \btheta_{i} \|_2 =O(a_{n2})
\end{equation}
with $(a_{n1} , a_{n2}) \rightarrow 0$, where $\hat{\bm{\theta}}_{i_1}$ is defined in \S\ref{sec: construct}. Then $\hat{\varrho}_{i,j} + \mu_{i,j}$ can be expressed as 
\begin{align} \notag
	 \hat{\varrho}_{i,j} + \mu_{i,j}& = -b_{ij} \frac{ [ \bOmega_1^*]_{i,j}}{ [ \bOmega_1^*]_{i,i} [ \bOmega_1^*]_{j,j}}  +\frac{m_1}{(n-1)m}   \sum \limits_{l=1}^{n} \sum  \limits_{i_2=1}^{m_2} \cdots  \sum \limits_{i_K=1}^{m_K} \big ( \tilde{\xi}_{l ; i, \ldots , i_K} \tilde{\xi}_{l ; j,\ldots , i_K}   - \EE \tilde{\xi}_{l ; i, \ldots , i_K} \tilde{\xi}_{l ; j, \ldots , i_K} \big ) \\
  	 &\quad \quad \quad \quad \quad + O_P \big(  (a_{n1}^2 + a_{n1}+1) \sqrt{\frac{\log m_1}{nm/m_1}}  +a_{n2}^2 \big) 
\end{align}
where 
$$b_{ij} = [ \bOmega_1^*]_{i,i} [\Xi_1]_{ii;\xi}+ [ \bOmega_1^*]_{j,j} [\Xi_1]_{jj;\xi} - m_1 \prod_{k=\{2, \ldots , K\}}^K \tr (\bSigma_k) / m .$$ 
\end{lemma}

\noindent{{\bf Proof of Lemma \ref{lem: test_expansion}}:}
To ease the presentation, we show that Lemma \ref{lem: test_expansion} holds when $K=3$. The proof can easily be generalized to the case with $K>3$.

Our proof strategy is first transforming $\hat{\varrho}_{i,j}$ in terms of $\tilde{\xi}_{l ; i_1, i_2 , i_3}$, then applying the convergency results of $\tilde{\xi}_{l ; i_1, i_2 , i_3}$ in Lemma \ref{lem: lem1} \& \ref{lem: lem2} to get the desired approximation of $\hat{\varrho}_{i,j} + \mu_{i,j}$.

Recall the definition of the residuals $\hat{\xi}_{l ; i_1, i_2 , i_3}$, together with \eqref{eqn: xi_tilde}, 
$$\hat{\xi}_{l ; i_1, i_2 , i_3} = \tilde{\xi}_{l ; i_1, i_2 ,i_3}  - ( {\cal T }_{l ; - i_1, i_2 ,i_3} - \bar{{\cal T }}_{ - i_1, i_2 , i_3} )^\top (\hat{\btheta}_{i_1} - \btheta_{i_1})$$ 
Therefore, 
\begin{align} \notag
\hat{\xi}_{l ; i, i_2 , i_3} \hat{\xi}_{l ; j, i_2 , i_3} = & \;\tilde{\xi}_{l ; i, i_2 , i_3} \tilde{\xi}_{l ; j, i_2 , i_3}   \\\notag
& - \underbrace{\tilde{\xi}_{l ; i, i_2 , i_3}  ( {\cal T }_{l ; - j, i_2 , i_3} - \bar{{\cal T }}_{ - j, i_2 , i_3} )^\top (\hat{\btheta}_{j} - \btheta_{j})}_{\bf{I}_{1}} \\\notag
& - \underbrace{\tilde{\xi}_{l ; j, i_2 ,i_3}  ( {\cal T }_{l ; - i, i_2 , i_3} - \bar{{\cal T }}_{ - i, i_2 , i_3} )^\top (\hat{\btheta}_{i} - \btheta_{i})}_{\bf{I}_2} \\
& + \underbrace{(\hat{\btheta}_{i} - \btheta_{i})^\top ( {\cal T }_{l ; - i, i_2 , i_3} - \bar{{\cal T }}_{ - i, i_2 , i_3} ) ( {\cal T }_{l ; - j, i_2 , i_3} - \bar{{\cal T }}_{ - j, i_2 , i_3} )^\top (\hat{\btheta}_{j} - \btheta_{j})}_{\bf{I}_3} \label{eqn: ee}
\end{align}

Next we bound the last three terms of \eqref{eqn: ee}. For term \textbf{$\bf{I}_1$}, it can be re-formulated into,  $\forall 1 \le i \le j \le m_1$, 
\begin{align} \notag 
{\bf{I}}_1 &=  \tilde{\xi}_{l ; i, i_2 , i_3}  ({\cal T }_{l ; i, i_2 , i_3} - \bar{{\cal T }}_{ i, i_2 , i_3})(\hat{\theta}_{i ,j} - \theta_{i ,j})\ind \{  i\ne j \}  \\
& \quad \quad \quad \quad \quad \quad +   \sum \limits_{h \in \{1, \ldots , m_1 \} /\{ i,j\} } \tilde{\xi}_{l ; i, i_2 , i_3}  ({\cal T }_{l ; h, i_2 , i_3} - \bar{{\cal T }}_{ h, i_2 , i_3}) (\hat{\theta}_{h ,j} - \theta_{h ,j}) \label{eqn: ind}
\end{align}
where $\hat{\btheta}_{i} = (\hat{\theta}_{1 ,i} , \ldots , \hat{\theta}_{m_1 -1  ,i} )^\top$ and $\hat{\theta}_{m_1 ,i} =0$. 
The second term of \eqref{eqn: ind} can be bounded as follows,
\begin{align} \notag
& \max \limits_{ 1 \le i \le j \le m_1 } \bigg | \sum \limits_{h \in \{1, \ldots , m_1 \} /\{ i,j\}  }  \frac{1}{(n-1)m_2m_3}   \sum \limits_{l=1}^{n} \sum  \limits_{i_2=1}^{m_2}  \sum \limits_{i_3=1}^{m_3}   \tilde{\xi}_{l ; i, i_2 , i_3}  ({\cal T }_{l ; h, i_2 , i_3} - \bar{{\cal T }}_{ h, i_2 , i_3}) (\hat{\theta}_{h ,j} - \theta_{h ,j}) \bigg | \\  \label{eqn: I12}
 \le & \max \limits_{ 1 \le i \le j \le m_1 } \max \limits_{h \in \{1, \ldots , m_1 \} /\{ i,j\}  } \bigg |   \frac{1}{(n-1)m_2m_3}   \sum \limits_{l=1}^{n} \sum  \limits_{i_2=1}^{m_2}  \sum \limits_{i_3=1}^{m_3}  \tilde{\xi}_{l ; i, i_2 , i_3}  ({\cal T }_{l ; h, i_2 , i_3} - \bar{{\cal T }}_{ h, i_2 , i_3}) \bigg | \|\hat{\btheta}_{j} - \btheta_{j}\|_1 
\end{align}
Lemma \ref{lem: lem2} \eqref{eqn: lem2_1} implies 
$$\max_{1 \le i_1 \le m_1} \bigg | \frac{m_1}{nm}   \sum \limits_{l=1}^{n} \sum \limits_{i_2=1}^{m_2}  \cdots \sum \limits_{i_K=1}^{m_K}  \tilde{\xi}_{l ; i_1, \ldots , i_K} ({\cal T }_{l ; i_1, \ldots, i_K} - \bar{{\cal T }}_{ i_1, \ldots, i_K}) \bigg |= O_p \big ( \sqrt{\frac{\log  m_1 }{nm/m_1}} \big )$$
Together with \eqref{eqn: xi_approx}, we have
$${\bf{I}}_1 = \tilde{\xi}_{l ; i, i_2 , i_3}  ({\cal T }_{l ; i, i_2 , i_3} - \bar{{\cal T }}_{ i, i_2 , i_3})(\hat{\theta}_{i ,j} - \theta_{i ,j})\ind \{  i\ne j \} +  O_P (a_{n1} \sqrt{\frac{\log m_1}{nm_2m_3}}  ). $$
Convergency result of $\bf{I}_2$ has the similar arguments of $\bf{I}_1$. Then we turn to term ${\bf{I}}_3$. Let 
$$ \Xi_1 = \frac{1}{(n-1) m_2 m_3} \sum \limits_{l=1}^{n}  \sum \limits_{i_2=1}^{m_2} \sum \limits_{i_3=1}^{m_3} ({\cal T}_{l; : , i_2,  i_3} - \bar{{\cal T}}_{: , i_2,i_3}) ({\cal T}_{l; : , i_2, i_3} - \bar{{\cal T}}_{ : , i_2, i_3})^\top.$$ 
By Cauchy-Schwarz inequality, we get
\begin{align} \notag
& \big | \frac{1}{(n-1)m_2m_3}   \sum \limits_{l=1}^{n} \sum  \limits_{i_2=1}^{m_2}  \sum \limits_{i_3=1}^{m_3}  (\hat{\btheta}_{i} - \btheta_{i})^\top ( {\cal T }_{l ; - i, i_2 ,i_3} -  \bar{{\cal T }}_{ - i, i_2 , i_3} ) ( {\cal T }_{l ; - j, i_2 , i_3}  - \bar{{\cal T }}_{ - j, i_2 , i_3} )^\top (\hat{\btheta}_{j} - \btheta_{j}) \big |  \\ 
\label{eqn: I31}
\le &  \max \limits_{i \in \{1, \ldots , m_1\}} \big |  (\hat{\btheta}_{i} - \btheta_{i})^\top [\Xi_{1}]_{-i , -i } (\hat{\btheta}_{i} - \btheta_{i}) \big | , 
\end{align}
Denote $\vartheta = \tr (\bSigma_2) \tr (\bSigma_3) /(m_2 m_3) $. Condition \ref{con:eigenvalue_tensor} implies that $C_1 < \vartheta < C_1^{-1}$. Then we use triangle inequality into \eqref{eqn: I31}
\begin{align} \notag 
\big |  (\hat{\btheta}_{i} - \btheta_{i})^\top [\Xi_{1}]_{-i , -i } (\hat{\btheta}_{i} - \btheta_{i}) \big | \le & \big |  (\hat{\btheta}_{i} - \btheta_{i})^\top \big( [\Xi_{1}]_{-i , -i } - \vartheta [\bSigma_1^*]_{-i,-i} \big) (\hat{\btheta}_{i} - \btheta_{i}) \big | \\ \label{eqn: I32}
&\quad\quad\quad +  \vartheta  \big |  (\hat{\btheta}_{i} - \btheta_{i})^\top  [\bSigma_1^*]_{-i,-i} (\hat{\btheta}_{i} - \btheta_{i}) \big | 
\end{align}
Lemma \ref{lem: lem1} implies 
\begin{equation} \label{eqn: I36}
[\Xi_{1}]_{-i , -i } - \vartheta [\bSigma_1^*]_{-i,-i} = O_p (\sqrt{\frac{\log m_1}{n m_2 m_3}}).
\end{equation}
Together with 
$$\max \limits_{i \in \{1 ,\ldots , m_1\}}\|\hat{\btheta}_{i} - \btheta_{i} \|_1 =O(a_{n1}),$$
the first term of \eqref{eqn: I32} is bounded as 
\begin{equation} \label{eqn: I34}
\max \limits_{i \in \{ 1, \ldots, m_1\}}\big |  (\hat{\btheta}_{i} - \btheta_{i})^\top \big( [\Xi_{1}]_{-i , -i } - \vartheta [\bSigma_1^*]_{-i,-i} \big) (\hat{\btheta}_{i} - \btheta_{i}) \big |  = O_p \bigg (  a_{n1}^2 \sqrt{\frac{\log m_1}{nm_2m_3}} \bigg )
\end{equation}
and the second term is 
\begin{equation} \label{eqn: I33}
\big |  (\hat{\btheta}_{i} - \btheta_{i})^\top  [\bSigma_1^*]_{-i,-i} (\hat{\btheta}_{i} - \btheta_{i}) \big |  = O_P(a_{n2}^2 \big ).
\end{equation}
Combining \eqref{eqn: I31}, \eqref{eqn: I32}, \eqref{eqn: I34} and \eqref{eqn: I33}, $\bf{I}_3$ is upper bounded by
\begin{align} \notag
& \big | \frac{1}{(n-1)m_2m_3}   \sum \limits_{l=1}^{n} \sum  \limits_{i_2=1}^{m_2}  \sum \limits_{i_3=1}^{m_3}  (\hat{\btheta}_{i} - \btheta_{i})^\top  ( {\cal T }_{l ; - i, i_2 ,i_3}  -  \bar{{\cal T }}_{ - i, i_2 , i_3} ) ( {\cal T }_{l ; - j, i_2 , i_3}  - \bar{{\cal T }}_{ - j, i_2 , i_3} )^\top (\hat{\btheta}_{j} - \btheta_{j}) \big |  \\
 = & O_P \big ( a_{n_2}^2+ a_{n1}^2 \sqrt{\frac{\log m_1}{nm_2m_3}}  \big ).
\end{align}
 So far, $\hat{\varrho}_{i,j}$ becomes
\begin{align} \notag
\hat{\varrho}_{i,j} =& \frac{1}{(n-1)m_2m_3}   \sum \limits_{l=1}^{n} \sum  \limits_{i_2=1}^{m_2}  \sum \limits_{i_3=1}^{m_3} \hat{\xi}_{l ; i, i_2 , i_3} \hat{\xi}_{l ; j, i_2 , i_3}\\ \notag
 =  &  \frac{1}{(n-1)m_2m_3}   \sum \limits_{l=1}^{n} \sum  \limits_{i_2=1}^{m_2}  \sum \limits_{i_3=1}^{m_3} \tilde{\xi}_{l ; i, i_2 , i_3} \tilde{\xi}_{l ; j, i_2 , i_3}  \\ \notag
& -  \frac{1}{(n-1)m_2m_3}   \sum \limits_{l=1}^{n} \sum  \limits_{i_2=1}^{m_2}  \sum \limits_{i_3=1}^{m_3}\tilde{\xi}_{l ; i, i_2 , i_3}  ({\cal T }_{l ; i, i_2 , i_3} - \bar{{\cal T }}_{ i, i_2 , i_3})(\hat{\theta}_{i ,j} - \theta_{i ,j})\ind \{  i\ne j \} \\ \notag
& -  \frac{1}{(n-1)m_2m_3}   \sum \limits_{l=1}^{n} \sum  \limits_{i_2=1}^{m_2}  \sum \limits_{i_3=1}^{m_3}\tilde{\xi}_{l ; j, i_2 , i_3}  ({\cal T }_{l ; j, i_2 , i_3} - \bar{{\cal T }}_{ j, i_2 , i_3})(\hat{\theta}_{j-1 ,i} - \theta_{j-1 ,i})\ind \{  i\ne j \} \\
& + O_P \big(  (a_{n1}^2 + a_{n1}) \sqrt{\frac{\log m_1}{nm_2m_3}}  +a_{n2}^2 \big) \label{eqn: ind1}
\end{align}
for $1 \le i \le j \le m_1 $. 

We then bound the second and the third term in \eqref{eqn: ind1}. 
Recall the definition of $\tilde{\xi}_{l; i_1, i_2 , i_3}$ in \eqref{eqn: xi_tilde}, we have
\begin{align} \notag
& \frac{1}{(n-1)m_2m_3}   \sum \limits_{l=1}^{n} \sum  \limits_{i_2=1}^{m_2}  \sum \limits_{i_3=1}^{m_3}\tilde{\xi}_{l ; i_1, i_2 , i_3}  ({\cal T }_{l ; i_1, i_2 , i_3} - \bar{{\cal T }}_{ i_1, i_2 , i_3}) \\  \notag
 = &  \frac{1}{(n-1)m_2m_3}   \sum \limits_{l=1}^{n} \sum  \limits_{i_2=1}^{m_2}  \sum \limits_{i_3=1}^{m_3}   (\tilde{\xi}_{l ; i_1, i_2 , i_3}  )^2 \\ \label{eqn: I41}
 & \quad + \frac{1}{(n-1)m_2m_3}   \sum \limits_{l=1}^{n} \sum  \limits_{i_2=1}^{m_2}  \sum \limits_{i_3=1}^{m_3} \tilde{\xi}_{l ; i_1, i_2 , i_3}  ({\cal T }_{l ; - i_1, i_2 , i_3} - \bar{{\cal T }}_{- i_1, i_2 , i_3})^\top \btheta_{i_1 } 
 \end{align}
Lemma \ref{lem: lem2} \eqref{eqn: lem2_2} implies 
\begin{equation} \label{eqn: I42}
\max_{1 \le i_1 \le m_1} \bigg | \frac{m_1}{nm}   \sum \limits_{l=1}^{n} \sum \limits_{i_2=1}^{m_2}  \cdots \sum \limits_{i_K=1}^{m_K}  \tilde{\xi}_{l ; i_1, \ldots , i_K} ({\cal T }_{l ; -i_1, \ldots, i_K} - \bar{{\cal T }}_{ -i_1, \ldots, i_K})^\top \btheta_{i_1} \bigg |= O_p \big ( \sqrt{\frac{\log  m_1 }{nm/m_1}} \big ).
\end{equation}
When $ 1 \le i = j  \le m_1$, \eqref{eqn: ind1} becomes 
\begin{align} \notag
\frac{1}{(n-1)m_2m_3}   \sum \limits_{l=1}^{n} \sum  \limits_{i_2=1}^{m_2}  \sum \limits_{i_3=1}^{m_3} (\hat{\xi}_{l ; i, i_2 , i_3} )^2 = &    \frac{1}{(n-1)m_2m_3}   \sum \limits_{l=1}^{n} \sum  \limits_{i_2=1}^{m_2}  \sum \limits_{i_3=1}^{m_3}(\tilde{\xi}_{l ; j, i_2 , i_3} )^2  \\
& \quad  + O_P \big(  (a_{n1}^2 + a_{n1}) \sqrt{\frac{\log m_1}{nm_2m_3}} +a_{n2}^2 \big) \label{eqn: ind2}
\end{align}
Applying \eqref{eqn: ind2} and \eqref{eqn: I42}  into \eqref{eqn: I41}, we have
\begin{align} \notag
\frac{1}{(n-1)m_2m_3}   \sum \limits_{l=1}^{n} \sum  \limits_{i_2=1}^{m_2}  \sum \limits_{i_3=1}^{m_3}\tilde{\xi}_{l ; i_1, i_2 , i_3}  ({\cal T }_{l ; i_1, i_2 , i_3} - \bar{{\cal T }}_{ i_1, i_2 , i_3})  =& \frac{1}{(n-1)m_2m_3}   \sum \limits_{l=1}^{n} \sum  \limits_{i_2=1}^{m_2}  \sum \limits_{i_3=1}^{m_3}   (\hat{\xi}_{l ; i_1, i_2 , i_3}  )^2   \\ 
& + O_P \big(  (a_{n1}^2 + a_{n1} +1 ) \sqrt{\frac{\log m_1}{nm_2m_3}}  +a_{n2}^2 \big)  \label{eqn: et}
\end{align}
Note that 
\begin{equation} \label{eqn: I44}
\max \limits_{i \in \{1 \ldots , m_1\}} \| \hat{\btheta}_{i } - \btheta_{i } \|_{\infty} \le \max \limits_{i \in \{1 \ldots , m_1\}} \| \hat{\btheta}_{i } - \btheta_{i } \|_{1} = O_P (a_{n1}) = O_P(1) .
\end{equation}
Following \eqref{eqn: et} and \eqref{eqn: I44}, simple algebra implies that \eqref{eqn: ind2} can be rewritten as, $\forall 1 \le i < j \le m_1$,
\begin{align} \notag
\hat{\varrho}_{i,j}=  &  \frac{1}{(n-1)m_2m_3}   \sum \limits_{l=1}^{n} \sum  \limits_{i_2=1}^{m_2}  \sum \limits_{i_3=1}^{m_3} \tilde{\xi}_{l ; i, i_2 , i_3} \tilde{\xi}_{l ; j, i_2 , i_3}  -  \frac{1}{(n-1)m_2m_3}   \sum \limits_{l=1}^{n} \sum  \limits_{i_2=1}^{m_2}  \sum \limits_{i_3=1}^{m_3}(\hat{\xi}_{l ; i, i_2 , i_3})^2(\hat{\theta}_{i ,j} - \theta_{i ,j}) \\\label{eqn: e_hate_hat}
& -  \frac{1}{(n-1)m_2m_3}   \sum \limits_{l=1}^{n} \sum  \limits_{i_2=1}^{m_2}  \sum \limits_{i_3=1}^{m_3} (\hat{\xi}_{l ; j, i_2 , i_3})^2(\hat{\theta}_{j-1 ,i} - \theta_{j-1 ,i})   + O_P \big(  (a_{n1}^2 + a_{n1}+1) \sqrt{\frac{\log m_1}{nm_2m_3}}  +a_{n2}^2 \big)   
\end{align}
Given the convergency result in \eqref{eqn: e_hate_hat}, the bias-corrected sample covariance is 
\begin{align} \notag
\hat{\varrho}_{i,j} + \mu_{i,j} 
	& =  \frac{1}{(n-1)m_2m_3}   \sum \limits_{l=1}^{n} \sum  \limits_{i_2=1}^{m_2}  \sum \limits_{i_3=1}^{m_3} \tilde{\xi}_{l ; i, i_2 , i_3} \tilde{\xi}_{l ; j, i_2 , i_3}+ \frac{1}{(n-1)m_2m_3}   \sum \limits_{l=1}^{n} \sum  \limits_{i_2=1}^{m_2}  \sum \limits_{i_3=1}^{m_3}(\hat{\xi}_{l ; i, i_2 , i_3})^2 \theta_{i ,j} \\\label{eqn: trans1}
	& +  \frac{1}{(n-1)m_2m_3}   \sum \limits_{l=1}^{n} \sum  \limits_{i_2=1}^{m_2}  \sum \limits_{i_3=1}^{m_3} (\hat{\xi}_{l ; j, i_2 , i_3})^2 \theta_{j-1 ,i}  + O_P \big(  (a_{n1}^2 + a_{n1}+1) \sqrt{\frac{\log m_1}{nm_2m_3}}  +a_{n2}^2 \big). 
\end{align}
Let 
$$ [\Xi_1]_{ii;\xi} = \frac{1}{(n-1)m_2 m_3 }  \sum \limits_{l=1}^{n} \sum  \limits_{i_2=1}^{m_2}  \sum \limits_{i_3=1}^{m_3}(\hat{\xi}_{l ; i, i_2 , i_3})^2  .$$ 
Recall that $\theta_{i,j} =  - [ \bOmega_1^*]_{j,j}^{-1}   [ \bOmega_1^*]_{i,j} $ and $\theta_{j-1 , i} =  - [ \bOmega_1^*]_{ i , i}^{-1}  [ \bOmega_1^*]_{j , i }$. \eqref{eqn: trans1} can be rewritten into
\begin{align} \notag
\hat{\varrho}_{i,j} + \mu_{i,j}	 & =  \frac{1}{(n-1)m_2m_3}   \sum \limits_{l=1}^{n} \sum  \limits_{i_2=1}^{m_2}  \sum \limits_{i_3=1}^{m_3} \big ( \tilde{\xi}_{l ; i, i_2 , i_3} \tilde{\xi}_{l ; j, i_2 , i_3}   - \EE \tilde{\xi}_{l ; i, i_2 , i_3} \tilde{\xi}_{l ; j, i_2 , i_3} \big ) \\ \notag
	 &\quad \quad \quad \quad \quad + [\Xi_1]_{ii;\xi} ( - [ \bOmega_1^*]_{j,j}^{-1} )  [ \bOmega_1^*]_{i,j} + [\Xi_1]_{jj;\xi} ( - [ \bOmega_1^*]_{i,i}^{-1} )  [ \bOmega_1^*]_{j,i} \\ \notag
	 &\quad \quad \quad \quad \quad + \frac{1}{(n-1)m_2m_3}   \sum \limits_{l=1}^{n} \sum  \limits_{i_2=1}^{m_2}  \sum \limits_{i_3=1}^{m_3} \EE \tilde{\xi}_{l ; i, i_2 , i_3} \tilde{\xi}_{l ; j, i_2 , i_3}  \\ \notag
 	 &\quad \quad \quad \quad \quad + O_P \big(  (a_{n1}^2 + a_{n1}+1) \sqrt{\frac{\log m_1}{nm_2m_3}}  +a_{n2}^2 \big).
\end{align}	 
Define 
$$b_{ij} = [ \bOmega_1^*]_{i,i} [\Xi_1]_{ii;\xi}+ [ \bOmega_1^*]_{j,j} [\Xi_1]_{jj;\xi} - m_1 \tr (\bSigma_2) \tr (\bSigma_3) / m.$$
Following 
$$ \EE \tilde{\xi}_{l ; i, i_2 , i_3} \tilde{\xi}_{l ; j, i_2 , i_3} = (n-1)[ \bSigma_2^*]_{i_2,i_2} [ \bSigma_3^*]_{i_3,i_3}  [ \bOmega_1^*]_{i,j}/ (  n [ \bOmega_1^*]_{i,i}[ \bOmega_1^*]_{j,j} ) $$
 \cite{peng2012}, simple algebra implies
\begin{align}	   \notag
\hat{\varrho}_{i,j} + \mu_{i,j}  & = -b_{ij} \frac{ [ \bOmega_1^*]_{i,j}}{ [ \bOmega_1^*]_{i,i} [ \bOmega_1^*]_{j,j}}  +\frac{1}{(n-1)m_2m_3}   \sum \limits_{l=1}^{n} \sum  \limits_{i_2=1}^{m_2}  \sum \limits_{i_3=1}^{m_3} \big ( \tilde{\xi}_{l ; i, i_2 , i_3} \tilde{\xi}_{l ; j, i_2 , i_3}   - \EE \tilde{\xi}_{l ; i, i_2 , i_3} \tilde{\xi}_{l ; j, i_2 , i_3} \big ) \\
  	 &\quad \quad \quad \quad \quad + O_P \big(  (a_{n1}^2 + a_{n1}+1) \sqrt{\frac{\log m_1}{nm_2m_3}}  +a_{n2}^2 \big) .
\end{align} 
The proof is complete.   \hfill $\blacksquare$\\

\begin{lemma} \label{lem: lem1}
Assume i.i.d. tensor data ${\cal T}, {\cal T}_1,\ldots, {\cal T}_n \in \mathbb R^{m_1 \times m_2 \times \cdots \times m_K}$ follows the tensor normal distribution $\textrm{TN}({\bf0}; \bSigma_1^*, \ldots, \bSigma_K^*)$. Let 
$$ \Xi_1 = \frac{m_1}{(n-1) m} \sum \limits_{l=1}^{n}  \sum \limits_{i_2=1}^{m_2} \cdots \sum \limits_{i_K=1}^{m_K} ({\cal T}_{l; : , i_2, \ldots ,i_K} - \bar{{\cal T}}_{: , i_2, \ldots ,i_K}) ({\cal T}_{l; : , i_2, \ldots ,i_K} - \bar{{\cal T}}_{ : , i_2, \ldots ,i_K})^\top.$$
Assume Condition \ref{con:eigenvalue_tensor} and $\log  m_1 = o(nm/m_1)$ hold, we have
$$\EE ( \Xi_1 ) = \frac{m_1 \prod_{k \in \{2, \ldots , K\} }\tr(\bSigma_k^*)}{m} \bSigma_1^* $$
and 
$$\big \| \Xi_1 - \frac{m_1 \prod_{k \in \{2, \ldots , K\} }\tr(\bSigma_k^*)}{m} \bSigma_1^* \big \|_{\infty} = O_p \big ( \sqrt{\frac{m_1 \log m_1 }{ n m}}\big ).$$
\end{lemma}

\noindent{{\bf Proof of Lemma \ref{lem: lem1}}:}
To ease the presentation, we show that Lemma \ref{lem: lem1} holds when $K=3$. The proof can easily be generalized to the case when $K> 3$. 

We prove it by transforming $[\Xi_1]_{i,j}$ into summation of product of two random normal variables, then computing its expectation and applying Lemma 1 in \cite{cai2011adaptive} to get convergency result. 

We first transform $[\Xi_1]_{i,j}$. Let $\bO \in \RR^{n \times n}$ be an orthogonal matrix, the last row is $(1/\sqrt{n}, \ldots , 1/\sqrt{n})^T$. Define $\Yb_{i_1,i_2 , i_3} = (Y_{1 ; i_1,i_2 , i_3} , \ldots , Y_{n ; i_1,i_2 , i_3})^\top = \bO ({\cal T}_{1; i_1 , i_2 ,i_3} , \ldots, {\cal T}_{n; i_1 , i_2 ,i_3} )^\top$. Then, we rewrite $\Xi_1$ by
\begin{align} \notag
 \sum \limits_{l=1}^{n} ({\cal T}_{l; i , i_2 ,i_3} - \bar{{\cal T}}_{i, i_2 ,i_3}) ({\cal T}_{l; j , i_2 , i_3} - \bar{{\cal T}}_{ j, i_2 , i_3}) & = \Yb_{i,i_2 , i_3}^\top \Yb_{j,i_2 , i_3} - Y_{n ; i ,i_2 , i_3}Y_{n ; j ,i_2 , i_3}\\
& = \sum \limits_{l=1}^{n-1}Y_{l ; i ,i_2 , i_3}Y_{l ; j ,i_2 , i_3}
\end{align}
Let $\Yb_{l} = (Y_{l; i_1, i_2 ,i_3})_{i_k \in \{1 , \ldots , m_k  \} , k \in \{1,2,3\}}$ for every $l \in \{1 , \ldots, n-1 \}$. 
It is obvious to see that  $({\cal T}_{1; i_1 , i_2 ,i_3} , \ldots, {\cal T}_{n; i_1 , i_2 ,i_3} )^\top \sim \textrm{N} ({\bf{0}} , [\bSigma_1^*]_{i_1,i_1} [\bSigma_2^*]_{i_2,i_2}[\bSigma_3^*]_{i_3,i_3} \Ib_{n\times n})$ and 
$ (Y_{1 ; i_1,i_2 , i_3} , \ldots , Y_{n -1; i_1,i_2 , i_3})^\top \sim \textrm{N}( \bm{0}, [\bSigma_1^*]_{i_1,i_1} [\bSigma_2^*]_{i_2,i_2}[\bSigma_3^*]_{i_3,i_3} \Ib_{n-1\times n-1})$. Therefore $\Yb_l \sim \textrm{TN} ({\bm{0}} ; \bSigma_1^*, \bSigma_2^*, \bSigma_3^*)$, i.i.d. for $1 \le l \le n-1$. Let $\Zb_l = \Yb_l \times_3 {\bSigma_3^*}^{-1/2} \sim \textrm{TN} (\bm{0} ; \bSigma_1^*, \bSigma_2^*, \Ib_{m_3 \times m_3})$. Define $\Ub_3^{\top} \Db_3 \Ub_3$ as the eigenvalue decomposition of $\bSigma_3^*$, where $\Ub_3$ is an orthogonal matrix and $\Db_3=\diag(\lambda_1^{(3)} , \ldots , \lambda_{m_3}^{(3)})$ with $\bSigma_3^*$'s eigenvalue $\lambda_1^{(3)} \le  \ldots \le \lambda_{m_3}^{(3)}$. Let $\Wb_l = \Zb_l \times_3 \Ub_3 \sim \textrm{N}(\bm{0}; \bSigma_1^*, \bSigma_2^*, \Ib_{m_3 \times m_3})$. Similarly, we set $\Vb_l = \Wb_l \times_2 {\bSigma_2^*}^{-1/2} \sim \textrm{N}(\bm{0} ; \bSigma_1^*, \Ib_{m_2 \times m_2}, \Ib_{m_3 \times m_3})$ and define $\Ub_2^{\top} \Db_2 \Ub_2$ as the eigenvalue decomposition of $\bSigma_2^*$, where $\Ub_2$ is an orthogonal matrix and $\Db_2=\diag(\lambda_1^{(2)} , \ldots , \lambda_{m_2}^{(2)})$ with $\bSigma_2^*$'s eigenvalue $\lambda_1^{(2)} \le  \ldots \le \lambda_{m_2}^{(2)}$. Let $\Gb_l = \Vb_l \times_2 \Ub_2 \sim \textrm{N}( \bm{0};\bSigma_1^*, \Ib_{m_2 \times m_2}, \Ib_{m_3 \times m_3} ).$ $[\Xi_1]_{i,j}$ is transformed as follows,
\begin{align} \notag
[\Xi_1]_{i,j} & =\frac{1}{(n-1)m_2m_3} \sum \limits_{l=1}^{n-1} \sum \limits_{i_2=1}^{m_2}  \Yb_{l; i, i_2 , :}^\top \Yb_{l; j, i_2 , :} \\\notag
& =\frac{1}{(n-1)m_2m_3} \sum \limits_{l=1}^{n-1}  \sum \limits_{i_2=1}^{m_2}  (\Ub_3 \Zb_{l; i, i_2 , :})^\top \Db_3 (\Ub_3 \Zb_{l; j, i_2 , :}) \\\notag
& =\frac{1}{(n-1)m_2m_3} \sum \limits_{l=1}^{n-1}  \sum \limits_{i_2=1}^{m_2} \sum \limits_{i_3 =1}^{m_3} \lambda_{i_3}^{(3)} W_{l ; i, i_2, i_3} W_{l ; j, i_2, i_3}  \\\notag
& =\frac{1}{(n-1)m_2m_3} \sum \limits_{l=1}^{n-1}   \sum \limits_{i_3 =1}^{m_3} \lambda_{i_3}^{(3)} \sum \limits_{i_2=1}^{m_2} W_{l ; i, i_2, i_3} W_{l ; j, i_2, i_3}  \\\notag
& =\frac{1}{(n-1)m_2m_3} \sum \limits_{l=1}^{n-1}   \sum \limits_{i_3 =1}^{m_3} \lambda_{i_3}^{(3)}  \Wb_{l ; i, : , i_3}^\top \Wb_{l ; j, :, i_3}  \\\notag
& =\frac{1}{(n-1)m_2m_3} \sum \limits_{l=1}^{n-1}   \sum \limits_{i_3 =1}^{m_3} \lambda_{i_3}^{(3)}   (\Ub_2 \Vb_{l; i, : , i_3})^\top \Db_2 (\Ub_2 \Vb_{l; j, : , i_3}) \\ \label{eqn: xi_trans}
& =\frac{1}{(n-1)m_2m_3} \sum \limits_{l=1}^{n-1}  \sum \limits_{i_2=1}^{m_2} \sum \limits_{i_3 =1}^{m_3}    \lambda_{i_2}^{(2)} \lambda_{i_3}^{(3)} G_{l ; i, i_2, i_3} G_{l ; j, i_2, i_3}
\end{align}

Next we compute the expectation of $[\Xi_1]_{i,j}$ based on \eqref{eqn: xi_trans}.
Since
\[
(G_{l ; i, i_2, i_3} G_{l ; j, i_2, i_3})  \sim  \textrm{N} \bigg{\{} \bm{0} ; 
  \begin{pmatrix}
    [\bSigma_1^*]_{i,i} &     [\bSigma_1^*]_{i,j}\\
    [\bSigma_1^*]_{j,i} &     [\bSigma_1^*]_{j,j}
  \end{pmatrix} \bigg{\}},
\] 
we have $\EE G_{l ; i, i_2, i_3} G_{l ; j, i_2, i_3} =   [\bSigma_1^*]_{i,j}.$ Therefore, the expectation of $[\Xi_1]_{i,j}$ is 
$$\EE [\Xi_1]_{i,j} =  \frac{1}{(n-1)m_2m_3} \sum \limits_{l=1}^{n-1}  \sum \limits_{i_2=1}^{m_2} \sum \limits_{i_3 =1}^{m_3}    \lambda_{i_2}^{(2)} \lambda_{i_3}^{(3)}   [\bSigma_1^*]_{i,j} =  \frac{\tr(\bSigma_2)\cdot \tr(\bSigma_3)}{m_2 m_3 } [\bSigma_1^*]_{i,j}.$$
Then we prove the conditions of Lemma 1 in \cite{cai2011adaptive}. Let 
$$G_{l ; i j, i_2, i_3}= G_{l ; i, i_2, i_3} G_{l ; j, i_2, i_3} - \EE G_{l ; i, i_2, i_3} G_{l ; j, i_2, i_3}.$$
There exists some sufficiently small constant $\eta > 0$ and large $C>0$, s.t. 
$$\EE \exp\{ 2\eta |\lambda_{i_2}^{(2)} \lambda_{i_3}^{(3)} G_{l ; i j, i_2, i_3} | \} \le C,$$
 uniformly in $i,j,i_2,i_3, l$. Hence, under Condition \ref{con:eigenvalue_tensor}, by Cauchy-Schwarz inequality, 
\begin{align} \notag
  & \sum \limits_{l=1}^{n-1}  \sum \limits_{i_2=1}^{m_2} \sum \limits_{i_3 =1}^{m_3}  \EE  (\lambda_{i_2}^{(2)} \lambda_{i_3}^{(3)} G_{l ; i j, i_2, i_3})^2  \exp\{ \eta |\lambda_{i_2}^{(2)} \lambda_{i_3}^{(3)} G_{l ; i j, i_2, i_3} | \}    \\ \notag
\le & C \sum \limits_{l=1}^{n-1}  \sum \limits_{i_2=1}^{m_2} \sum \limits_{i_3 =1}^{m_3}  (\EE  (\lambda_{i_2}^{(2)} \lambda_{i_3}^{(3)} G_{l ; i j, i_2, i_3})^4)^{1/2}   \\ \label{eqn: cond_cai}
= & C \sqrt{2} (n-1) \|\bSigma_2^*\|_F^2 \|\bSigma_3^*\|_F^2 \big  ( [\bSigma_1^*]_{i,i} [\bSigma_1^*]_{j,j}  + ([\bSigma_1^*]_{i,j} )^2 \big )
\end{align}
Applying the inequality $\|\bSigma_k^*\|_F^2 \le \lambda_{\max}^2 (\bSigma_k^*) m_k$, \eqref{eqn: cond_cai} is bounded by $O( nm_2m_3)$. So far, the conditions of Lemma 1 in \cite{cai2011adaptive} are proven. This lemma implies
$$\big \| \Xi_1  - \EE ( \Xi_1)  \big \|_{\infty} = O_p \bigg (\sqrt{\frac{\log m_1 }{n m_2 m_3 }} \bigg ).$$
The proof of Lemma \ref{lem: lem1} is complete.    \hfill $\blacksquare$\\

\begin{lemma} \label{lem: lem2}
Assume i.i.d. tensor data ${\cal T}, {\cal T}_1,\ldots, {\cal T}_n \in \mathbb R^{m_1 \times m_2 \times \cdots \times m_K}$ follows the tensor normal distribution $\textrm{TN}({\bf0}; \bSigma_1^*, \ldots, \bSigma_K^*)$.  Let 
$$\tilde{\xi}_{l ; i_1, \ldots , i_K} = {\cal T }_{l ; i_1, \ldots , i_K} - \bar{{\cal T }}_{ i_1, \ldots  , i_K} - ( {\cal T }_{l ; - i_1, \ldots  , i_K} - \bar{{\cal T }}_{ - i_1, \ldots , i_K} )^\top \btheta_{i_1},$$
 where $\bm{\theta}_{i_1}$ is defined in \S\ref{sec: construct}. Assume Condition \ref{con:eigenvalue_tensor} and $\log m_1 = o(nm/m_1)$ hold, we have
\begin{equation} \label{eqn: lem2_1}
\max_{1 \le i_1 \le m_1} \bigg | \frac{m_1}{nm}   \sum \limits_{l=1}^{n} \sum \limits_{i_2=1}^{m_2}  \cdots \sum \limits_{i_K=1}^{m_K}  \tilde{\xi}_{l ; i_1, \ldots , i_K} ({\cal T }_{l ; i_1, \ldots, i_K} - \bar{{\cal T }}_{ i_1, \ldots, i_K}) \bigg |= O_p \big ( \sqrt{\frac{\log  m_1 }{nm/m_1}} \big )
\end{equation}
and 
\begin{equation} \label{eqn: lem2_2}
\max_{1 \le i_1 \le m_1} \bigg | \frac{m_1}{nm}   \sum \limits_{l=1}^{n} \sum \limits_{i_2=1}^{m_2}  \cdots \sum \limits_{i_K=1}^{m_K}  \tilde{\xi}_{l ; i_1, \ldots , i_K} ({\cal T }_{l ; -i_1, \ldots, i_K} - \bar{{\cal T }}_{ -i_1, \ldots, i_K})^\top \btheta_{i_1} \bigg |= O_p \big ( \sqrt{\frac{\log  m_1 }{nm/m_1}} \big )
\end{equation}
\end{lemma} 
\noindent{{\bf Proof of Lemma \ref{lem: lem2}}:}
To ease the presentation, we show that Lemma \ref{lem: lem2} holds when $K=3$. The proof can easily be generalized to the case when $K> 3$. 

We prove it by first constructing tensor normal variables that combine $\xi_{l ; i_1, i_2 ,i _3 }$ with ${\cal T }_{l ; i_1, i_2 , i_3}$ and ${\cal T }_{l ; - i_1, i_2 , i_3}^\top\bm{\theta}_{i_1}$, then adopting the proof strategy of Lemma \ref{lem: lem1} to get similar convergency results.

Recall that 
$$\xi_{l ; i_1, i_2 , i_3} =  {\cal T }_{l ; i_1, i_2 ,  i_3} - {\cal T }_{l ; - i_1, i_2 , i_3}^\top \btheta_{i_1}.$$ 
Let 
$$
\bxi_{l ; i_1} = (\xi_{l ; i_1, i_2 , i_3}  )_{  i_2 \in \{1 , \ldots , m_2\} ,  i_3 \in \{1 , \ldots , m_3\} } \in  \RR^{m_2 \times m_3} ,
$$
and
$$
\Yb_{l ; i_1} = ({\cal T }_{l ; - i_1, i_2 , i_3}^\top \btheta_{i_1})_{ i_2 \in \{1 , \ldots , m_2\} , i_3 \in \{1 , \ldots , m_3\} } \in  \RR^{m_2 \times m_3}.
$$
Applying Lemma 1 in \cite{peng2012}, we have
$$\bxi_{l ; i_1}\sim \textrm{TN}(\bm{0};[\bOmega^*_1]_{i_1 , i_1 }^{-1}, \bSigma_2^* ,\bSigma_3^* ), \; \Yb_{l ; i_1} \sim \textrm{TN}(\bm{0}; \frac{[\bSigma_1^*]_{i_1 , i_1 }[\bOmega_1^*]_{i_1 , i_1 } -1 }{ [\bOmega_1^*]_{i_1,i_1}}, \bSigma_2^* ,\bSigma_3^*), $$
and $\bxi_{l ; i_1}$ is independent of $\Yb_{l ; i_1}$ for $1 \le l \le n$, $1 \le i_1 \le m_1$.

Next we first construct tensor normal variables by combining $\xi_{l ; i_1, i_2 ,i _3 }$ with ${\cal T }_{l ; i_1, i_2 , i_3}$. Let  ${\cal T }_{l ; h} = ({\cal T }_{l ; h, i_2 ,  i_3})_{ i_2 \in \{1 , \ldots , m_2\} , i_3 \in \{1 , \ldots , m_3\} } \in  \RR^{m_2 \times m_3}.$ Obviously ${\cal T }_{l ; h} \sim \textrm{TN} (\bm{0}; [\bSigma_1^*]_{hh} , \bSigma_2^* , \bSigma_3^*)$, and ${\cal T }_{l ; h}$ is independent of $\bxi_{l ; i_1}$, for $ h,i_1 \in \{1 , \ldots , m_1 \} , h\ne i_1 , l \in \{1 ,\ldots , n\}$. Let $\Zb_{l;[i_1,h]} \in \RR^{2\times m_2 \times m_3}$ with $[\Zb_{l;[i_1,h]}]_{1,: , : }= \bxi_{l ; i_1}$  and $[\Zb_{l;[i_1,h]}]_{2,: , : }= {\cal T }_{l ; h}.$ It is clear that 
 \[
\Zb_{l;[i_1,h]}  \sim  \textrm{TN} \bigg{\{} \bm{0} ; 
  \begin{pmatrix}
    [\bOmega^*_1]_{i_1 , i_1 }^{-1} &     0\\
    0 &     [\bSigma_1^*]_{h,h}
  \end{pmatrix} , \bSigma_2^*, \bSigma_3^* \bigg{\}}.
\]
We then combine $\xi_{l ; i_1, i_2 ,i _3 }$ with  ${\cal T }_{l ; - i_1, i_2 , i_3}^\top\bm{\theta}_{i_1}$. Define $\Ub_{l;[i,j]} \in \RR^{2\times m_2 \times m_3}$ with $[\Ub_{l;[i,j]}]_{1,: , : }= \bxi_{l ; i}$  and $[\Ub_{l;[i,j]}]_{2,: , : }= \Yb_{l ; j}.$ Hence, 
\[
\Ub_{l;[i_1,i_1]}  \sim  \textrm{TN} \bigg{\{} \bm{0} ; 
  \begin{pmatrix}
    [\bOmega^*_1]_{i_1 , i_1 }^{-1} &     0\\
    0 &     \frac{[\bSigma_1^*]_{i_1 , i_1 }[\bOmega_1^*]_{i_1 , i_1 } -1 }{ [\bOmega_1^*]_{i_1,i_1}}
  \end{pmatrix} , \bSigma_2^*, \bSigma_3^* \bigg{\}}.
\]
Finallly, the rest of the proof exactly follows Lemma \ref{lem: lem1}. Details are eliminated. The proof is complete.    \hfill $\blacksquare$\\

\begin{lemma} \label{lem: lem3}
Assume i.i.d. tensor data ${\cal T}, {\cal T}_1,\ldots, {\cal T}_n \in \mathbb R^{m_1 \times m_2 \times \cdots \times m_K}$ follows the tensor normal distribution $\textrm{TN}({\bf0}; \bSigma_1^*, \ldots, \bSigma_K^*)$. $\bOmega_k^*$ is the inverse of $\bSigma_k^*$. Let 
$$\tilde{\xi}_{l ; i_1, \ldots , i_K} = {\cal T }_{l ; i_1, \ldots , i_K} - \bar{{\cal T }}_{ i_1, \ldots  , i_K} - ( {\cal T }_{l ; - i_1, \ldots  , i_K} - \bar{{\cal T }}_{ - i_1, \ldots , i_K} )^\top \btheta_{i_1}$$
 and 
$$[\Xi_1]_{ij ; \tilde{\xi}} = \frac{m_1}{(n-1) m} \sum \limits_{l=1}^{n} \sum \limits_{i_2=1}^{m_2}  \cdots \sum \limits_{i_K=1}^{m_K}  \tilde{\xi}_{l;i, \ldots, i_K} \tilde{\xi}_{l; j, \ldots, i_K},$$
 where $\bm{\theta}_{i_1}$ is defined in \S\ref{sec: construct}.

(i). As $nm/m_1 \rightarrow \infty$, 
$$\frac{\sum \limits_{l=1}^{n} \sum \limits_{i_2=1}^{m_2}  \cdots \sum \limits_{i_K=1}^{m_K} \big (\tilde{\xi}_{l;i, \ldots, i_K} \tilde{\xi}_{l; j, \ldots, i_K} - \EE \tilde{\xi}_{l;i, \ldots, i_K} \tilde{\xi}_{l; j, \ldots, i_K} \big ) }{\sqrt{(n-1) \prod_{k=2}^K \| \bSigma_k^*\|_F^2 }} \rightarrow \textrm{N} \bigg  ( 0 ; \frac{1}{[\bOmega_1^*]_{i,i} [\bOmega_1^*]_{j,j}} + \frac{([\bOmega_1^*]_{i,j})^2}{([\bOmega_1^*]_{i,i}[\bOmega_1^*]_{j,j})^2}   \bigg  )$$
in distribution. 

(ii). Assume Condition \ref{con:eigenvalue_tensor} and $\log  m_1 = o(nm/m_1)$ hold, we have
$$ \max \limits_{i,j \in \{ 1, \ldots, m_1\}} \bigg | [\Xi_1]_{ij ; \tilde{\xi}} - \frac{m_1 \prod_{k \in \{2, \ldots , K\} }\tr(\bSigma_k^*)}{m}     \frac{[\bOmega^*_1]_{i , j }}{[\bOmega^*_1]_{i,i }[\bOmega^*_1]_{j , j } }  \bigg | = O_p  \bigg (  \sqrt{\frac{\log m_1 }{nm/m_1}} \bigg ) . $$
\end{lemma}
\noindent{{\bf Proof of Lemma \ref{lem: lem3}}:} To ease the presentation, we show that Lemma \ref{lem: lem3} holds when $K=3$. The proof can easily be generalized to the case with $K>3$.

Let $\bxi_{l;i} = (\xi_{l;i, i_2, i_3})_{i_2 \in \{1 , \ldots , m_2\} , i_3 \in \{1 , \ldots , m_3\}} \in \RR^{m_2 \times m_3}$. We prove it by constructing a tensor normal variable that combines $\bxi_{l;i}$ and $\bxi_{l;j}$, then following similar strategy in Lemma \ref{lem: lem1} to get its mean and variance, finally applying Lindeberg-Feller central limit theorem and Lemma 1 in \cite{cai2011adaptive} to get convergency results.

Define $\Ub_{l ; i j} \in \RR^{2 \times m_2 \times m_3}$ with $[\Ub_{l; ij}]_{1, : ,:} = \bxi_{l;i}$ and $[\Ub_{l; ij}]_{2, : ,:} = \bxi_{l;j}$.
From Lemma 1 in \cite{peng2012}, we have 
 \[
\Ub_{l;[ij]}  \sim  \textrm{TN} \bigg{\{} \bm{0} ; 
  \begin{pmatrix}
    [\bOmega^*_1]_{i , i }^{-1} &    \frac{[\bOmega^*_1]_{i , j }}{[\bOmega^*_1]_{i,i }[\bOmega^*_1]_{j , j }}\\
    \frac{[\bOmega^*_1]_{i , j }}{[\bOmega^*_1]_{i,i }[\bOmega^*_1]_{j , j } } &      [\bOmega^*_1]_{j , j }^{-1}
  \end{pmatrix} , \bSigma_2^*, \bSigma_3^* \bigg{\}}.
\]
Under similar arguments in the proof of Lemma \ref{lem: lem1} \eqref{eqn: xi_trans}, we have 
$$
 \sum \limits_{l=1}^{n} \sum \limits_{i_2=1}^{m_2}  \sum \limits_{i_3=1}^{m_3} \tilde{\xi}_{l;i, i_2, i_3} \tilde{\xi}_{l; j, i_2, i_3} = \sum \limits_{l=1}^{n-1} \sum \limits_{i_2=1}^{m_2}  \sum \limits_{i_3=1}^{m_3} \lambda_{i_2}^{(2)}\lambda_{i_3}^{(3)} \zeta_{l; i , i_2 ,i_3}\zeta_{l; j , i_2 ,i_3}, 
$$
where 
 \[
(\zeta_{l; i , i_2 ,i_3} , \zeta_{l; j , i_2 ,i_3} )   \sim  \textrm{TN} \bigg{\{} \bm{0} ; 
  \begin{pmatrix}
    [\bOmega^*_1]_{i , i }^{-1} &    \frac{[\bOmega^*_1]_{i , j }}{[\bOmega^*_1]_{i,i }[\bOmega^*_1]_{j , j }}\\
    \frac{[\bOmega^*_1]_{i , j }}{[\bOmega^*_1]_{i,i }[\bOmega^*_1]_{j , j } } &      [\bOmega^*_1]_{j , j }^{-1}
  \end{pmatrix}  \bigg{\}}
\]
i.i.d. for $1 \le l \le n, \; 1 \le i_2 \le m_2 , \; 1 \le i_3 \le m_3$. Therefore, its expectation is
\begin{align*}
\Eb \bigg [ \sum \limits_{l=1}^{n} \sum \limits_{i_2=1}^{m_2}  \sum \limits_{i_3=1}^{m_3} \tilde{\xi}_{l;i, i_2, i_3} \tilde{\xi}_{l; j, i_2, i_3} \bigg ] & = \sum \limits_{l=1}^{n-1} \sum \limits_{i_2=1}^{m_2}  \sum \limits_{i_3=1}^{m_3} \lambda_{i_2}^{(2)}\lambda_{i_3}^{(3)}  \frac{[\bOmega^*_1]_{i , j }}{[\bOmega^*_1]_{i,i }[\bOmega^*_1]_{j , j }}   \\
& = (n-1) \tr (\bSigma_2^*)\tr (\bSigma_3^*) \frac{[\bOmega^*_1]_{i , j }}{[\bOmega^*_1]_{i,i }[\bOmega^*_1]_{j , j }}
\end{align*}
Applying Lemma 1 in \cite{cai2011adaptive}, (ii) is proved.
Moreover, its variance is 
\begin{align*}
\Var \bigg( \sum \limits_{l=1}^{n} \sum \limits_{i_2=1}^{m_2}  \sum \limits_{i_3=1}^{m_3} \tilde{\xi}_{l;i, i_2, i_3} \tilde{\xi}_{l; j, i_2, i_3} \bigg) & = \Var \bigg ( \sum \limits_{l=1}^{n-1} \sum \limits_{i_2=1}^{m_2}  \sum \limits_{i_3=1}^{m_3} \lambda_{i_2}^{(2)}\lambda_{i_3}^{(3)} \zeta_{l; i , i_2 ,i_3}\zeta_{l; j , i_2 ,i_3} \bigg ) \\
& =   \sum \limits_{l=1}^{n-1} \sum \limits_{i_2=1}^{m_2}  \sum \limits_{i_3=1}^{m_3}  \Var \bigg (\lambda_{i_2}^{(2)}\lambda_{i_3}^{(3)} \zeta_{l; i , i_2 ,i_3}\zeta_{l; j , i_2 ,i_3} \bigg ) \\
& = (n-1)\sum \limits_{i_2=1}^{m_2}  \sum \limits_{i_3=1}^{m_3}   (\lambda_{i_2}^{(2)})^2 ( \lambda_{i_3}^{(3)})^2 \Var \bigg ( \zeta_{l; i , i_2 ,i_3}\zeta_{l; j , i_2 ,i_3} \bigg ) \\
& = (n-1) \bigg (  \frac{1}{[\bOmega_1^*]_{i,i} [\bOmega_1^*]_{j,j}} + \frac{([\bOmega_1^*]_{i,j})^2}{([\bOmega_1^*]_{i,i}[\bOmega_1^*]_{j,j})^2}      \bigg )  \| \bSigma_2^*\|_F^2 \| \bSigma_3^*\|_F^2 \\
\end{align*}
By Lindeberg-Feller central limit theorem, we prove the limiting distribution in (i). The proof is completed.
\hfill $\blacksquare$\\

\begin{lemma} \label{lem: Ahat_consistency}
Assume i.i.d. tensor data ${\cal T}, {\cal T}_1,\ldots, {\cal T}_n \in \mathbb R^{m_1 \times m_2 \times \cdots \times m_K}$ follows the tensor normal distribution $\textrm{TN}({\bf0}; \bSigma_1^*, \ldots, \bSigma_K^*)$. Let 
$$\varpi_0^2 =  \frac{m \cdot \|\bSigma_2^*\|_F^2 \cdots \|\bSigma_K^*\|_F^2 }{ m_1 \cdot (\tr(\bSigma_2^*))^2 \cdots (\tr(\bSigma_K^*))^2}.$$
Under the same conditions of Lemma \ref{lemma:sample_cov_max} and $(m_k +s_k ) \log m_k = o(nm)$ for all $k \in \{ 2 , \ldots , K \}$, we have 
$$\varpi^2  /\varpi_0^2  \rightarrow 1 ,$$
in probability as $nm \rightarrow \infty$ where $\varpi^2 $ is defined in \S\ref{sec: construct}. 
\end{lemma}

\noindent{{\bf Proof to Lemma \ref{lem: Ahat_consistency}}:}
To ease the presentation, we show that Lemma \ref{lem: Ahat_consistency} holds when $K=3$. The proof can easily be generalized to the case with $K>3$.

Lemma \ref{lemma:sample_cov_max} gives
$$\max_{s,t} \big[ \widehat{\Sbb}_k -  \bSigma_k^* \big]_{s,t} = O_P\left( \max_{j=1,\ldots, K}  \sqrt{ \frac{(m_j+s_j) \log m_j }{nm} } \right).$$
For simplicity, denote 
$$ a_n =\max_{j=1,\ldots, K}  \sqrt{ \frac{(m_j+s_j) \log m_j }{nm} } .$$
Then we have 
$$\tr ( \widehat{\Sbb}_k ) / \tr ( \bSigma_k^*)  = 1+ O_p ( a_n ) \; \text{, and } \;  \|\widehat{\Sbb}_k\|_F^2 / \| \bSigma_k^* \|_F^2 = 1 + O_p(a_n).$$
Therefore
$$\frac{\| \bSigma_k^* \|_F^2}{ \|\widehat{\Sbb}_k\|_F^2 } \cdot \frac{(\tr ( \widehat{\Sbb}_k ))^2}{(\tr ( \bSigma_k^*))^2} \rightarrow 1 $$
in probability. The proof is completed.  \hfill $\blacksquare$\\

\begin{lemma}\label{lem: beta_consistency}
Under the same conditions of Theorem \ref{thm:maxnorm}, we have 
$$\max \limits_{i_1 \in \{1 ,\ldots , m_1\}}\|\hat{\btheta}_{i_1} - \btheta_{i_1}\|_2 =  \sqrt{\frac{d_1 m_1 \log m_1 }{nm}}
\; \text{, and} \; \max \limits_{i_1 \in \{1 ,\ldots , m_1\}}\|\hat{\btheta}_{i_1} - \btheta_{i_1}\|_1 = d_1 \sqrt{\frac{m_1 \log m_1 }{nm}},$$
where $\hat{\btheta}_{i_1}$ and $\btheta_{i_1}$ are defined in \S\ref{sec: construct}.
\end{lemma}
\noindent{{\bf Proof of Lemma \ref{lem: beta_consistency}}:}
To ease the presentation, we show that Lemma \ref{lem: beta_consistency} holds when $K=3$. The proof can easily be generalized to the case with $K>3$.

First, we derive upper bounds on $l_2$ norm.
\begin{align} \notag 
\max \limits_{i_1 \in \{1 ,\ldots , m_1\}}\|\hat{\btheta}_{i_1} - \btheta_{i_1}\|_2 & = \max \limits_{i_1 \in \{1 ,\ldots , m_1\}} \| [ \widehat{\bOmega}_1]^{-1}_{i_1 i_1}[ \widehat{\bOmega}_1]_{i_1 , - i_1} - [ \bOmega^*_1]^{-1}_{i_1 i_1}[\bOmega_1^*]_{i_1 , - i_1} \|_2 \\  \notag 
& \le \max \limits_{i_1 \in \{1 ,\ldots , m_1\}}  \|[\bOmega_1^*]_{i_1 , - i_1}\|_2 | [ \widehat{\bOmega}_1]^{-1}_{i_1 i_1} - [\bOmega_1^*]^{-1}_{i_1 i_1} | \\  \notag 
&\quad \quad\quad\quad + \max \limits_{i_1 \in \{1 ,\ldots , m_1\}}  \|[\widehat{\bOmega}_1]_{i_1 , - i_1} - [\bOmega_1^*]_{i_1 , - i_1}\|_2 | [ \widehat{\bOmega}_1]^{-1}_{i_1 i_1}  |    \\  \notag
& \le \max \limits_{i_1 \in \{1 , \ldots , m_1\}}\| [ \bOmega_1^*]_{i_1, :} \|_2\max \limits_{i,j \in \{1 ,\ldots , m_1\}} | [ \widehat{\bOmega}_1]^{-1}_{i j} - [\bOmega_1^*]^{-1}_{i j} |  \\  
& \quad\quad\quad\quad + \max \limits_{i_1 \in \{1 ,\ldots , m_1\}}  \|[\widehat{\bOmega}_1]_{i_1 , :} - [\bOmega_1^*]_{i_1 , :}\|_2   \max \limits_{i,j \in \{1 ,\ldots , m_1\}} | [ \widehat{\bOmega}_1]^{-1}_{i j}  |
\end{align}
Theorem \ref{thm:maxnorm} implies that 
$$\big\| \widehat{\bOmega}_1 -  \bOmega_1^* \big\|_{\infty} = O_P\left(\sqrt{\frac{m_1 \log m_1}{n m }}\right).$$
Together with the fact $\|\bOmega_1^*\|_F= 1$ and $\|\hat{\bOmega}_1^*\|_F= 1$, $l_2$ norm is bounded as 
$$\max \limits_{i_1 \in \{1 ,\ldots , m_1\}}\|\hat{\btheta}_{i_1} - \btheta_{i_1}\|_2 \le C_1 \sqrt{\frac{\log m_1}{n m_2 m_3 }} +  C_2 \sqrt{d_1 }\sqrt{ \frac{ \log m_1 }{nm_2 m_3}}$$

Similarly, we bound $l_1$ norm as follow:
\begin{align} \notag 
\max \limits_{i_1 \in \{1 ,\ldots , m_1\}}\|\hat{\btheta}_{i_1} - \btheta_{i_1}\|_1 & = \max \limits_{i_1 \in \{1 ,\ldots , m_1\}} \| [ \widehat{\bOmega}_1]^{-1}_{i_1 i_1}[ \widehat{\bOmega}_1]_{i_1 , - i_1} - [ \bOmega^*_1]^{-1}_{i_1 i_1}[\bOmega_1^*]_{i_1 , - i_1} \|_1 \\  \notag 
& \le \smallvertiii{ \bOmega_1^*}_\infty \max \limits_{i,j \in \{1 ,\ldots , m_1\}} | [ \widehat{\bOmega}_1]^{-1}_{i j} - [\bOmega_1^*]^{-1}_{i j} | \\\notag
&\quad\quad\quad\quad\quad\quad + \smallvertiii{ \widehat{\bOmega}_1 - \bOmega_1^*}_\infty  \max \limits_{i,j \in \{1 ,\ldots , m_1\}} | [ \widehat{\bOmega}_1]^{-1}_{i j}  |\\
& \le C_1 \sqrt{\frac{ \log m_1}{n m_2m_3 }} +  C_2  d_1 \sqrt{ \frac{\log m_1 }{nm_2m_3}}
\end{align}
The proof is completed.     \hfill $\blacksquare$\\

\section{Auxiliary lemmas}
\label{sec:other_lemma}

\begin{lemma}
\label{lemma:compute_cov}
Assume a random matrix $\Xb \in \mathbb R^{p\times q}$ follows the matrix-variate normal distribution such that $\textrm{vec}(\Xb) \sim \textrm{N}(\textbf{0}; \bPsi^* \otimes \bSigma^*)$ with $\bPsi^* \in \mathbb R^{q\times q}$ and $\bSigma^* \in \mathbb R^{p\times p}$. Then for any symmetric and positive definite matrix $\bOmega \in \mathbb R^{p\times p}$, we have $\EE(\Xb^{\top} \bOmega \Xb) = \bPsi^{*} \textrm{tr}(\bOmega \bSigma^*)$.
\end{lemma}

\noindent{\bf Proof of Lemma \ref{lemma:compute_cov}:} Since the matrix $\bOmega$ is symmetric and positive definite, it has the Cholesky decomposition $\bOmega = \Vb^{\top}\Vb$, where $\Vb$ is upper triangular with positive diagonal entries. Let $\Yb := \Vb \Xb$ and denote the $j$-th row of matrix $\Yb$ as $\yb_j = (y_{j,1},\ldots, y_{j,q})$. We have $\EE(\Xb^{\top} \bOmega \Xb) = \EE(\Yb^{\top} \Yb) = \sum_{j=1}^p \EE(\yb_j^{\top} \yb_j)$. Here $\yb_j = \vb_j \Xb$ with $\vb_j$ the $j$-th row of $\Vb$. Denote the $i$-th column of matrix $\Xb$ as $\xb_{(i)}$, we have $y_{j,i} = \vb_j \xb_{(i)}$. Therefore, the $(s,t)$-th entry of $\EE(\yb_j^{\top} \yb_j)$ is 
$$
\bigl[ \EE(\yb_j^{\top} \yb_j) \bigr]_{(s,t)} = \EE[ \vb_j \xb_{(s)} \vb_j \xb_{(t)} ] = \vb_j \EE[  \xb_{(s)}  \xb_{(t)}^{\top} ] \vb_j^{\top} = \vb_j \bPsi^*_{s,t} \bSigma^* \vb_j^{\top},
$$
where $\bPsi^*_{s,t}$ is the $(s,t)$-th entry of $\bPsi^*$. The last equality is due to $\textrm{vec}(\Xb) = (\xb_{(1)}^{\top}, \ldots, \xb_{(q)}^{\top})^{\top} \sim \textrm{N}(\textbf{0}; \bPsi^* \otimes \bSigma^*)$  Therefore, we have
$$
\EE(\Xb^{\top} \bOmega \Xb) = \sum_{j=1}^p \EE(\yb_j^{\top} \yb_j) = \bPsi^*  \sum_{j=1}^p  \vb_j \bSigma^* \vb_j^{\top} = \bPsi^*  \textrm{tr} \Big( \sum_{j=1}^p \vb_j^{\top} \vb_j \bSigma^*\Big) = \bPsi^{*} \textrm{tr}(\bOmega \bSigma^*).
$$
This ends the proof of Lemma \ref{lemma:compute_cov}. \hfill $\blacksquare$\\

The following lemma is stated by \cite{ledoux2011}.

\begin{lemma}
\label{lemma:ledoux}
Let random variables $x_1,\ldots,x_n \in \mathbb R$ be i.i.d. drawn from standard normal $\textrm{N}(0;1)$ and denote $\xb = (x_1,\ldots,x_n)^{\top} \in \mathbb R^n$ be a random vector. For a function $f: \mathbb R^n \rightarrow \mathbb R$ with Lipschitz constant $L$, that is, for any vectors $\vb_1,\vb_2\in \mathbb R^n$, there exists $L\ge 0$ such that $|f(\vb_1) - f(\vb_2)| \le L \|\vb_1 - \vb_2\|_2$. Then, for any $t>0$, we have
$$
\mathbb P\left\{ |f(\xb) - \EE[f(\xb)]| > t \right\} \le 2 \exp\left(-\frac{t^2}{2L^2}\right).
$$
\end{lemma}

The following lemma is useful for the proof of Lemma \ref{lemma:sample_cov}. A similar statement was given in Lemma I.2 of \cite{negahban2011}.

\begin{lemma}
\label{lemma:NW11}
Suppose that a $d$-dimensional Gaussian random vector $\yb \sim \textrm{N}(0;\Qb)$, Then, for any $t > 2/{\sqrt{d}}$, we have
$$
\mathbb P\Big[ \frac{1}{d} \big| \|\yb\|_2^2  - \EE(\|\yb\|_2^2 ) \big| > 4 t \|\Qb\|_2 \Big] \le 2 \exp\biggl\{-\frac{d \big(t-{2}/{\sqrt{d}}\big)^2}{2}\biggr\} + 2 \exp(-d/2).
$$
\end{lemma}

\noindent{\bf Proof of Lemma \ref{lemma:NW11}:} Note that $\EE(\|\yb\|_2^2 ) \le [\EE(\|\yb\|_2 )]^2$ and hence
$$
\|\yb\|_2^2  - \EE(\|\yb\|_2^2 ) \le [\|\yb\|_2  - \EE(\|\yb\|_2 )][\|\yb\|_2  + \EE(\|\yb\|_2 )].
$$
The term $(\|\yb\|_2  - \EE(\|\yb\|_2 )$ can be bounded via the concentration inequality in Lemma \ref{lemma:ledoux} by noting that $\|\yb\|_2$ is a Lipschitz function of Gaussian random vector $\yb$. The term $\|\yb\|_2  + \EE(\|\yb\|_2 )$ can also be bounded by the large deviation bound since $\yb$ is a Gaussian random vector. This ends the proof of Lemma \ref{lemma:NW11}. \hfill $\blacksquare$\\

The following lemma, which is key to Theorem \ref{thm: FDR_control}, is Lemma 6.4 in \cite{chen2015}. Let $\bxi_1 , \ldots , \bxi_n$ be independent $d$-dimensional random vectors with mean zero. Define $G(t)=2 - 2 \Phi(t)$ and $|\cdot|_{(d)}$ by $|\bz|_{(d)} = \min \{|z_i|; 1 \le i \le d  \}$ for $\bz = (z_1 ,\ldots , z_d)^\top$. Let $(p,n)$ be a sequence of positive integers and the constants $c,r,b,\gamma,K,C$ mentioned below is irrelevant to $(p,n)$.
\begin{lemma} \label{lem: liu2013lem1}
Suppose that $p \le cn^r$ and $\max_{1\le k \le n} \EE \|\bxi_k\|_2^{bdr+2+\epsilon} \le K$ for some fixed $c > 0$, $r > 0$, $b > 0$, $K>0$ and $\epsilon > 0$. Assume that $\|\frac{1}{n} \textrm{cov}(\sum_{k=1}^n \bxi_k) - \Ib_{d\times d} \|_2 \le C(\log p )^{-2-\gamma}$ for some $\gamma > 0 $ and $C > 0$. Then we have 
$$\sup \limits_{0 \le t \le \sqrt{b\log p}} \bigg |  \frac{P(\big|  \sum_{k=1}^n \bxi_k  \big |_{(d)} \ge t\sqrt{n})}{(G(t))^d} -1 \bigg | \le C(\log p)^{-1-\gamma_1},$$
 where $\gamma_1 = \min \{ \gamma, 1/2 \}$.
\end{lemma}

The following lemma is Lemma 6.5 in \cite{chen2015}, serving a significant role in Theorem \ref{thm: FDR_control}. Let $\eta_k = (\eta_{k1},\eta_{k2})^\top$, $1\le k \le n$, are independent $2$-dimensional random vectors with mean zero. 
\begin{lemma} \label{lem: liu2013lem2}
Suppose that $p \le cn^r$ and $\max_{1\le k \le n} \EE \|\theta_k\|_2^{bdr+2+\epsilon} \le \infty$ for some fixed $c > 0$, $r > 0$, $b > 0$ and $\epsilon > 0$. Assume that $\sum_{k=1}^n \Var (\eta_{k1})= \sum_{k=1}^n \Var (\eta_{k2}) =n $ and $|\frac{1}{n}\sum_{k=1}^n \textrm{cov} (\eta_{k1},\eta_{k2}) | \le \delta$ for some $0 \le \delta < 1$. Then, we have 
$$P\bigg (  |\sum\limits_{k=1}^{n} \eta_{k1}| \ge t\sqrt{n}, |\sum\limits_{k=1}^{n} \eta_{k2}| \ge t\sqrt{n}  \bigg ) \le C (t+1)^{-2} \exp ( -t^2/(1+\delta) )$$
uniformly for $0 \le t \le \sqrt{b\log p }$, where $C$ only depend on $c,d,r,\epsilon, \delta$.
\end{lemma}


\section{Sensitivity Analysis of Tuning Parameter}
\label{sec:sensitivity}
In this section, additional numerical results are presented of sensitivity analysis for $C$ in tuning parameter $\lambda_k = C \sqrt{{\log m_k}/(n m m_k)}$ of updating $\hat{\bOmega}$.

We compare the estimation accuracy of Tlasso algorithm under different choices of $C$. 
In particular, $C$ is chosen from $\{10,15,20,25,30\}$. 
Two simulations are considered, i.e., triangle graph and nearest neighbor graph. 
The scenarios of interest are s1, s2, and s3. Each repeats 100 times. The rest setup is the same as in \S\ref{sec:simulation}. 

Three criteria are selected to measure estimation accuracy. 
The first one is Frobenius estimation error of Kronecker product of precision matrices, see \eqref{eq:kro_F}. 
The rest two are the averaged estimation errors in Frobenius norm and max norm, see \eqref{eq:ave_measure}. 

\begin{figure}[h!]
\centering
\includegraphics[scale=0.2]{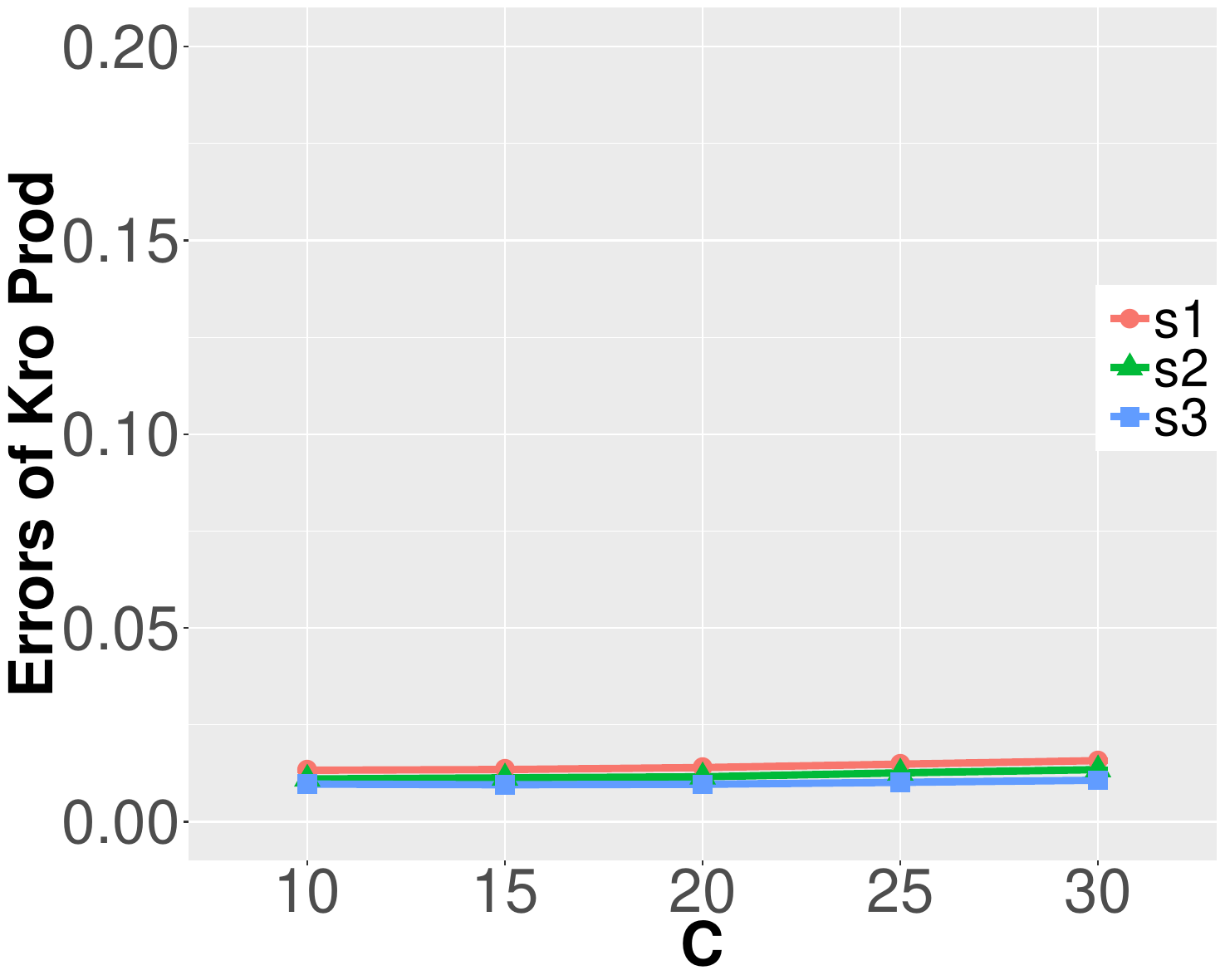}
\includegraphics[scale=0.2]{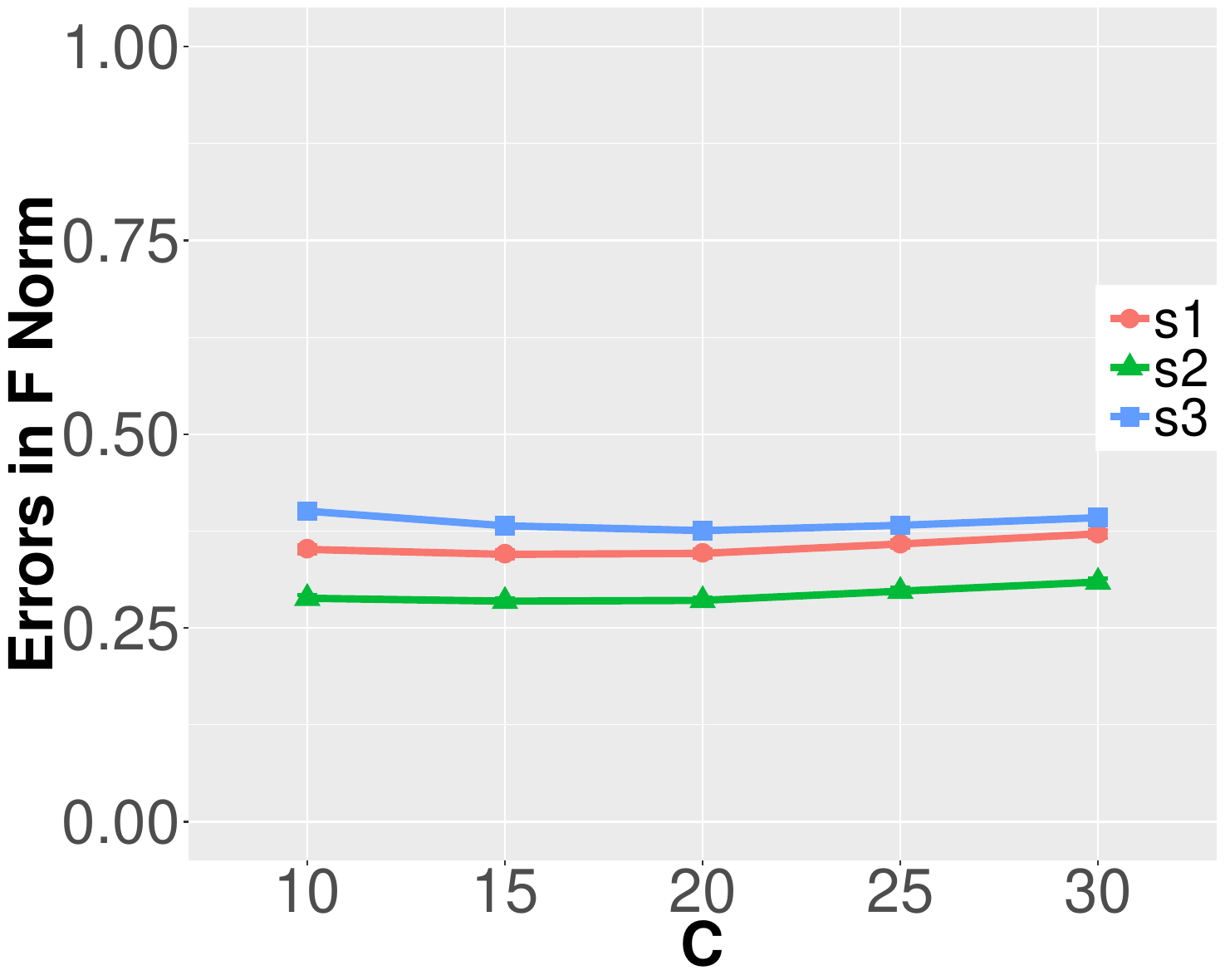}
\includegraphics[scale=0.2]{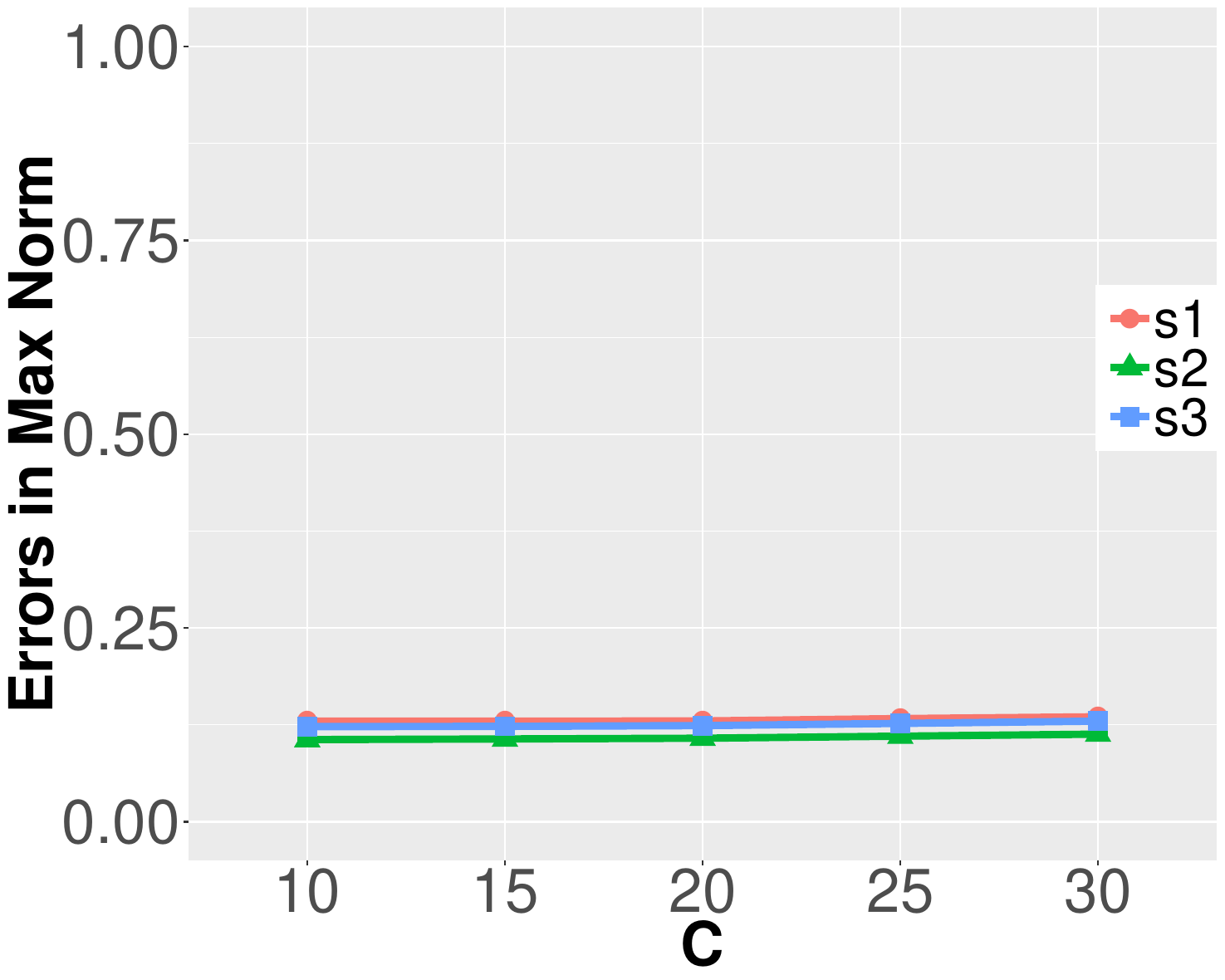}
\includegraphics[scale=0.2]{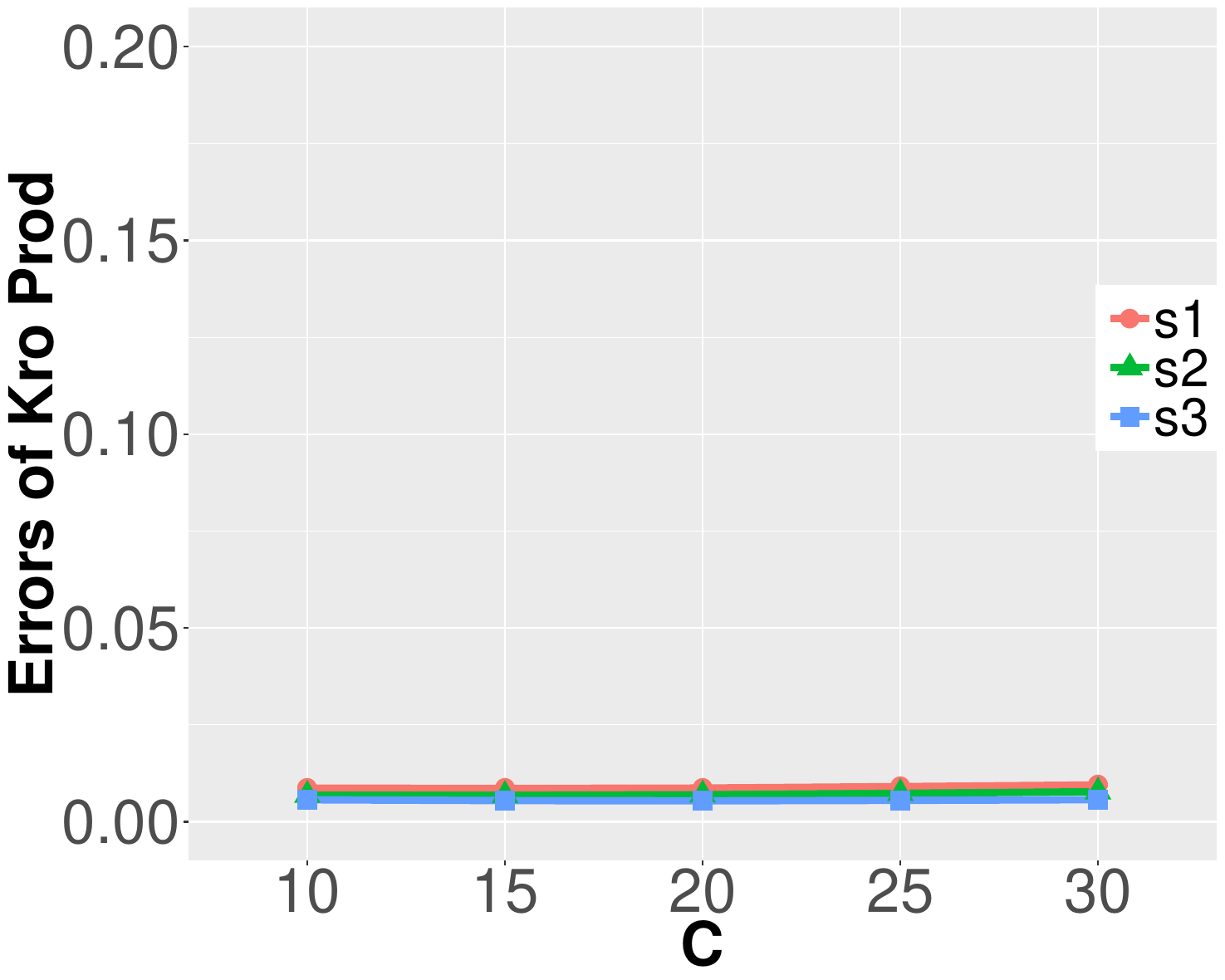}
\includegraphics[scale=0.2]{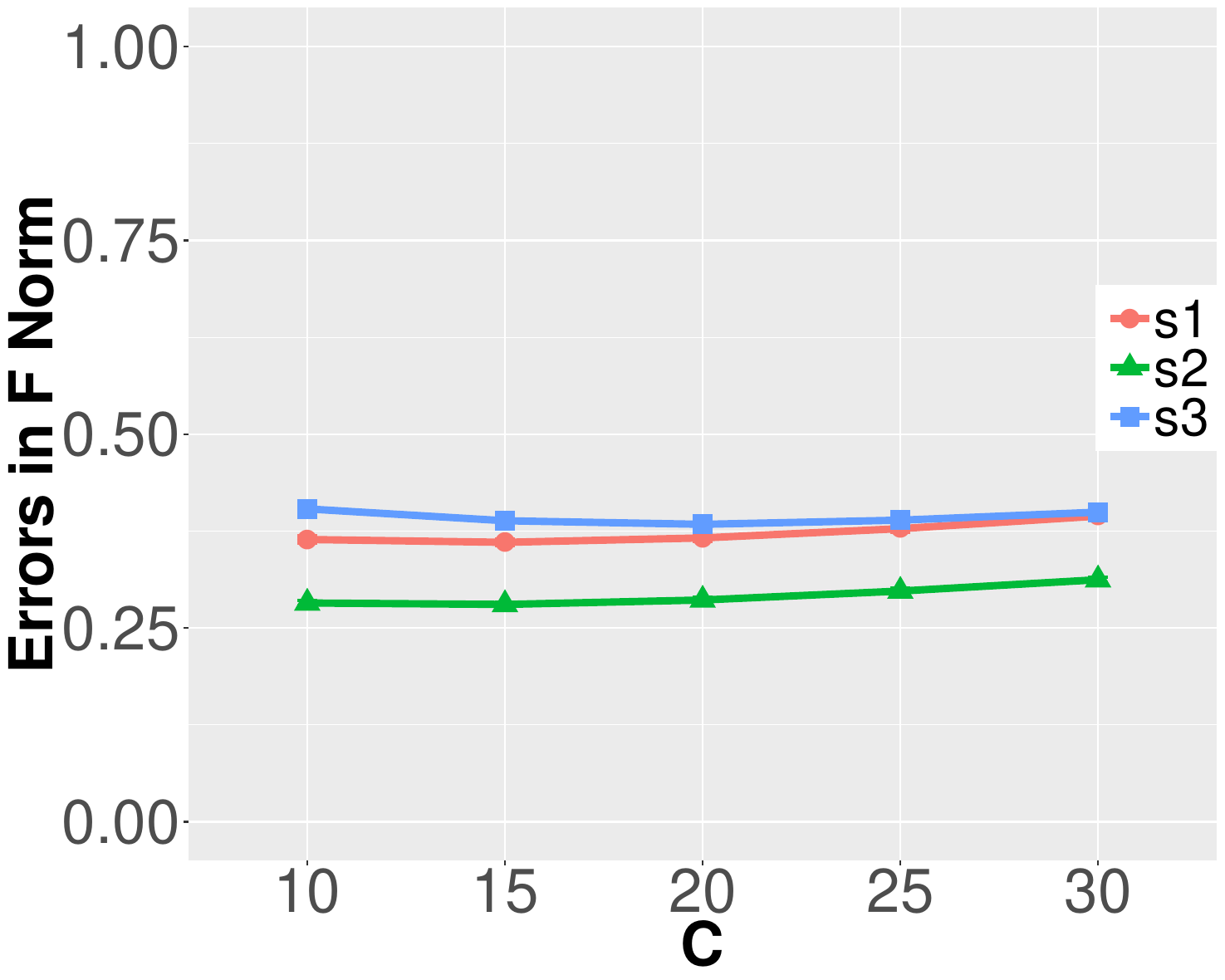}
\includegraphics[scale=0.2]{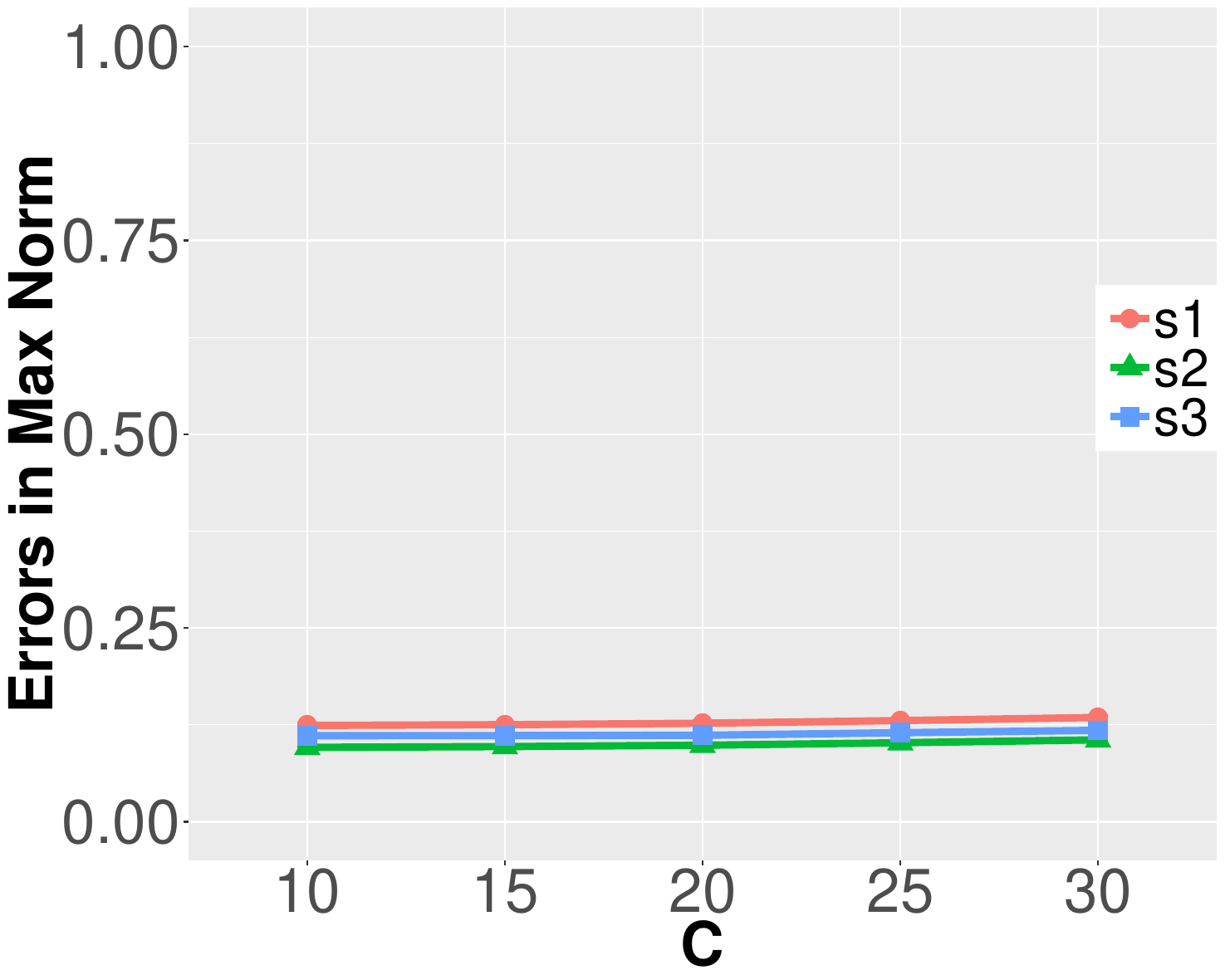}
\caption{Estimation errors of Tlasso under the choices of $C\in\{10,15,20,25,30\}$. From left to right column is the Frobenius error of Kronecker product of precision matrices, the averaged Frobenius error and the averaged max error. The first row is simulation 1, and the second  is simulation 2.}
\label{fig: sensitivity_estimation}
\vskip 0em
\end{figure}

Estimation accuracy of Tlasso under different values of $C$ are depicted in Figure \ref{fig: sensitivity_estimation}. 
The nearly horizontal lines in the figure demonstrates that Tlasso yields almost the same performance when varying $C$. 
This phenomenon suggests that the performance of Tlasso is not sensitive to the choice of $C$.
Therefore, we will set $C=20$ through the simulations and real data analysis in this paper.

\section{Effect of Sample Size and Dimensionality} 
\label{sec:add_simulation}

In this section, additional numerical results of the proposed inference procedure are presented on new scenarios to study the effects of sample size and dimensionality.

\noindent {\bf Sample size:}
We first show the results on the scenarios of varying sample size. 
Similarly,  two simulations are considered, i.e., triangle graph and nearest neighbor graph. 
Estimation of precision matrices is conducted by Tlasso algorithm under the same setting as in \S\ref{sec:simulation}. 
To study the effect of sample size, the following five scenarios are compared: 
\begin{itemize}
\item \textbf{Scenario s4:} sample size $n = 10$ and dimension $(m_1,m_2,m_3) = (10,10,10)$.
\item \textbf{Scenario s5:} sample size $n = 20$ and dimension $(m_1,m_2,m_3) = (10,10,10)$.
\item \textbf{Scenario s6:} sample size $n = 30$ and dimension $(m_1,m_2,m_3) = (10,10,10)$.
\item \textbf{Scenario s7:} sample size $n = 100$ and dimension $(m_1,m_2,m_3) = (10,10,10)$.
\item \textbf{Scenario s8:} sample size $n = 150$ and dimension $(m_1,m_2,m_3) = (10,10,10)$.
\end{itemize}
Each scenario repeats 100 times.

\begin{figure}[h!]
\centering
\includegraphics[scale=0.16]{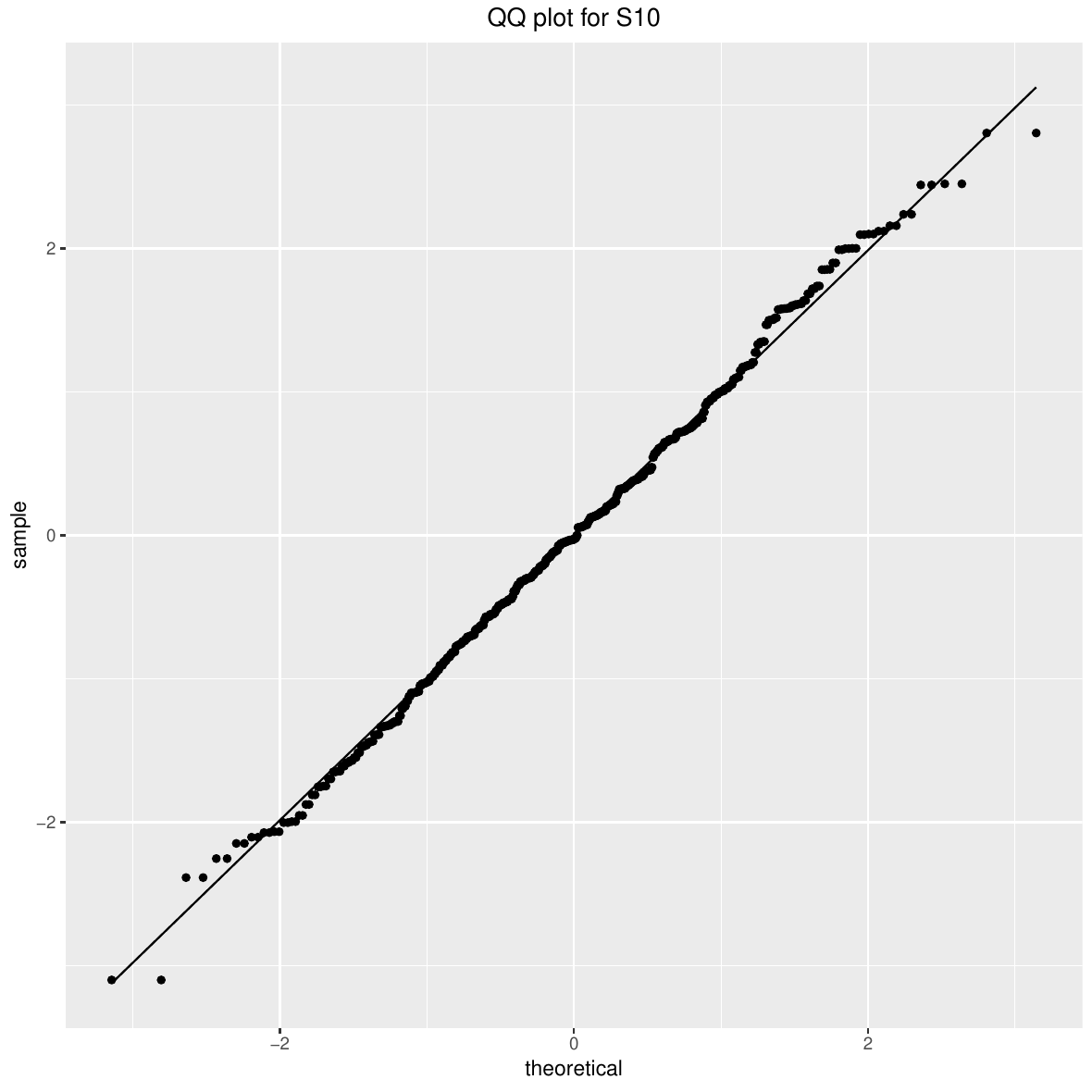}
\includegraphics[scale=0.16]{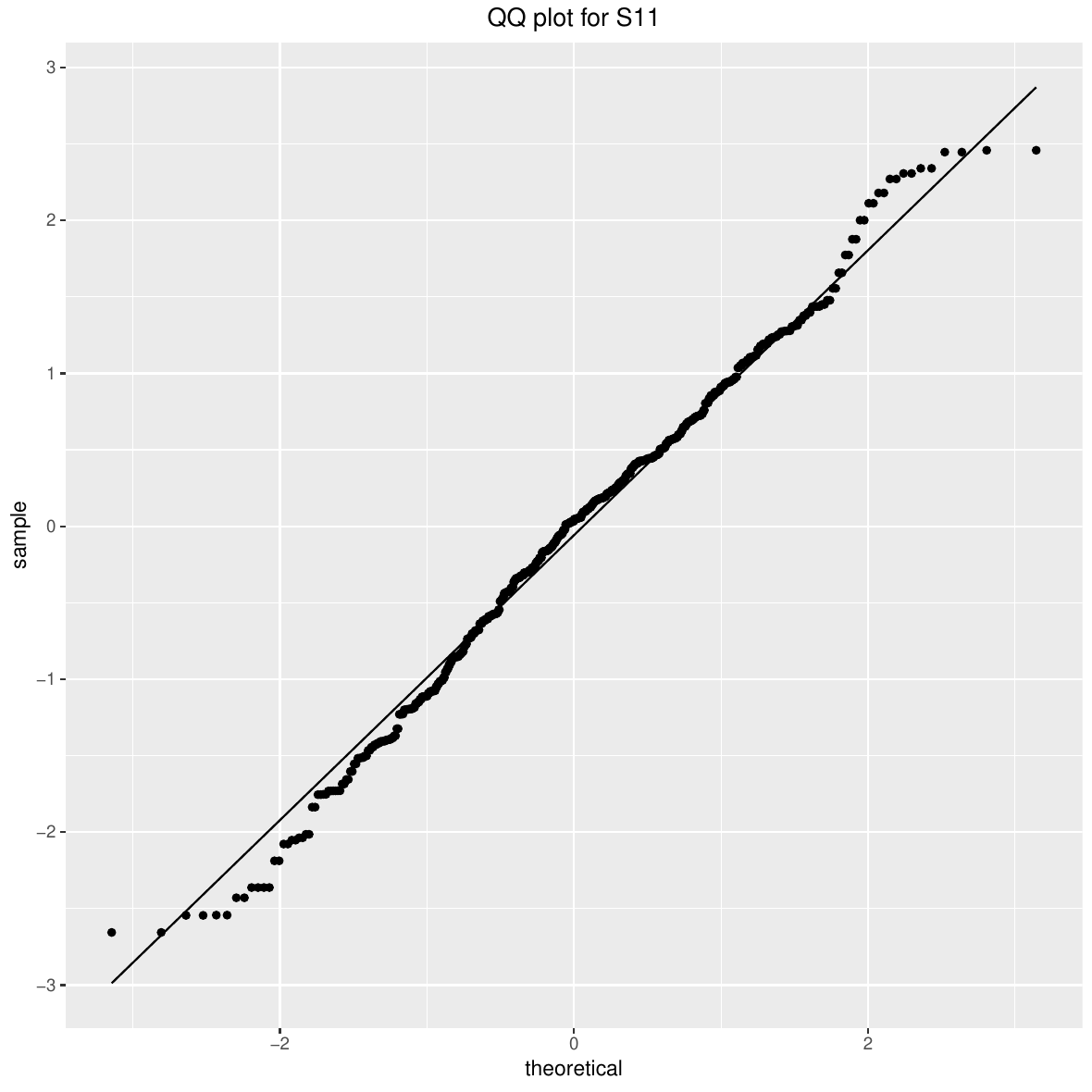}
\includegraphics[scale=0.16]{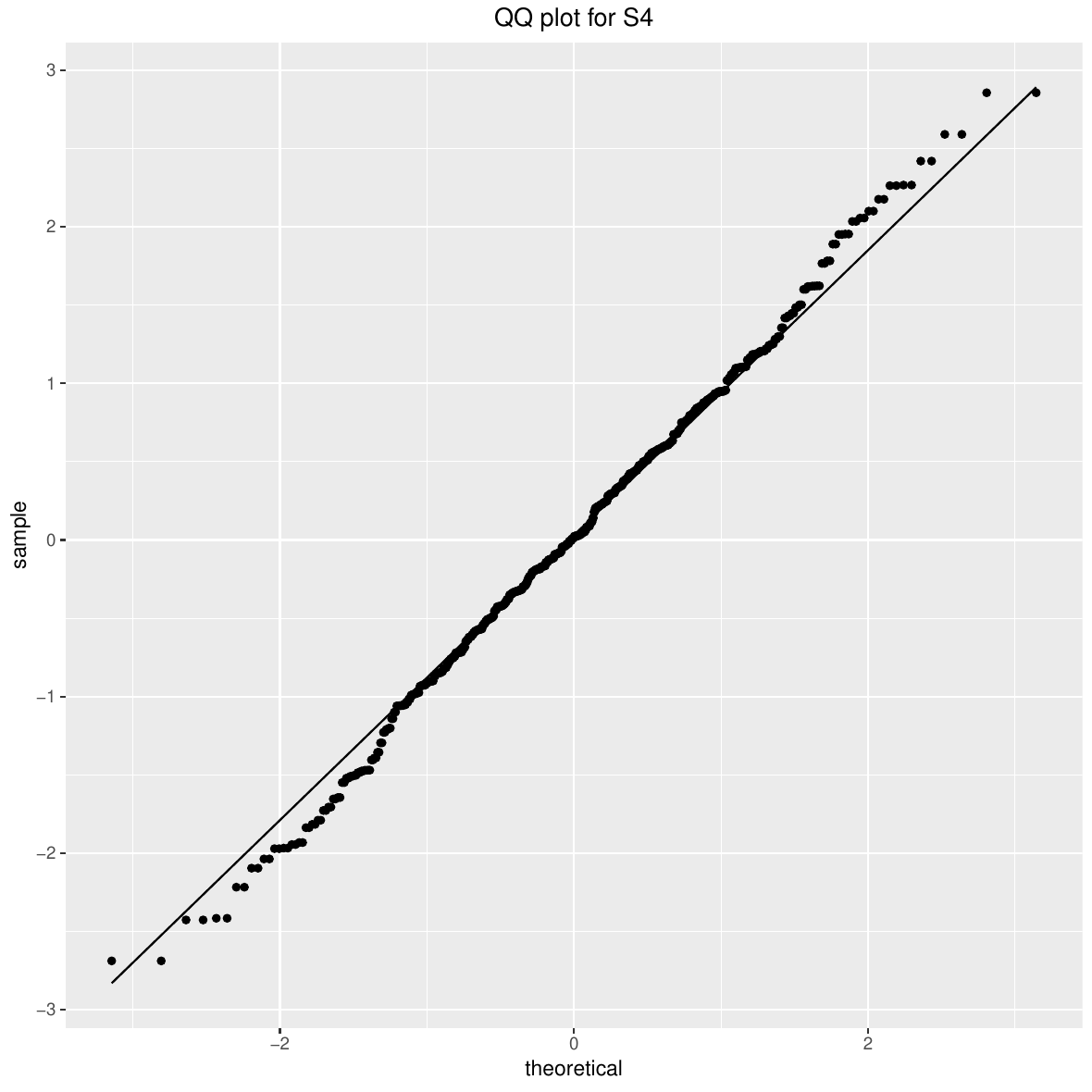}
\includegraphics[scale=0.16]{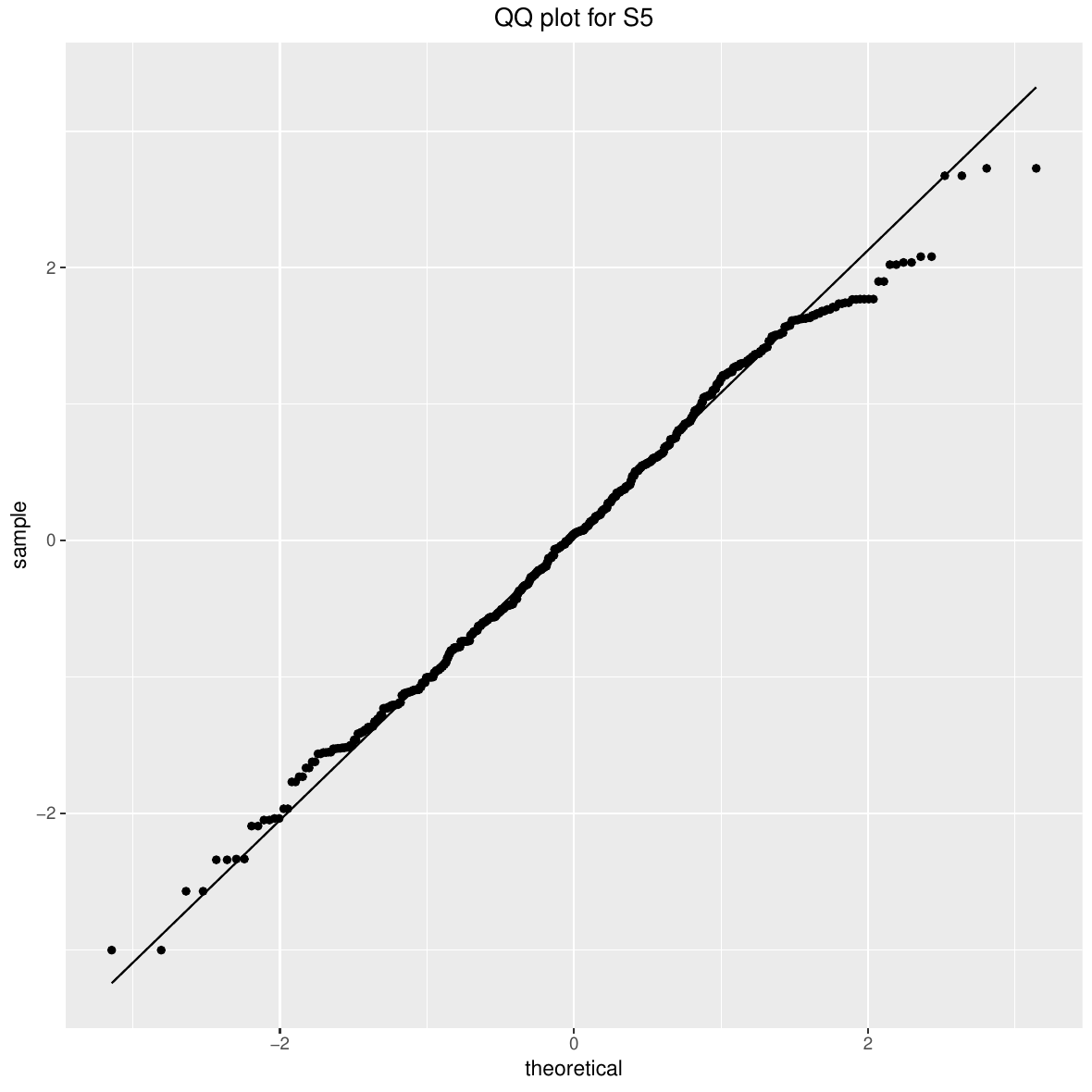}
\includegraphics[scale=0.16]{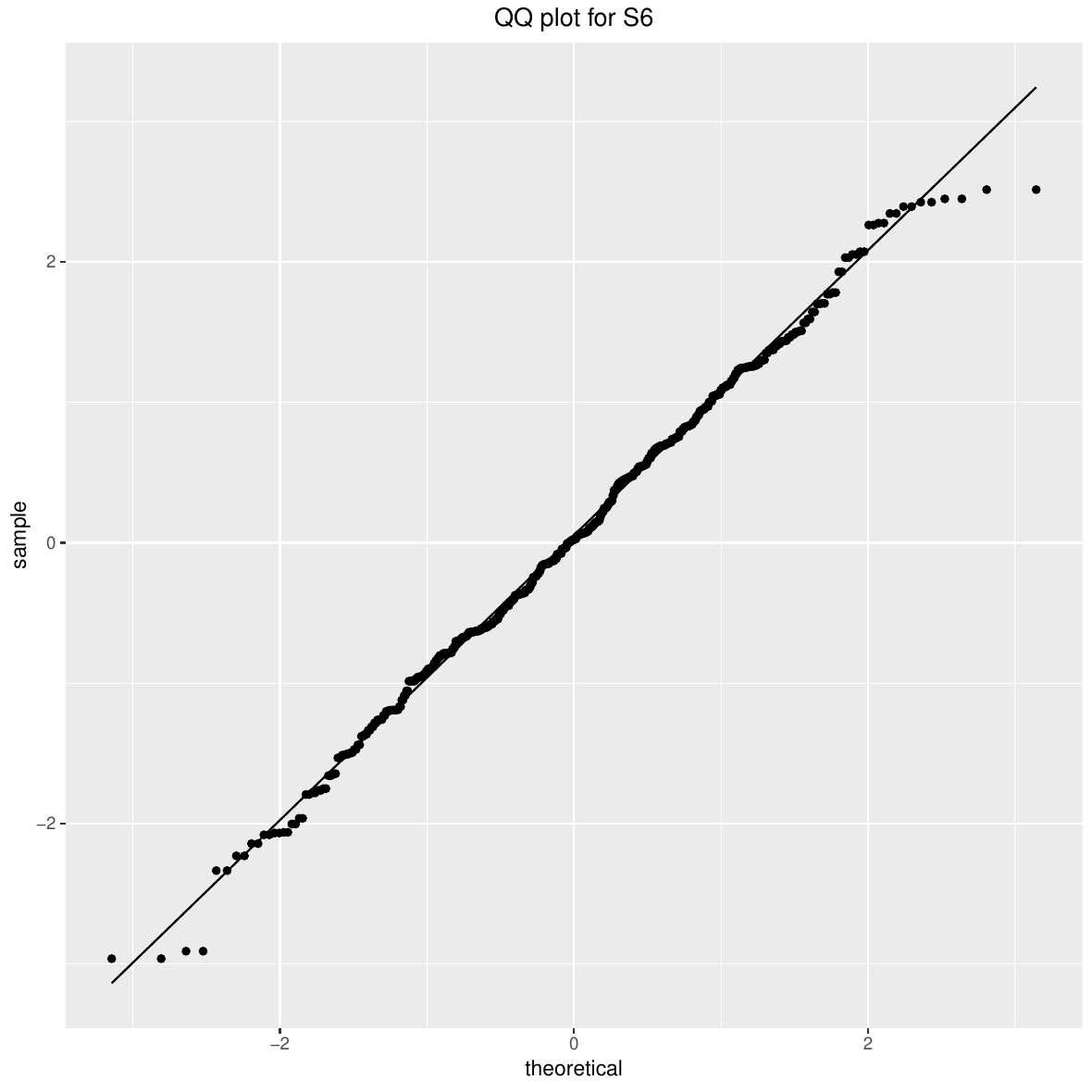}
\includegraphics[scale=0.16]{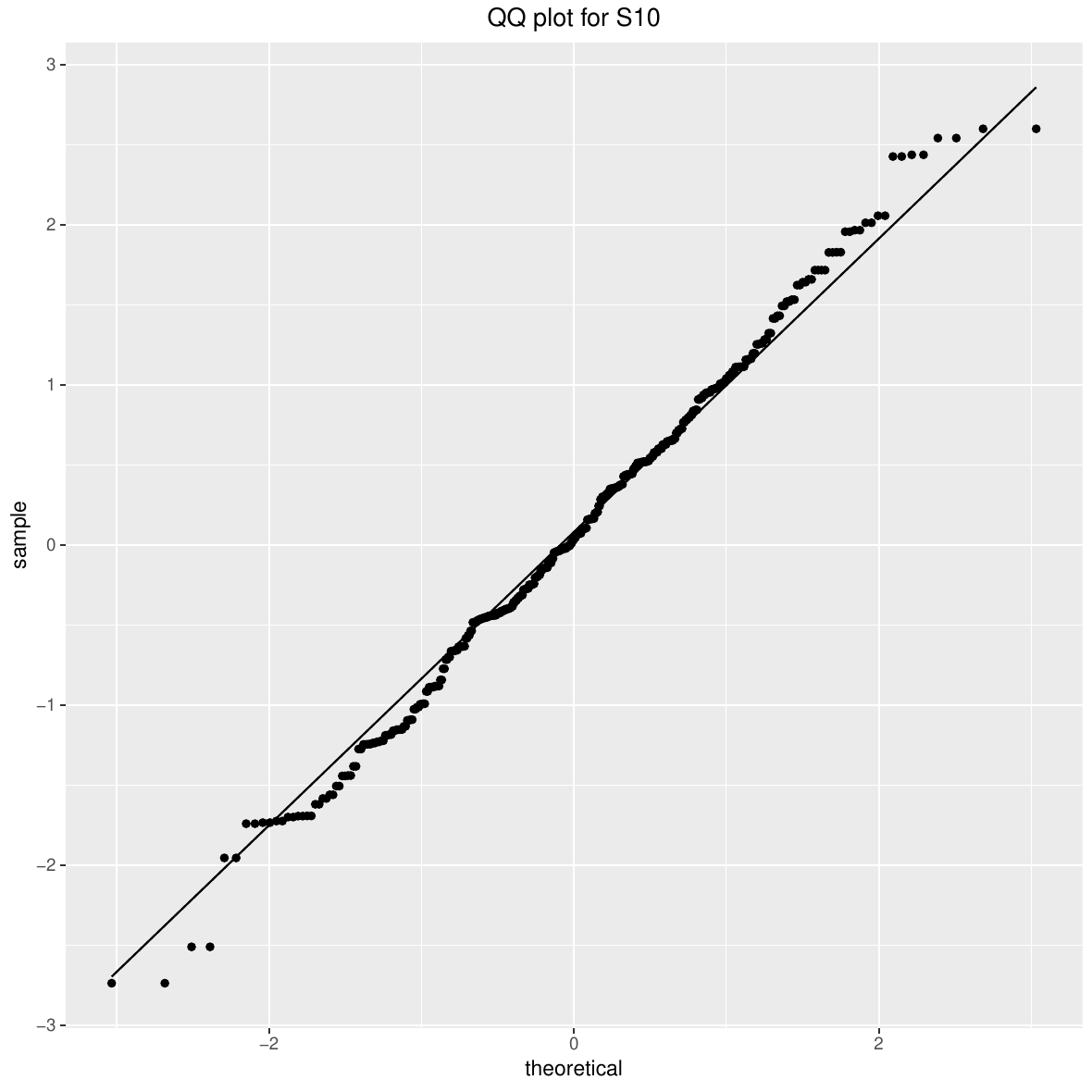}
\includegraphics[scale=0.16]{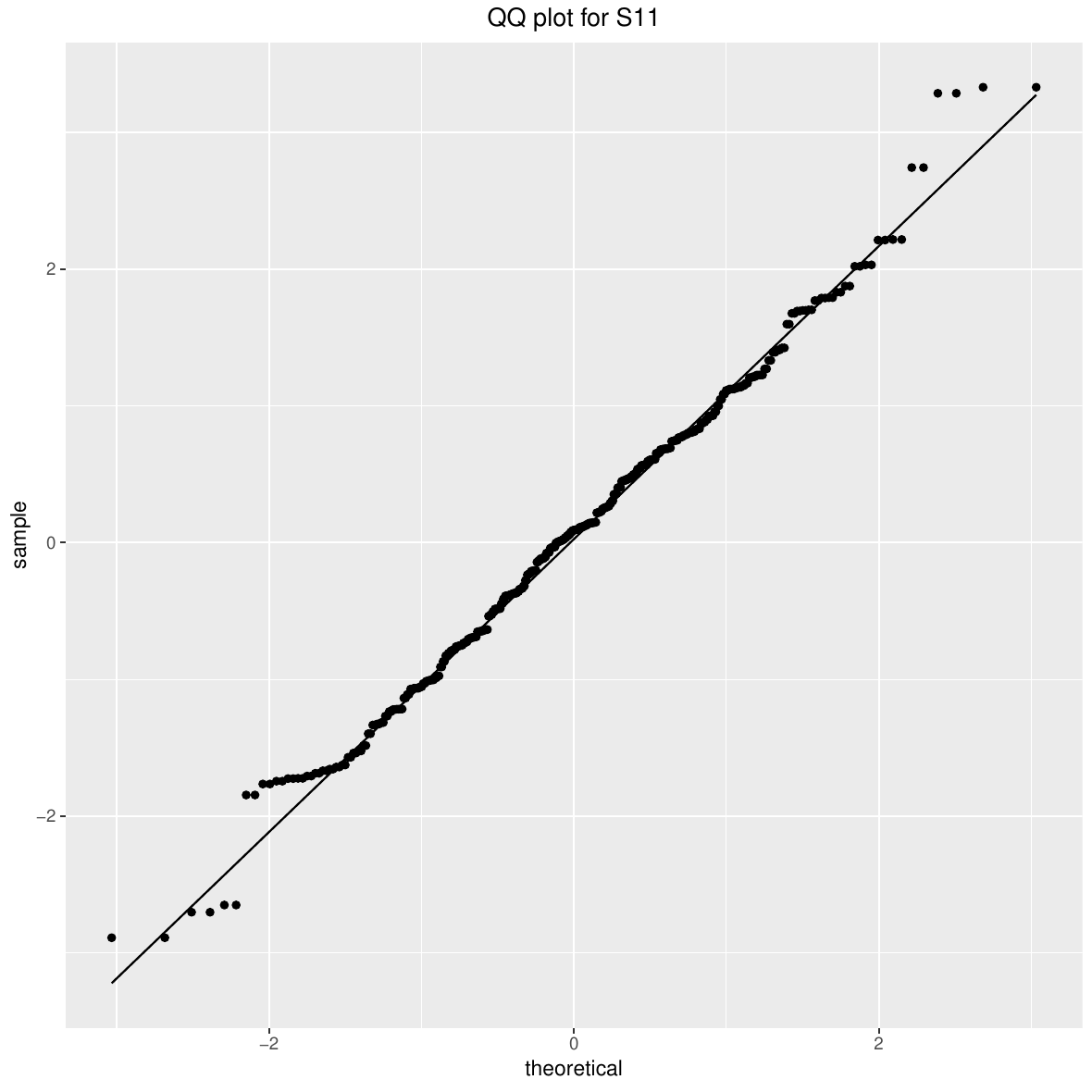}
\includegraphics[scale=0.16]{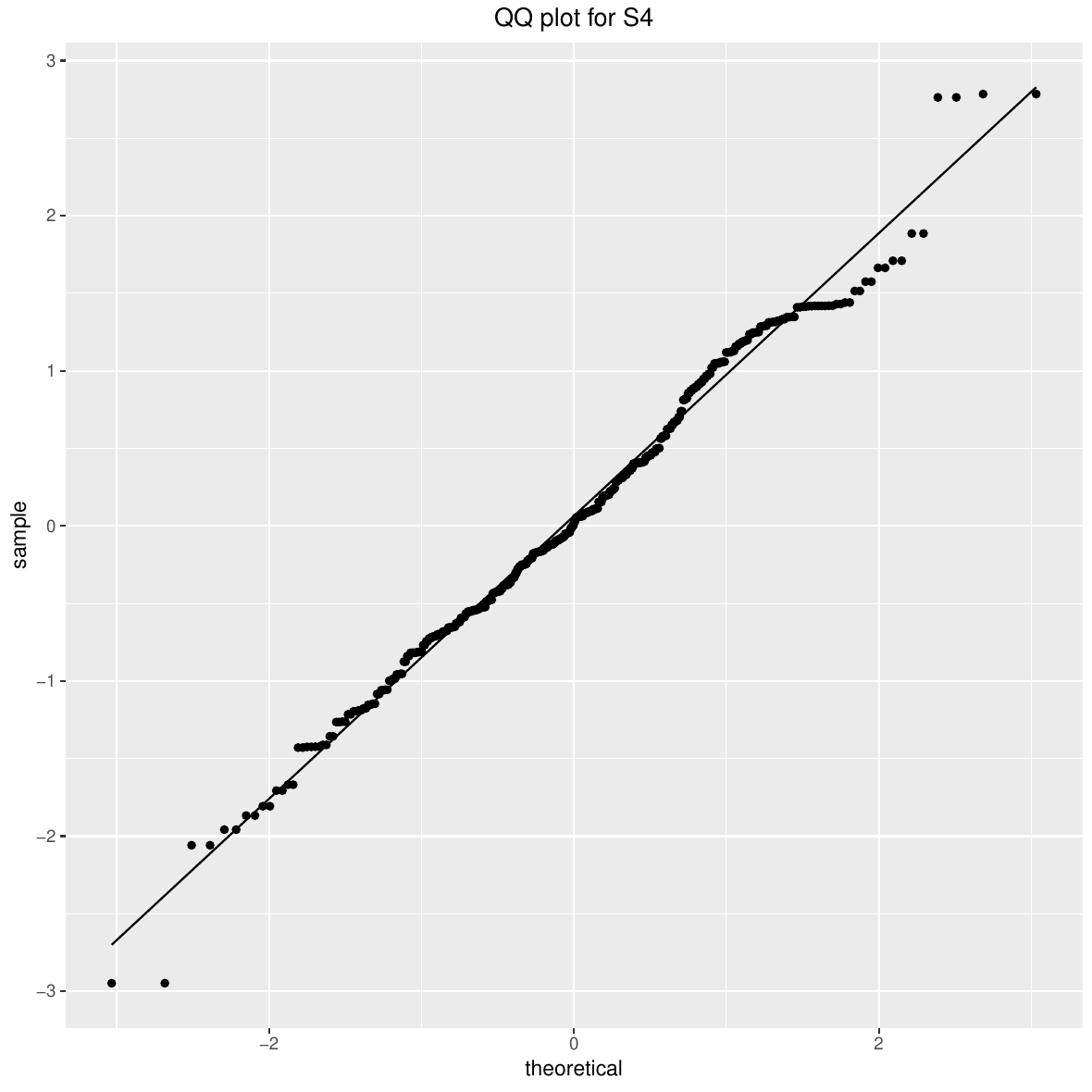}
\includegraphics[scale=0.16]{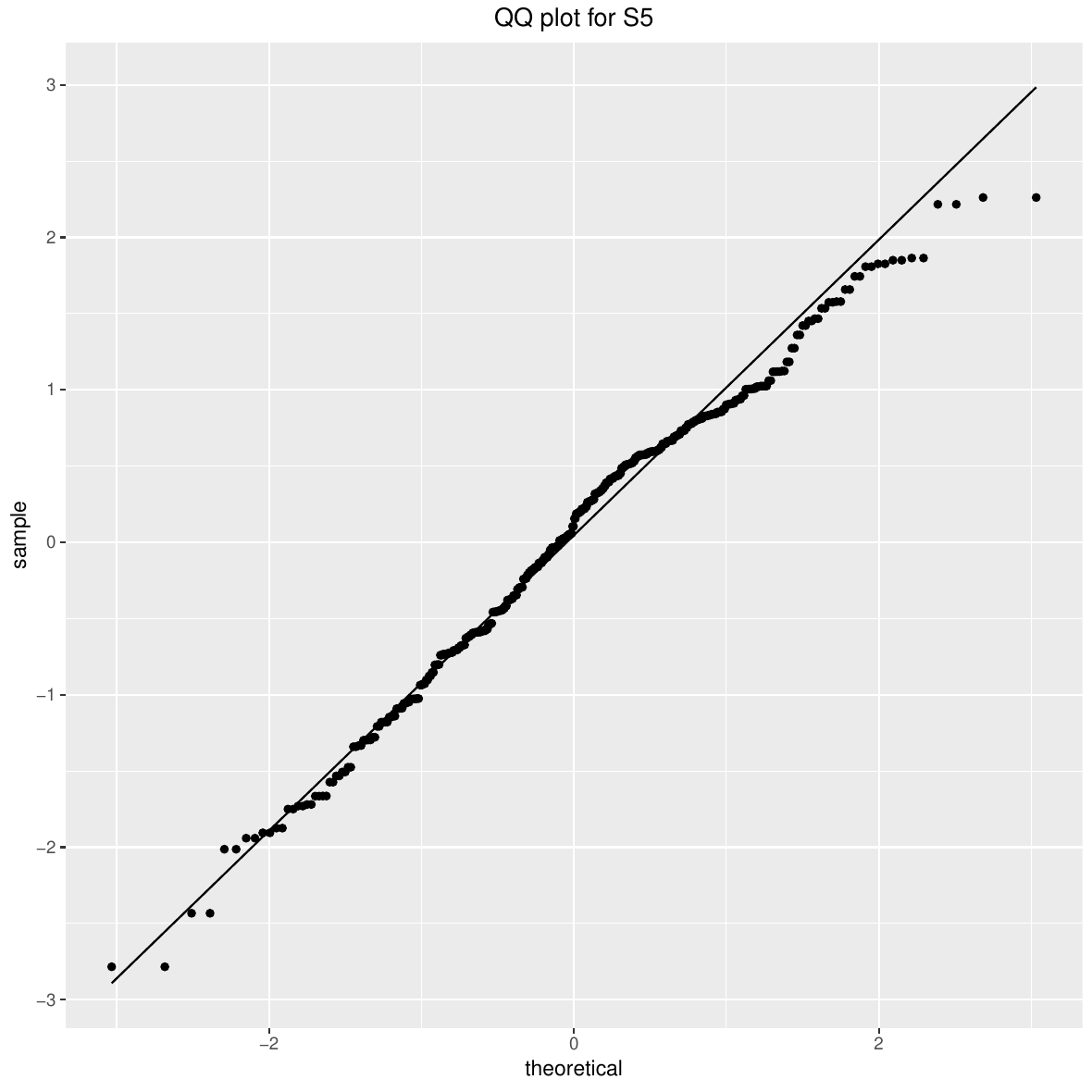}
\includegraphics[scale=0.16]{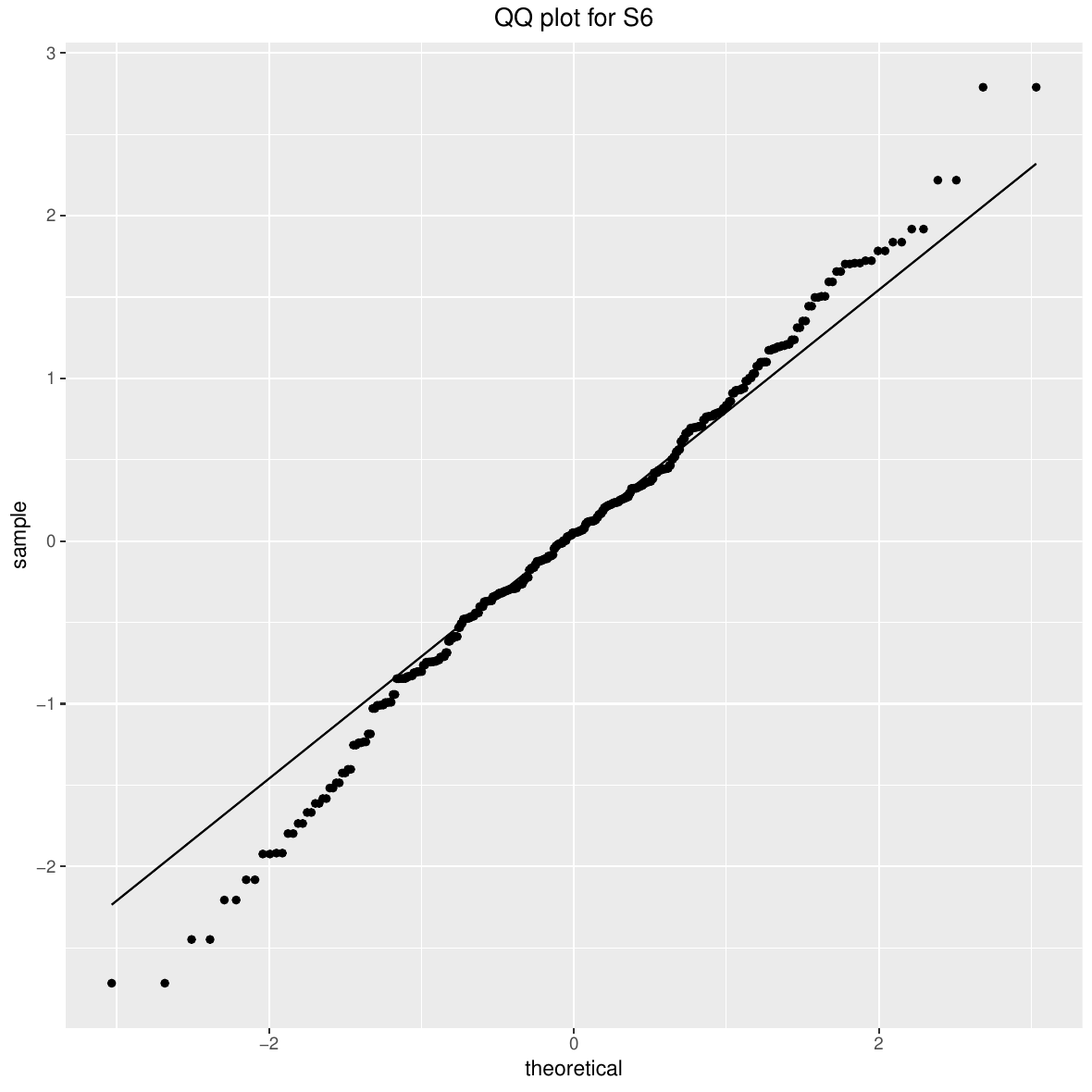}
\caption{QQ plots for fixed zero entry $[\bOmega_1^*]_{6,1}$. From left column to right column is scenario s4, s5, s6, s7, and s8. The first row is simulation 1, and the second  is simulation 2.}
\label{fig: add_consist_samplesize}
\vskip 0em
\end{figure}

The asymptotic normality of test statistic $\tilde{\tau}_{i,j}$ is illustrated in Figure \ref{fig: add_consist_samplesize}. These QQ plots of test statistic are for the same zero entry $[\bOmega_1^*]_{6,1}$ as in \S\ref{sec:simulation}.  As shown in the figure, our test statistic behaves very similar to standard normal even when sample size is extremely small (scenario s4).

The valid of our FDR control procedure is demonstrated in Table \ref{tab:FDP_dimension}, which contains empirical FDP, its theoretical limit $\tau$, and power (all in $\%$) of the Kronecker product of precision matrices under FDR control. 
Each mode has the same pre-specific level $\upsilon = 5\% $ or $10\%$. 
Empirical power of sim2 on scenario s4, s5, and s6 shows that, as sample size grows, power under FDR control would converge to one. 
Also, the low power of s4 in simulation 2 suggests that, if dimension of tensor is $10\times 10 \times 10$ and true precision matrices are generated from a nearest-neighbor like graph, we need sample size at least 20 to ensure acceptable power.

\begin{table}[htb]
  \small{
     \caption{Empirical $\textrm{FDP}$, its theoretical limit $\tau$, and $\textrm{power}$ (all in $\%$) of inference for the Kronecker product of precision matrices under FDR control in scenario s4, s5, s6, s7, and s8.}
	\label{tab:FDP_samplesize}
    \begin{center}
      \vskip -1 em
      \begin{tabular}{|p {0.5em}| p {2.5em} | p {2em} p {2em}p {2em}p {2em}p {2em} |p {2em}p {2em}p {2em}p {2em}p {2em}|}
        \hline 
        	 \multirow{2}{*}{}&  & \multicolumn{5}{c|  }{Sim1} & \multicolumn{5}{c   }{Sim2}  \vline \\[4pt]
	    $\upsilon$ & & s4   & s5 & s6  & s7& s8   & s4   & s5 & s6  & s7& s8   \\[4pt] \hline  
			 &&  \multicolumn{10}{c|}{Empirical FDP ($\tau$)} \\[4pt] \hline
	\multirow{2}{*}{5}  & oracle & 6.2  & 7  & 8.6 & 7.4  & 7.2  & 4.6  & 6.8  & 6.5  & 6.9 & 7.8\\[4pt]
			\hhline{~-----------} 
			& data-driven & 6.8 (9.9) & 7.3 (9.9) & 6.6 (9.9) & 7.3 (9.9) & 7.3 (9.9)& 5.2 (10.5) & 7.1 (11) & 7 (11) & 6.2 (11.1) & 7 (11.1)\\[4pt]
		\hline
	\multirow{2}{*}{10}  & oracle & 13.2  & 14.3  & 17.1  & 15.6  & 14.8  & 11.4 & 14.1 & 13.2  & 13.6  & 15.6 \\[4pt]
			\hhline{~-----------} 
			& data-driven & 14.2 (19.3) & 14.5 (19.3) & 13.9 (19.3) & 15.3 (19.3) & 15.6 (19.3)& 12.1 (20.7) & 14.3 (21.3) & 14.2 (21.4) & 13.2 (21.4) & 13.9 (21.4)\\[4pt]
		\hline 
	&&  \multicolumn{10}{c|}{Empirical Power } \\[4pt]	\hline
	\multirow{2}{*}{5} & oracle & 99.9  & 100  & 100  & 100  & 100& 60.4  & 93.2  & 98.4  & 100  & 100 \\[4pt]
			\hhline{~-----------} 
				& data-driven & 99.7  & 100  & 100  & 100  & 100 & 63.8  & 92.3  & 98.3  & 100  & 100 \\[4pt]
			\hline
	\multirow{2}{*}{10} & oracle & 99.9  & 100  & 100  & 100  & 100& 70.8  & 96.1  & 99.1  & 100  & 100 \\[4pt]
			\hhline{~-----------} 
			& data-driven & 99.8  & 100  & 100  & 100  & 100 & 74.2  & 95.2  & 99  & 100  & 100 \\
		\hline
	\end{tabular}
	\end{center} }
	\vskip -0.5em
\end{table} 
 

\noindent {\bf Dimensionality:} 
We then present numerical results on the scenarios of varying dimensionality. 
The effect of dimensionality are studied through the following four scenarios: 
\begin{itemize}
\item \textbf{Scenario s9:} sample size $n = 50$ and dimension $(m_1,m_2,m_3) = (10,20,20)$.
\item \textbf{Scenario s10:} sample size $n = 50$ and dimension $(m_1,m_2,m_3) = (20,20,20)$.
\item \textbf{Scenario s11:} sample size $n = 50$ and dimension $(m_1,m_2,m_3) = (10,10,30)$.
\item \textbf{Scenario s12:} sample size $n = 50$ and dimension $(m_1,m_2,m_3) = (10,20,30)$.
\end{itemize}
The rest setup is the same as the previous study on varying sample size.

\begin{figure}[htbp]
\centering
\includegraphics[scale=0.185]{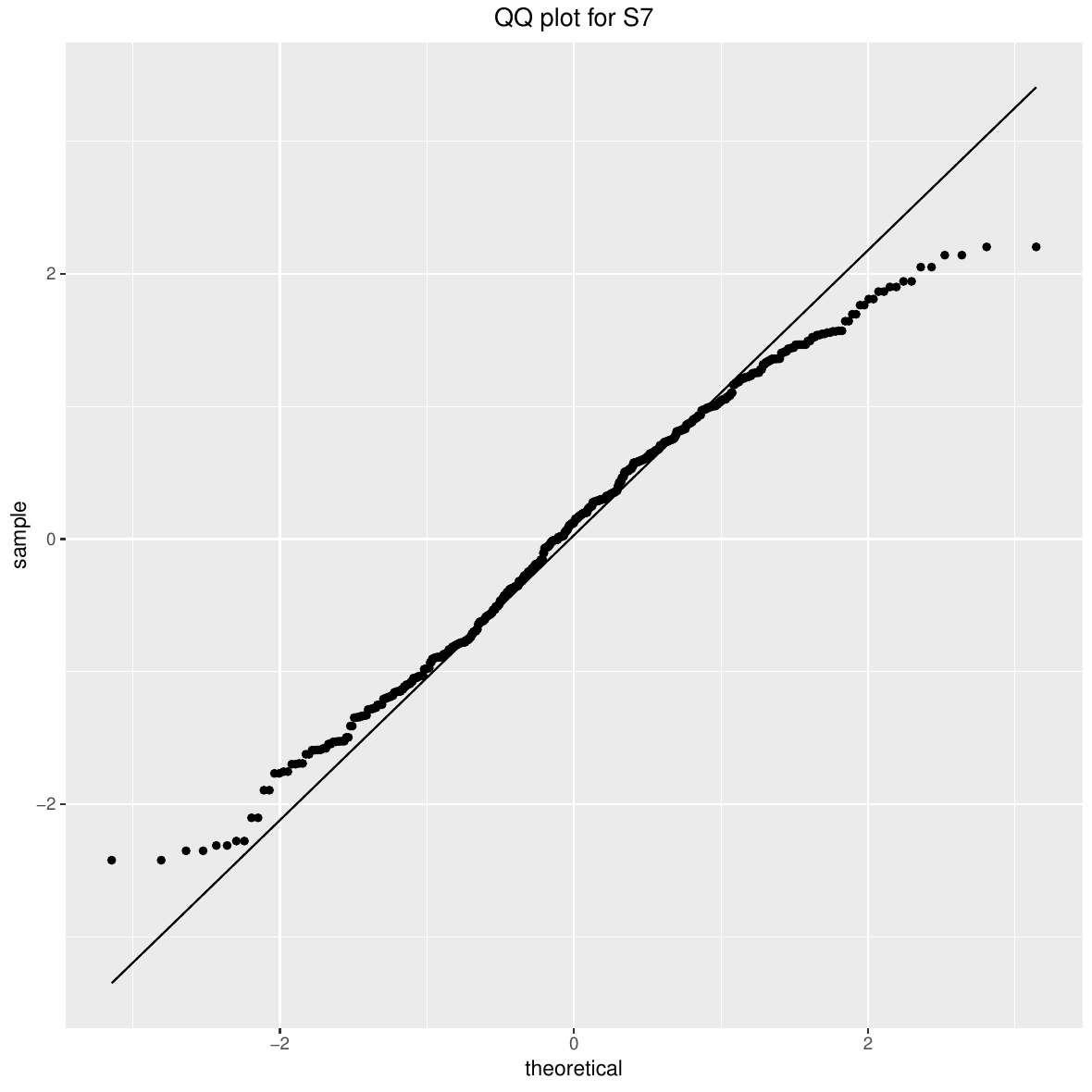}
\includegraphics[scale=0.185]{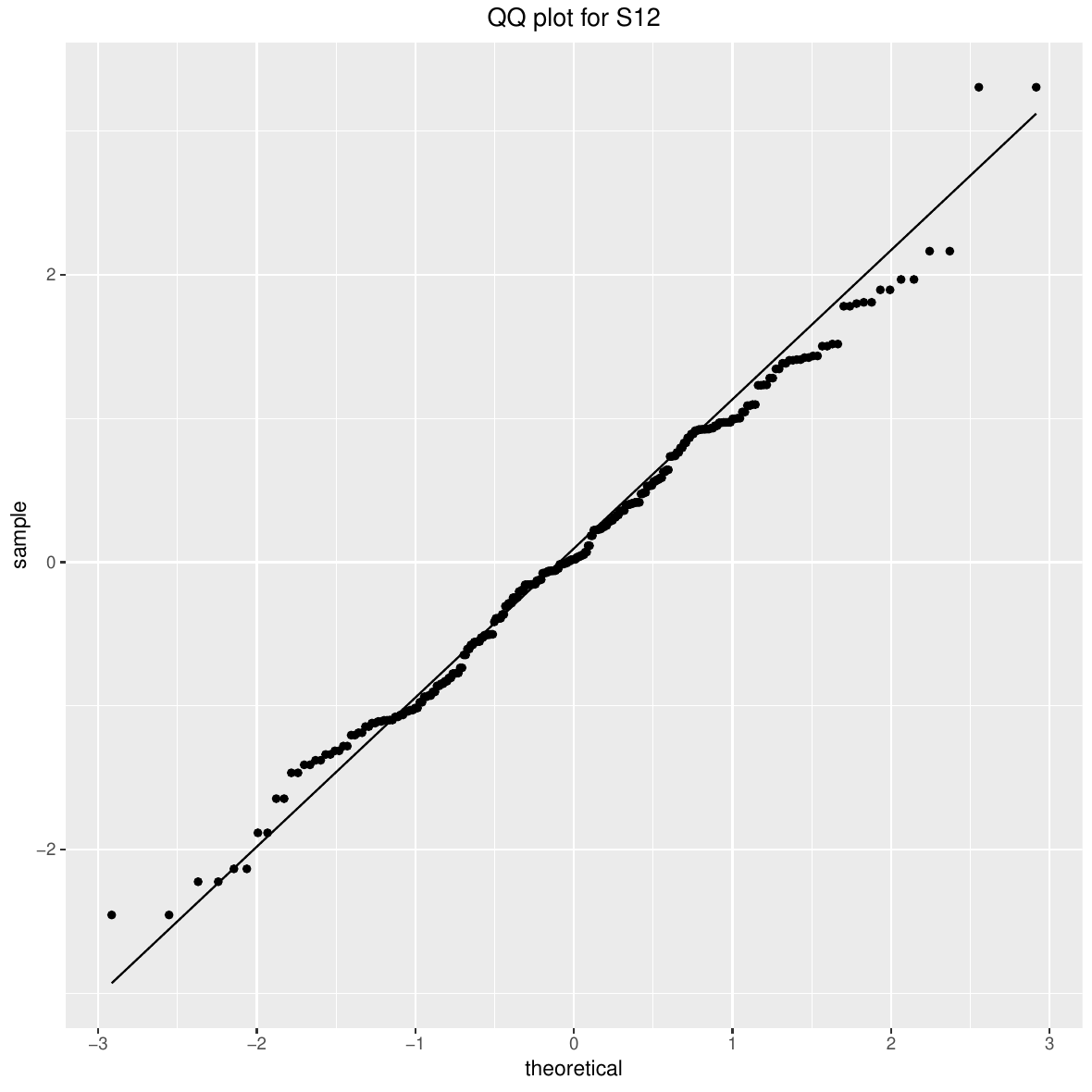}
\includegraphics[scale=0.185]{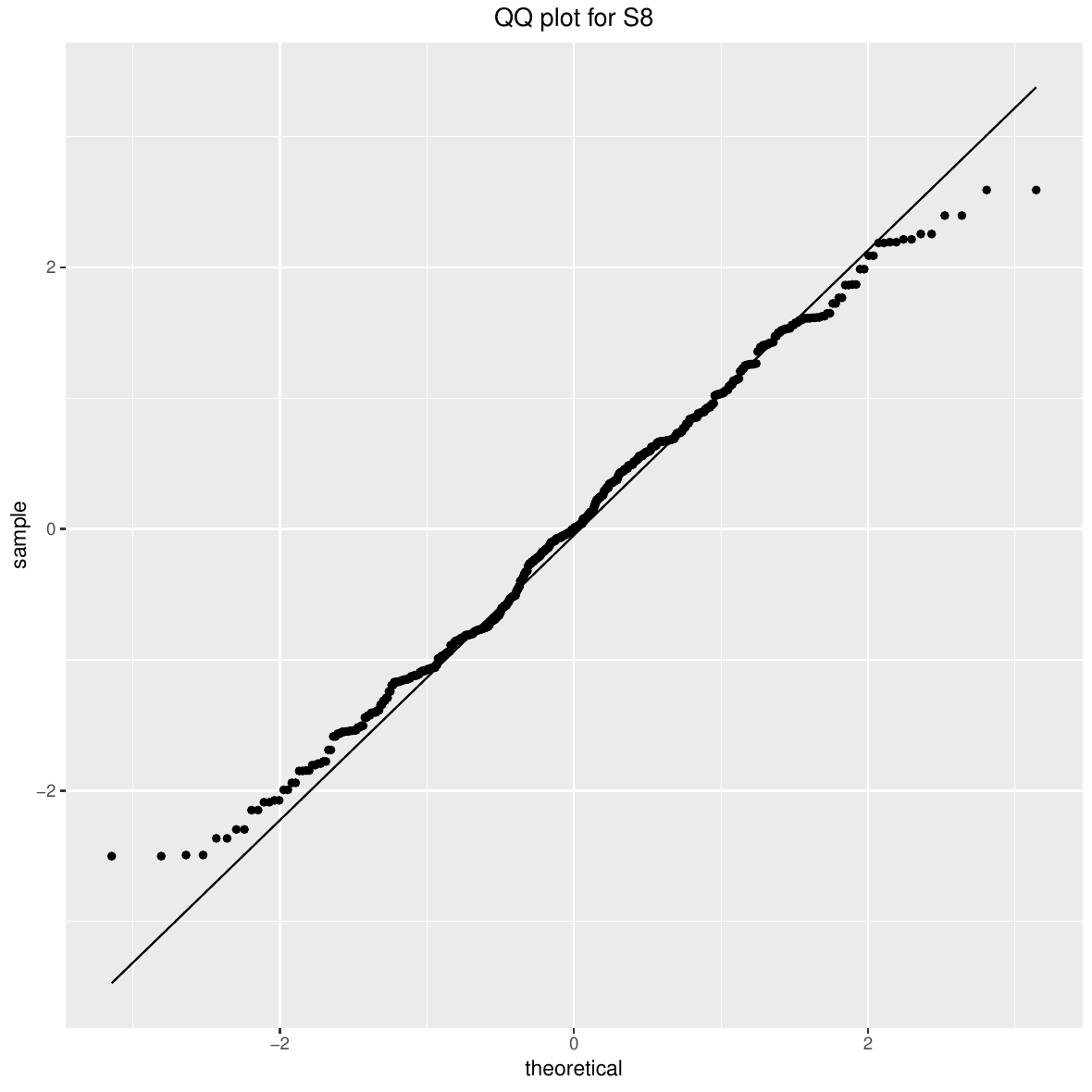}
\includegraphics[scale=0.185]{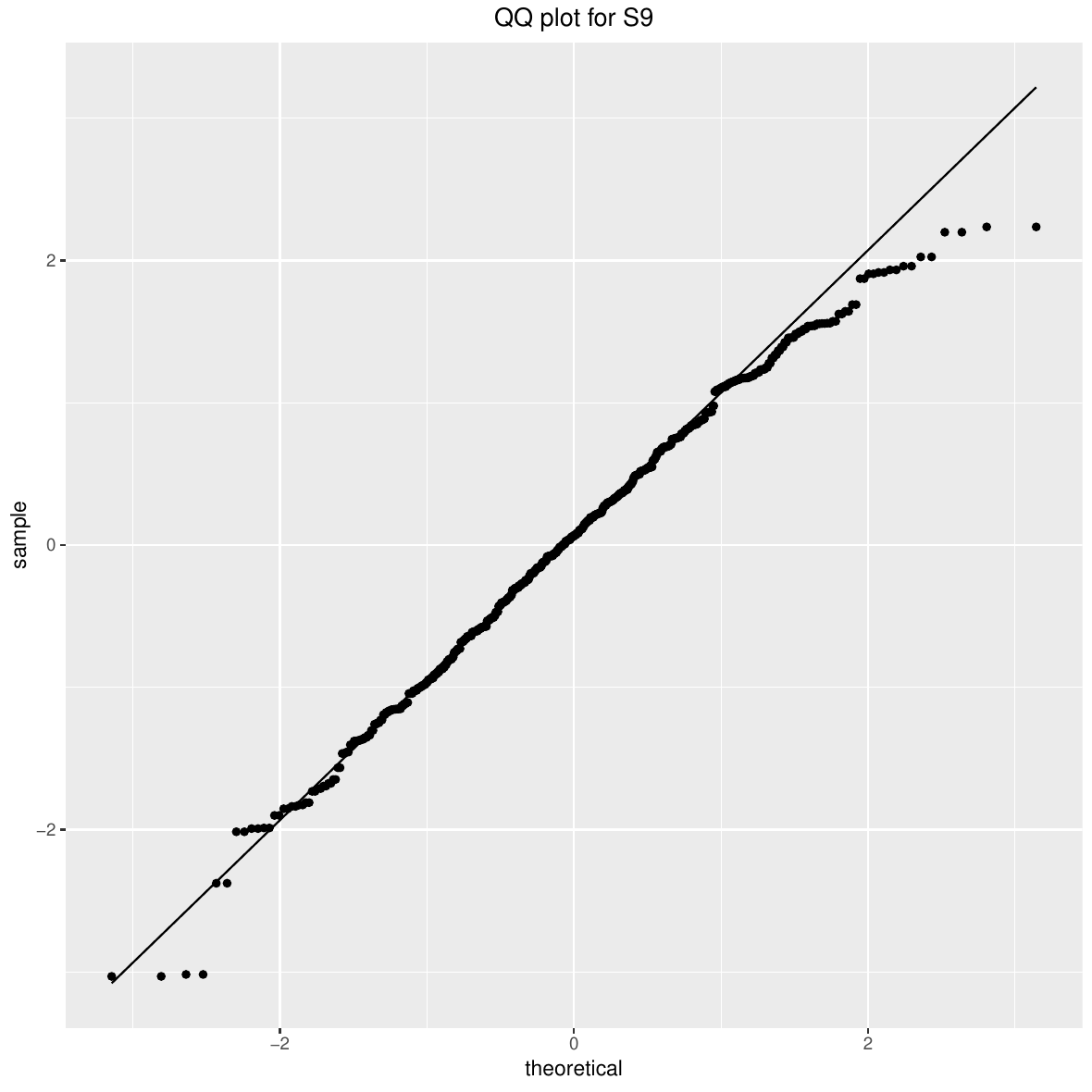}
\includegraphics[scale=0.185]{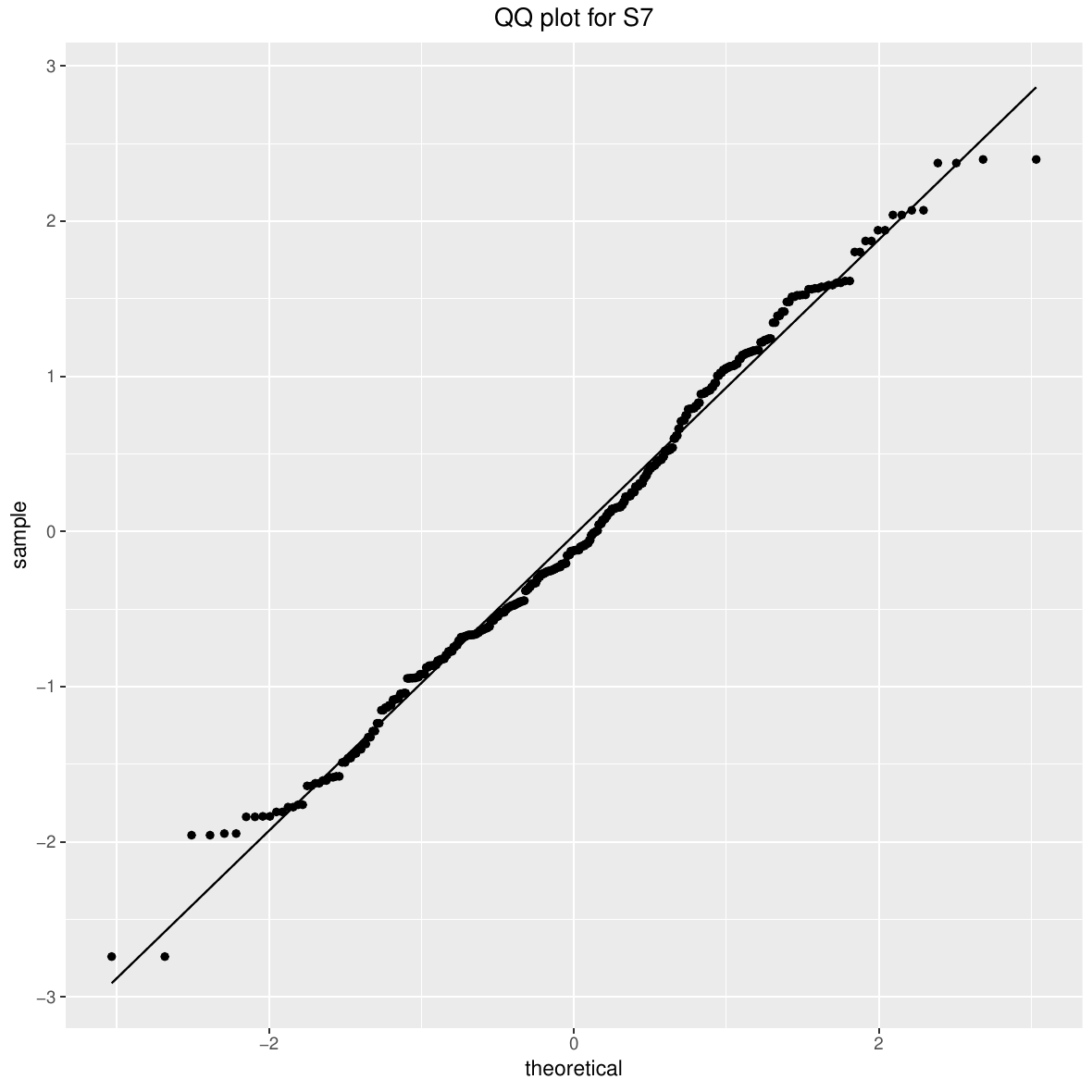}
\includegraphics[scale=0.185]{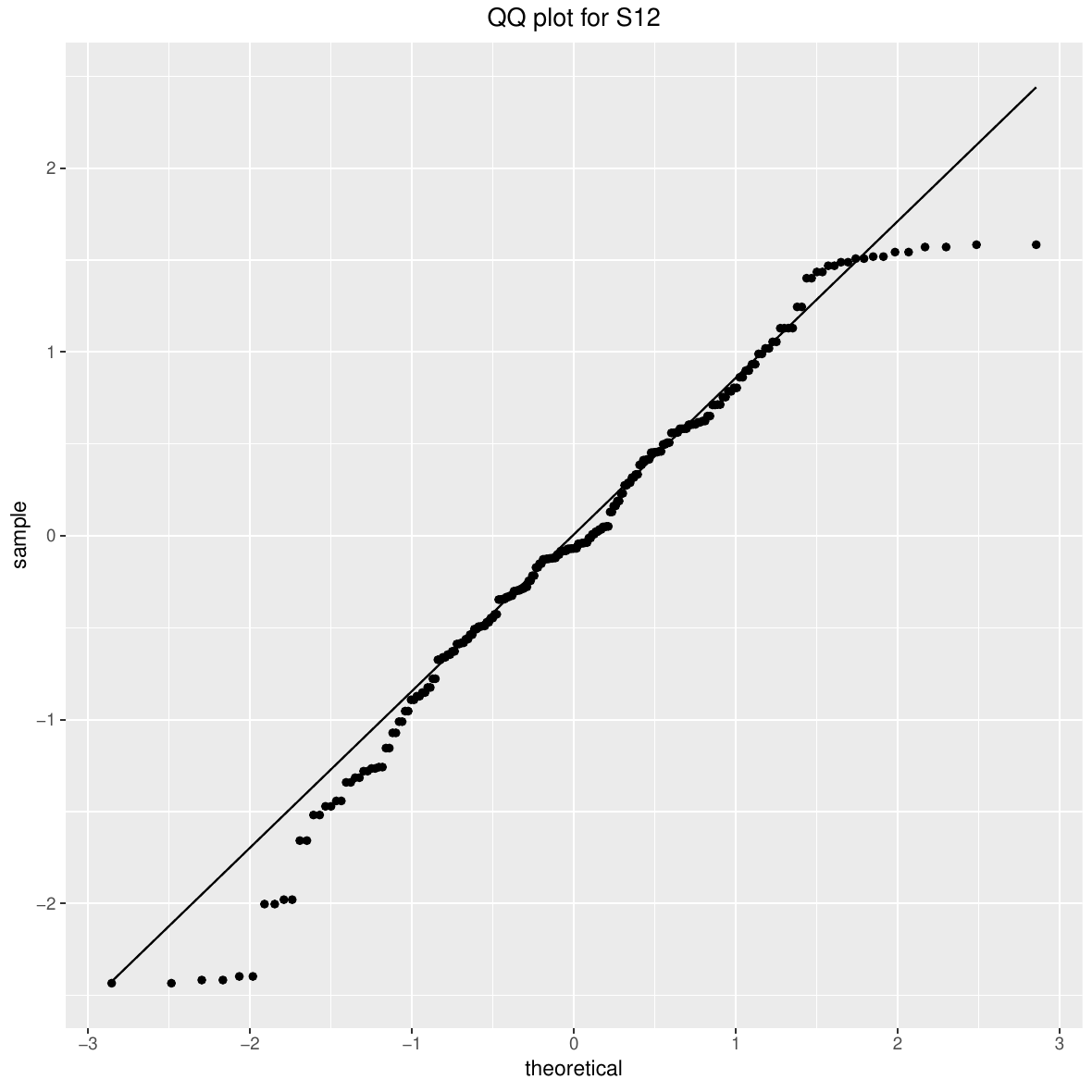}
\includegraphics[scale=0.185]{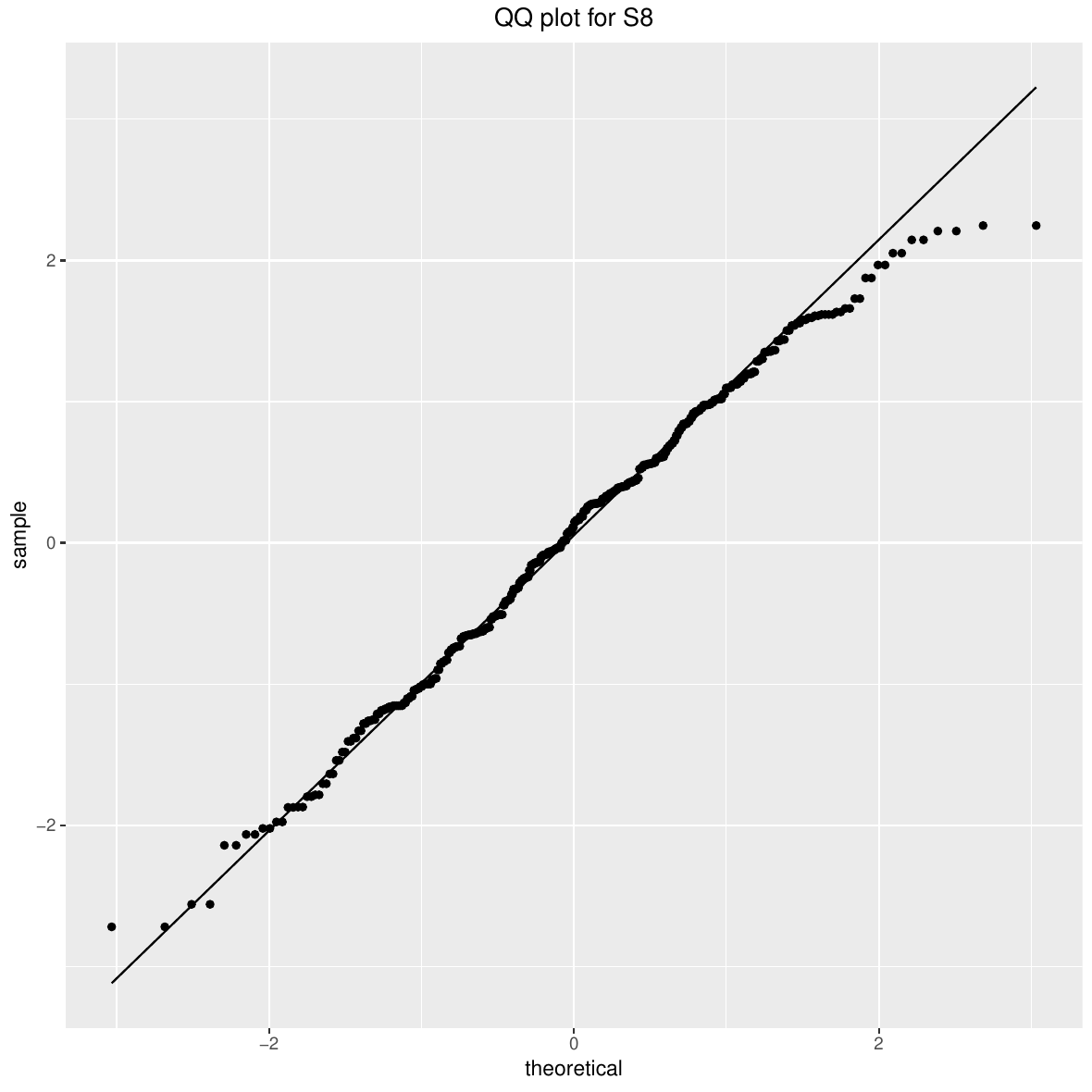}
\includegraphics[scale=0.185]{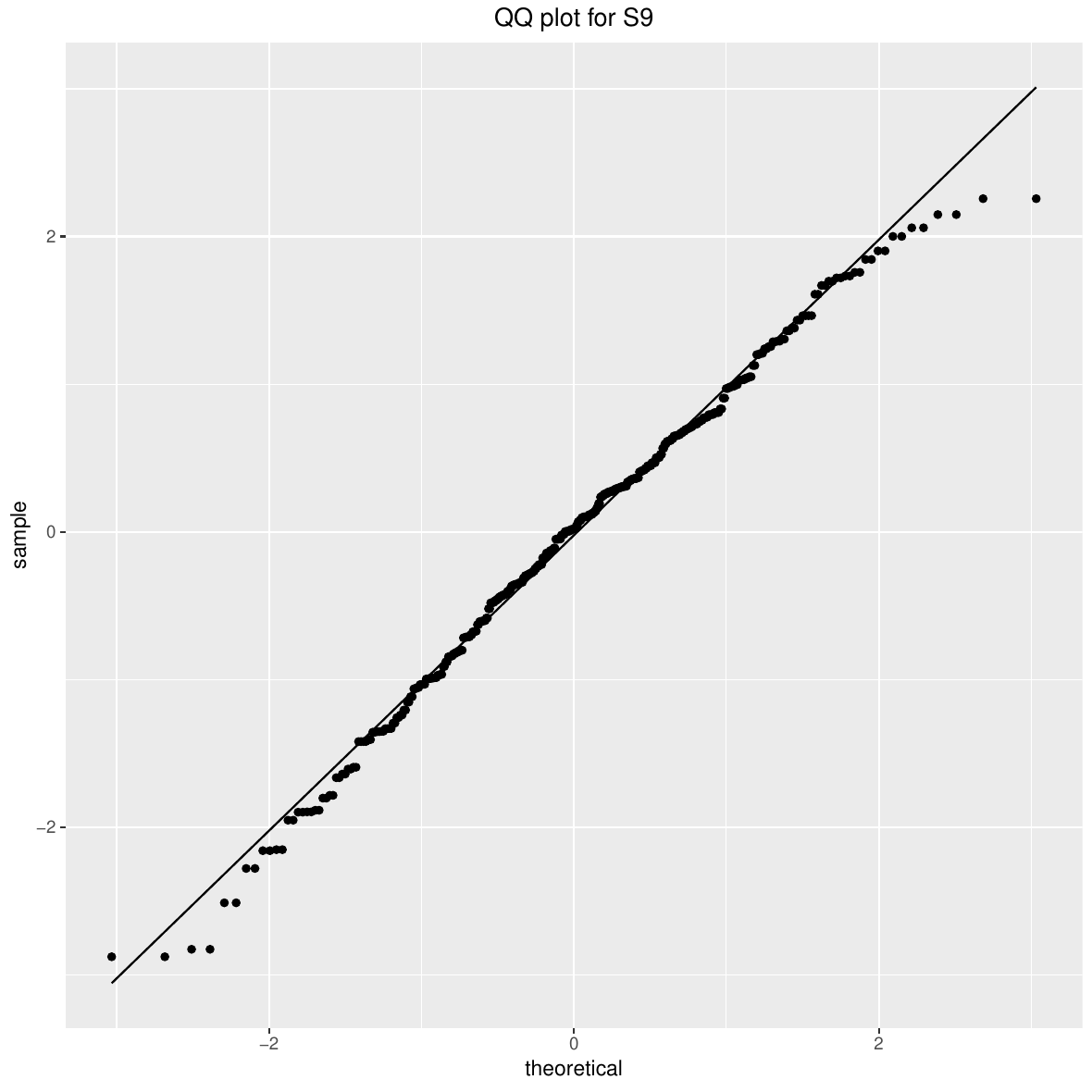}
\caption{QQ plots for fixed zero entry $[\bOmega_1^*]_{6,1}$. From left column to right column is scenario s9, s10, s11, and s12. The first row is simulation 1, and the second  is simulation 2.}
\label{fig: add_consist_dimension}
\vskip 0em
\end{figure}

Figure \ref{fig: add_consist_dimension} illustrates the asymptotic normality of test statistic $\tilde{\tau}_{i,j}$. QQ plots in the figure suggests that, even when dimensionality is very high (s10 and s12), our test statistic is still close to standard normal.

Table \ref{tab:FDP_dimension} summarizes empirical FDP, its theoretical limit $\tau$, and power (all in $\%$) of the Kronecker product of precision matrices under FDR control. 
Empirical FDP of  s9 v.s. s10, and s11 v.s. s12 shows that FDP gets closer to its theoretical limit $\tau$ if dimension grows. 
Besides, empirical power of s11 v.s. s12 in sim2 demonstrates that higher dimension also brings higher power. 
These phenomena not only suggest that our inference procedure fully utilizes tensor structure, but also back up the theoretical justification in Theorem \ref{thm: FDR_control}.

\begin{table}[htb]
  \small{
      \caption{Empirical $\textrm{FDP}$, its theoretical limit $\tau$, and $\textrm{power}$ (all in $\%$) of inference for the Kronecker product of precision matrices under FDR control in scenario s9, s10, s11, and s12.}
	\label{tab:FDP_dimension}
    \begin{center}
      \vskip -1 em
      \begin{tabular}{|p {0.5em}| p {2.7em} | p {2em} p {2em}  p {2em} p {2em}  |p {2em} p {2em}  p {2em} p {2em} |}
        \hline 
        	 \multirow{2}{*}{}&  & \multicolumn{4}{c | }{Sim1 } & \multicolumn{4}{c  }{Sim2}   \vline \\[4pt]
	    $\upsilon$ & & s9   & s10 & s11  & s12 & s9   & s10 & s11  & s12   \\[4pt] \hline 
			 &&  \multicolumn{8}{c|}{Empirical FDP ($\tau$)} \\[4pt] \hline
	\multirow{2}{*}{5}  & oracle & 7.9  & 9.3  & 7.6  & 8 & 8.1  & 10.1 & 9   & 8.4  \\[4pt]
			\hhline{~---------} 
			 & data-driven &7.7 (10) & 8.7 (10) & 7.3 (9.9) & 8.1 (10)  &8.1 (11.1) & 9.1 (11.1) & 8.2 (11.1) & 8 (11.1)\\[4pt]
			 \hline
	\multirow{2}{*}{10}  & oracle & 16.7  & 17.8  & 16  & 16.9  & 16.1  & 18.3  & 16.1  & 16.3 \\[4pt]
			\hhline{~---------} 
			 & data-driven &  15.2 (19.5) & 17.3 (19.6) & 14.8 (19.4) & 15.8 (19.5) & 16.6 (21.5) & 17.8 (21.5) & 15.7 (21.5) & 16.4 (21.5)\\[4pt]
		\hline 
	&&  \multicolumn{8}{c|}{Empirical Power  } \\[4pt]	\hline
	\multirow{2}{*}{5}  & oracle & 100 &100 &100 &100 & 100  & 100  & 99.8  & 100  \\[4pt]
			\hhline{~---------} 
				 & data-driven & 100 &100 &100 &100  & 100  & 100  & 99.7  & 100  \\[4pt]
			\hline
	\multirow{2}{*}{10} & oracle & 100 &100 &100 &100 & 100  & 100  & 99.9  & 100  \\[4pt]
			\hhline{~---------} 
	 		 & data-driven & 100 &100 &100 &100 & 100  & 100  & 99.8  & 100  \\[4pt]
 	\hline
	\end{tabular}
	\end{center} }
	\vskip -0.5em
\end{table} 

\end{document}